\documentclass{article}

\usepackage{microtype}
\usepackage{graphicx}
\usepackage{subfigure}
\usepackage{booktabs} % for professional tables

\usepackage{hyperref}

\PassOptionsToPackage{dvipsnames}{xcolor}
\usepackage[accepted]{arXiv}
\usepackage{amsmath}
\usepackage{amssymb}
\usepackage{mathtools}
\usepackage{amsthm}

\theoremstyle{plain}
\newtheorem{theorem}{Theorem}[section]
\newtheorem{proposition}[theorem]{Proposition}
\newtheorem*{proposition*}{Proposition}
\newtheorem{lemma}[theorem]{Lemma}
\newtheorem{corollary}[theorem]{Corollary}
\theoremstyle{definition}
\newtheorem{definition}[theorem]{Definition}
\newtheorem{assumption}[theorem]{Assumption}
\theoremstyle{remark}

\icmltitlerunning{Distributed Event-Based Learning via ADMM}

\usepackage{amsfonts}       % blackboard math symbols
\usepackage{nicefrac}       % compact symbols for 1/2, etc.
\usepackage{amsmath}
\usepackage{amsthm}
\usepackage{dsfont}

\usepackage{pgfplots}
\pgfplotsset{compat=1.18} % Specify the compatibility mode
\usepackage{tikz}
\usetikzlibrary{plotmarks}
\usetikzlibrary{arrows.meta,bending,positioning,calc} 
\usetikzlibrary[shapes,arrows]
\usepgfplotslibrary{fillbetween}

\usepackage{etoolbox}
\usepackage{accents}
\MakeRobust{\underaccent} % make \underaccent not fragile in moving arguments
\usepackage{dsfont}
\usepackage{duckuments}
\usepackage{algorithm}
\usepackage{algorithmic}
\usepackage{multirow}
\usepackage{graphicx}

\newcommand{\Cd}[0]{\mathcal{C}^d}
\newcommand{\grg}[0]{\nu}
\usepackage{svg}
\usepackage{caption}
\usepackage{booktabs}

\usepackage{filecontents}

\begin{document}
\twocolumn[\icmltitle{Distributed Event-Based Learning via ADMM}
\begin{icmlauthorlist}
\icmlauthor{Guener Dilsad Er}{mpi} 
\icmlauthor{Sebastian Trimpe}{rwth} 
\icmlauthor{Michael Muehlebach}{mpi}
\end{icmlauthorlist}
\icmlaffiliation{mpi}{Max Planck Institute for Intelligent Systems,  Tuebingen, Germany}
\icmlaffiliation{rwth}{Institute for Data Science in Mechanical Engineering, RWTH Aachen University, Aachen, Germany}
\icmlcorrespondingauthor{Guener Dilsad Er}{gder@tue.mpg.de}
\vskip 0.3in
]

\printAffiliationsAndNotice{} 
\begin{abstract}
We consider a distributed learning problem, where agents minimize a global objective function by exchanging information over a network. Our approach has two distinct features: (i) It substantially reduces communication by triggering communication only when necessary, and (ii) it is agnostic to the data-distribution among the different agents. We therefore guarantee convergence even if the local data-distributions of the agents are arbitrarily distinct. We analyze the convergence rate of the algorithm both in convex and nonconvex settings and derive accelerated convergence rates for the convex case. We also characterize the effect of communication failures and demonstrate that our algorithm is robust to these. The article concludes by presenting numerical results from distributed learning tasks on the MNIST and CIFAR-10 datasets. The experiments underline communication savings of 35\% or more due to the event-based communication strategy, show resilience towards heterogeneous data-distributions, and highlight that our approach outperforms common baselines such as FedAvg, FedProx, SCAFFOLD and FedADMM. 
\end{abstract}

\section{Introduction}\label{sec:intro}

Distributed learning refers to the minimization of a global objective function over a network of agents, where each agent has only access to a local cost function and can communicate with some or all agents in the network. Distributed learning systems provide a solution for handling the growing amount of data being generated everywhere on earth, by utilizing the computational power of individual devices in a network rather than relying on a central entity. This takes the burden off central processors and improves data privacy by avoiding a centralized training and storage of data. 

Distributed learning is particularly challenging when the data is not independent and identically distributed (non-i.i.d.) across the different agents. This situation often hinders the convergence to a globally optimal model. The non-i.i.d. nature leads to disparities in local datasets, preventing the local models from generalizing across the entire dataset, leading to a fundamental dilemma between minimizing local and global objective functions \citep{Acar_2021}. In addition, a second key challenge arises from the communication between agents, which is required to ensure convergence to the global solution and may lead to a substantial overhead. This communication overhead results in a waste of energy \citep{Li_Sahu_2020}, and is prone to delays and communication channel failures. As a result, both, non-i.i.d. datasets and communication overhead, constitute major bottlenecks for enabling large-scale learning systems. 

We provide an effective solution to both challenges. Inspired by the sent-on-delta concept \citep{miskowicz_send--delta_2006}, we reduce the communication load by introducing an event-based communication strategy, such that each agent (or computational node) communicates only if necessary. {Our communication rule enforces local constraints that collectively guarantee bounded overall error, a paradigm related to safe zone design strategies in distributed computing \citep{Garofalakis_2017}. Our approach is also rooted in event-based estimation, where communication is triggered by significant state changes.} We further base our approach on the Alternating Direction Method of Multipliers (ADMM). Our method is therefore robust against ill-conditioning and agnostic towards a disparity of the local data-distributions among the agents (these can be skewed in arbitrary ways). The approach further enables an explicit trade-off between communication load on the network and solution accuracy via a small set of hyperparameters that have a clear interpretation. We explicitly quantify the influence of these hyperparameters on the solution accuracy and analyze the effect of communication failures. The article concludes by highlighting the effectiveness of our algorithm in training neural networks, and solving LASSO problems in a distributed and communication-efficient manner.

Our theoretical analysis builds on a recent trend in the optimization literature \citep{jordan_variational_2016,su_differential_2016,muehlebach_icml,Tong_2023} that views algorithms as dynamical systems and leverages ideas from differential or symplectic geometry, as well as passivity and dissipativity \citep{Lessard_2016,Muehlebach_2020}. As we will show, this enables convergence proofs and a convergence rate analysis for our distributed algorithms, together with an analysis of robustness against communication failures. Our work provides important insights into the behavior of the event-based optimization under communication failures, an aspect, which has been overlooked in prior works, and thereby lays the groundwork for future research in this area.

\textbf{Related Work:} In the 1980s, \citet{Bertsekas_Tsitsiklis_1989} and others laid the foundation for the analysis of distributed algorithms. As machine learning became popular, distributed learning emerged, specifically focusing on parallelizing computation for empirical risk minimization. \citet{Shokri_Shmatikov_2015} explored collaborative deep learning with multiple agents using distributed stochastic gradient descent, which was later coined federated learning \citep{mcmahan_communicationefficient_2017} and advanced by subsequent contributions \citep{kairouz_advances_2019,Asad_2023}. A unifying element in these works is the consensus problem, where agents agree on a common value or decision. This problem is both central to distributed optimization and is also a special instance of distributed optimization \citep{Wei_Ozdaglar_2012}. 

The trade-off between communication and computation is inevitable in distributed optimization \citep{nedic_network_2018}. Recent work by \citet{Cao_2023} categorizes communication-efficient distributed learning into four main strategies: (1) minimizing the number of communications, (2) compression, (3) managing resources (e.g., bandwidth), and (4) using game theoretical approaches. We focus our review on the first category that aligns with our work, and reduces communication by transmitting information only if necessary.  
A first line of work \citep[][and many more]{mcmahan_communicationefficient_2017,wei_liu_decentralized_2021,Reisizadeh_2020} proposes algorithms with a periodical exchange of model parameters either among all agents or randomly selected subsets for decreasing communication load. While this approach is particularly straightforward and easy to implement, the random sampling  risks missing critical updates or performing redundant communications.
%Accelerated Gradients
A second line of work involves accelerated gradient methods for distributed optimization, reducing the need for many communication rounds to converge. For instance, \citet{Kovalev_2020} and \citet{Nabli_2023} optimize the number of gradient evaluations together with communication rounds. \citet{Shamir_2014} replaces gradient descent with Newton-like methods and \citet{Hendrikx_2020} proposes statistical preconditioning where both methods further improve convergence rates at the cost of a higher computational load per iteration. 
Additionally, \citet{Liu_decentralized_2021} propose a lazy evaluation of dual gradients, reducing communication by skipping redundant updates, while \citep{Chen_LAG_2018} adaptively reuse lagged gradients to meet target accuracy with fewer communication rounds.
%Compression
There has also been a third line of work that focuses on reducing communication via event-based triggering and compression of network parameters \citep{Liu_2019,ghadikolaei_lena_2021,Singh_2023,Zhang_2023_journal}. While \citep{Zhang_2022_privacy, Zhang_2023_journal} employ an ADMM-based strategy that is similar to ours, their focus lies on investigating different compression schemes, and not on analyzing convergence rates and the effect of communication failures.
Event-triggering has also been explored in contexts like dynamics model learning \citep{solowjow2020event,umlauft2019feedback}, and Bayesian optimization \citep{brunzema2022event}. While highlighting the benefit of triggering for reducing communication, these works do not consider distributed optimization problems as we do herein. 

In addition to the communication overhead, another major challenge for distributed learning arises from non-i.i.d. data distributions across agents \citep{Zhao_Li_2018,Li_Yang_2020,glasgow2022sharp}. 
SCAFFOLD \citep{Karimireddy_SCAFFOLD_2020} addresses this challenge by introducing a client control variate to improve convergence at the cost of doubling communication. Similarly, \citep{Gao_2022} enhances training with auxiliary drift variables, while \citep{Zheng_Ye_2024} selects representative clients and adjusts server gradients. Recent contributions by \citet{li_federated_2020,Acar_2021,Shi_2023} add a proximal regularization term to the local objective functions of the individual agents, whereas  \citet{Zhang_FedPD_2021} (FedPD) and \citet{Zhou_Li_2023,Wang_FedADMM_2022,Gong_Li_Freris_2022} (FedADMM) address the challenge with ADMM formulations. However, compared to our work, FedADMM \citep{Zhou_Li_2023,Wang_FedADMM_2022,Gong_Li_Freris_2022} relies on utilizing a random selection of agents that communicate and FedPD \citep{Zhang_FedPD_2021} considers full participation, whereas we use an event-triggered mechanism. Alternatively, other splitting schemes such as Douglas-Rachford method proposed by \citet{Tran_feddr_2021}, similarly align local and global objectives but remain constrained by random agent participation. 
\looseness-1

ADMM remains a widely-used tool for distributed learning, with recent advancements focusing on improving convergence rates and communication efficiency. For example, \citet{Wang_Wang_Li_Lei_2025} introduce inertia and adaptive iteration strategies to accelerate convergence, while \citet{Song_FedADMMInSa_2025} controls inexactness and dynamically tunes penalty parameters. In addition, \citet{He_Zheng_Feng_Chen_2023} explore dynamic tuning of ADMM hyperparameters, and hierarchical grouping approaches \citep{Qiu_Lei_Wang_2023}, inspired by \citet{Elgabli_GADMM_2020}, aim to reduce communication overhead by restricting updates to neighboring workers. These methods share the goal of improving the efficiency of ADMM in distributed settings, but they still rely on periodic or full-agent participation, whereas our approach uses event-triggered mechanisms to further reduce communication costs.

As we also highlight in numerical experiments, a random selection of agents might prevent important local changes from propagating quickly through the network, leading to a slower convergence.
To the best of our knowledge, this is the first work to provide a convergence analysis of distributed learning with event-triggered communication that addresses key aspects, such as packet drop and the presence of non-i.i.d. data.
\looseness-1

\textbf{Contributions} are summarized as follows:\\
\textbf{(i)} We propose an event-based communication scheme for distributed optimization, where a communication event is only triggered, when the current state has deviated by a predefined threshold $\Delta$, indicating a significant change in the local decision variables. Therefore, our approach is effective in reducing communication overhead and can adapt to the limited communication resources in heterogeneous networks.
Our method is also compatible with and complementary to gradient compression/quantization \citep{hegazy2023compression, mao2022adaptivequant, wang2018atomo} and fair aggregation techniques \citep{zhu2021broadcast}.

%Better communication
\textbf{(ii)} We characterize the effect of the communication threshold $\Delta$ on the solution accuracy and therefore quantify the trade-off between communication and solution accuracy. Compared to other ADMM-based approaches, such as \citep{Zhang_FedPD_2021,Zhou_Li_2023}, our method is versatile, both in the selection of variables that are being communicated (which is important for reducing communication in practice), as well as the different problem formulations that we can address. In particular, our approach goes beyond the scope of consensus problems, and can solve generic constrained optimization problems, sparse regression and LASSO problems, perform robust principal component analysis \citep{Candes_2011}, and solve distributed learning instances where the features but not the data points are distributed \citep{Boyd_2010}.

%Handling non-i.i.d.
\textbf{(iii)} Numerical experiments support the theoretical analysis and highlight that our approach even converges in the most extreme non-i.i.d. setting, where each agent has only access to training data from a single class (see the MNIST classifier example in Sec.~\ref{sec:results}). Comparisons to the baselines FedADMM \citep{Zhou_Li_2023}, SCAFFOLD \citep{Karimireddy_SCAFFOLD_2020}, FedProx \citep{li_federated_2020} and FedAvg \citep{mcmahan_communicationefficient_2017} demonstrate superiority both in terms of communication efficiency and classification accuracy. 

%Analysis
\textbf{(iv)} We demonstrate an accelerated convergence rate, and derive symbolic expressions that relate the convergence rate to instance-specific quantities such as the condition number and the topology of the communication network. The convergence analysis requires a Lyapunov-like function that is different compared to earlier work \citep{Nishihara_2015}, due to the presence of the event-based communication.

%Robustness and failures
\textbf{(v)} We study the robustness of our algorithm against communication failures, both in theory as well as in numerical experiments, which, to the best of our knowledge, is largely missing in the literature (a notable exception for the consensus problem is \citep{Bastianello_2021}). We address communication failures algorithmically by proposing a rare periodic reset strategy. We show that, without such a reset strategy, inter-agent errors accumulate rapidly in the presence of packet drops and prevent convergence.

\textbf{Outline:} The article is structured as follows: Sec.~\ref{sec:problem} describes the problem formulation and introduces our event-based learning algorithm in the consensus setting. The more general formulation is discussed in Sec.~\ref{sec:ADMM_dynamics}, where we also introduce a dynamical systems model for our algorithm. Sec.~\ref{sec:convergence} discusses the convergence analysis of the proposed algorithm and presents convergence rates, while empirical results that underline the theoretical findings are included in Sec.~\ref{sec:results} and in App.~\ref{app:add_experiments}. The appendix contains additional technical details about the communication structure in App.~\ref{app:comm} and the details of the convergence analysis in App.~\ref{app:dynamics} and \ref{app:convergence}.

\section{Event-Based Distributed Learning}\label{sec:eventbased}\label{sec:problem} 

We consider a distributed learning problem of the type
$
\min_{x\in \mathbb{R}^n}~ \sum_{i=1}^Nf^i(x),
$
where the overall cost function $f(x)$ is the sum of $N$ individual, potentially nonsmooth functions. The different $f^i$ typically arise from different training datasets stored on different computational nodes. In the most basic instance, our algorithm arises from the consensus formulation
\begin{align}
\begin{aligned}
&\min_{x^1, \ldots, x^N \in \mathbb{R}^n}~ \sum_{i=1}^Nf^i(x^i) + g(z),\\
&\text { subject to }~ x^i=z, \quad i=1,\ldots,N,\end{aligned}
\label{eqn:DL_problem}
\end{align}
where we impose the constraints $x^i=z$ by corresponding dual variables $u^i$. Thus, by guaranteeing constraint satisfaction, we can ensure consensus between the agents despite different local problems and, in particular, arbitrary non-i.i.d. data distributions among the computational nodes. In addition, we assume the function $f^i: \mathbb{R}^n\rightarrow \mathbb{R}$ to be smooth, while $g\!: \!\mathbb{R}^n\rightarrow \bar{\mathbb{R}}$ is allowed to be nonsmooth ($g$ typically represents a regularizer) and maps to the extended real numbers.

\begin{figure}
    \centering    
    \tikzstyle{block} = [draw, rectangle, minimum height=2em, minimum width=3em]
\tikzstyle{largeblock} = [draw, rectangle, minimum height=1em, minimum width=5em]

\tikzstyle{block} = [draw, rectangle, minimum height=2em, minimum width=2em, fill=YellowGreen]
\tikzstyle{thickblock} = [draw, rectangle, minimum height=1em, minimum width=1em, thick]
\tikzstyle{smallblock} = [draw, rectangle, minimum height=1em, minimum width=1em]
\tikzstyle{longblock} = [draw, rectangle, minimum height=1em, minimum width=18em, thick, fill=Apricot]
\tikzstyle{input} = [coordinate]

\tikzstyle{output} = [coordinate]
\tikzstyle{pinstyle} = [pin edge={to-,thin,black}]

\begin{tikzpicture}[auto, node distance=2cm,>=latex',every node/.style={font=\small}]
    % Blocks
\node [block] (A1) {Agent 1};

\node [block, right of=A1, node distance=1.8cm] (A2) {Agent 2};
    
     \node [block, right of=A2, node distance=1.8cm] (A3) {Agent 3};
       \node [block, right of=A3, node distance=1.8cm] (A4) {Agent 4};

    \node [longblock,  below=1.5cm of $(A2)!0.5!(A3)$] (bus) {Agent 5};

    \draw [->,dashed] (A1.south) -- ([xshift=-2.7cm]bus.north)  node[near start, left, font=\tiny]{$d^1_{k+1}\!\!\!-\!\!d^1_{[k]}$};

    \draw [->,dashed] (A2.south) -- ([xshift=-0.9cm]bus.north) node[near start, left, font=\tiny]{$d^2_{k+1}\!-\!d^2_{[k]}$};
    \draw [->,dashed]  (A3.south)  -- ([xshift=0.9cm]bus.north) node[near start, left, font=\tiny]{$d^3_{k+1}\!-\!d^3_{[k]}$} ;
    \draw [->,dashed] (A4.south) --  ([xshift=2.7cm]bus.north) node[near start, left, font=\tiny]{$d^4_{k+1}\!-\!d^4_{[k]}$};
    
    \draw [->,dashed] ([xshift=-2.6cm]bus.north) -- ([xshift=0.1cm]A1.south) node[near start, right, font=\tiny]{$z_{k+1}\!-\!z_{[k]}$ };
    \draw [->,dashed] ([xshift=-0.8cm]bus.north) -- ([xshift=0.1cm]A2.south) node[near start, right, font=\tiny]{$z_{k+1}\!-\!z_{[k]}$ };
    \draw [->,dashed] ([xshift=1.0cm]bus.north) -- ([xshift=0.1cm]A3.south) node[near start, right, font=\tiny]{$z_{k+1}\!-\!z_{[k]}$ };
    \draw [->,dashed] ([xshift=2.8cm]bus.north) --  ([xshift=0.1cm]A4.south) node[near start, right, font=\tiny]{$z_{k+1}\!\!-\!\!z_{[k]}$ };

\end{tikzpicture}
    \caption{The figure illustrates the distributed learning setup.  The Agents $1\!-\!4$ store $x^i$, $u^i$ and perform updates based on the information received by Agent $5$, according to Alg.~\ref{alg:over_relaxed_consensus}. Agent $5$, stores $z$ and performs updates based on the information received by Agent $1\!-\!4$. This architecture is common in distributed learning, where a single server aggregates updates from multiple distributed clients to collaboratively train a model.}
    \label{fig:dist_lr_server}
\end{figure}
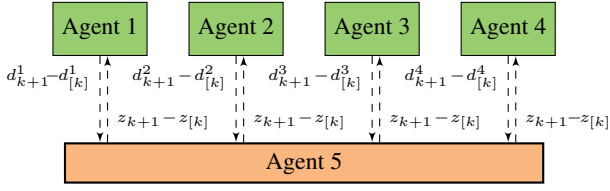

Our event-based algorithm, stated in Alg.~\ref{alg:over_relaxed_consensus}, works as follows. Each agent (or computational node) $i$, $i=1,\dots,N$, has access to the local objective function $f^i$, its local solution $x^i$, its local multiplier $u^i$, and an estimate $\hat{z}^i$ of the consensus variable $z$. We further introduce the agent $N\!+\!1$ (acting as server) that has access to $g$, the variable $z$, and maintains an estimate $\hat{\zeta}$ of the average 
\begin{align*}
    \zeta_k:=\frac{1}{N} \sum_{i=1}^N \left( \alpha x^i_{k+1}+u_k^i\right).
\end{align*} 
Following the communication structure in Fig.~\ref{fig:dist_lr_server}, the algorithm proceeds in two steps:

\textbf{i) Parallel update of agents $i=1,\dots,N$:} Each agent $i=1,\dots,N$ first updates its estimate $\hat{z}_k^i$ based on whether it receives an event-based communication from the agent $N\!+\!1$. The agent then updates its multiplier $u_k^i$ and solves a local minimization over $f^i$, which also includes a quadratic regularization term. The regularization term ensures that the minimization is well-conditioned (a key advantage to dual ascent, for example) and the local solution $x^i$ is biased towards $\hat{z}_k^i$. In practice, the minimization is replaced by a fixed number of (stochastic) gradient descent steps. If the resulting value $d^i_{k+1}:=\alpha x_{k+1}^i\!+\!u_{k}^i$ of the agent $i$ is significantly different from the value that it last communicated to the agent $N\!+\!1$, an event-based communication is triggered and the difference of $d^i_{k+1}$ to the last communicated value is sent to the agent $N\!+\!1$.

\textbf{ii) Update of agent $N\!+\!1$:} The agent $N\!+\!1$ updates its estimate $\hat{\zeta}$ of $\zeta$ by accumulating the $d^i$ variables that it receives from all agents. It then updates the consensus variable $z_{k+1}$ by solving a local minimization over $g$ with a quadratic regularization term. Note that if the nonsmooth component $g$ is missing, $z_{k+1}$ is simply set to $\hat{\zeta}_{k}\!-\!(1\!-\!\alpha)z_k$. Finally, the agent $N\!+\!1$ triggers an event-based communication if the value $z_{k+1}$ is significantly different from the value that it last communicated to the agents $i=1,\dots,N$.

\begin{algorithm}[tb]
\caption{Event-Based Distributed Learning with Over-Relaxed ADMM }
\label{alg:over_relaxed_consensus}
\begin{algorithmic}
\REQUIRE Local objective functions $f^i$, parameters $\rho$, $\Delta^d, \Delta^z$, reset period $T$
\REQUIRE Initialize $\hat{x}^i_0=x_0$, $\hat{z}_0=\zeta_0=x_0$, $\hat{u}^i_{-1}=u_0^i$
\FOR{$k=0$ to $t_{\mathrm{max}}$}
\FOR{$i=1$ to $N$} %{Each agent}
\hspace*{-\fboxsep}{\colorbox{YellowGreen}{\parbox{\linewidth}{\STATE $\hat{z}_k^i \gets$ receive $z_{k}\!-\!z_{[k-1]}$ \hfill \COMMENT{Agent $i$}
\STATE $u^i_{k}= u^i_{k-1}+ \alpha x^i_{k}\!-\!\hat{z}_{k}^i +(1-\alpha)\hat{z}_{k-1}^i$ 
\STATE $x^i_{k+1} = \arg \min_{x^i} f^i(x^i)+\frac{\rho}{2}|x^i-\hat{z}_k^i+u^i_k|^2$
\STATE event-based send of $d^i_{k+1}-d^i_{[k]}$ \hfill \COMMENT{See \eqref{eqn:event_definition}}}}}\ENDFOR
    
\hspace*{-\fboxsep}{\colorbox{Apricot}{\parbox{\linewidth}{\STATE $\hat{\zeta}_k \gets$ receive  $\frac{1}{N}\sum_{i\in \Cd_{k+1}} (d^i_{k+1}\!-\!d^i_{[k]})$ \hfill \COMMENT{Agent $N\!+\!1$}
\STATE $z_{k+1}= \arg\min_z g(z) + \frac{N\rho}{2} |z-\hat{\zeta}_k{-}(1-\alpha)z_k|^2$
\STATE event-based send of $z_{k+1}\!-\!z_{[k]}$}}}

\IF{$\mathrm{mod}(k+1,T)=0$}
\STATE perform reset $\rightarrow$ $\hat{\zeta}_k=\zeta_k$, $\hat{z}_k=z_k$
\ENDIF
\ENDFOR 
\end{algorithmic}
\end{algorithm}

Next, we explain the details of the event-based communication protocol on the example of the communication of $d^i$, which is related to the primal $x^i$ and dual $u^i$ variables of agent $i$. The other event-based communications proceed similarly. The protocol comes in two variants, \textbf{vanilla event-based} and \textbf{randomized event-based}.  

\textbf{Vanilla event-based:} This communication rule is inspired from the sent-on-delta concept \citep{miskowicz_send--delta_2006}, which aims to reduce the number of communications by only sending updates when significant changes occur. A communication is triggered, if the value $d_{k+1}^i$ has deviated by more than the predefined threshold $\Delta^d\!>\!0$ compared to the value that was last communicated. We introduce the variable $d_{[k]}^i$ to denote the value $d_{k}^i$ that was last communicated and add the index $i$ to the set $\Cd_{k+1}$. The set $\Cd_{k+1}$ denotes the set of agents that trigger a communication of $d^i_{k+1}$ at time-step $k$, that is, \looseness-1
\begin{align}|d^i_{k+1}\!-\!d^i_{[k]}|\!>\!\Delta^{d}\!\iff\!i\!\in\!\Cd_{k+1},\label{eqn:event_definition}\end{align} 
and $d^i_{k+1}-d^i_{[k]}$ is sent out. Similarly, agent $N+1$ triggers a communication if $|z_{k+1}-z_{[k]}|>\Delta^z$. We also model communication failures as drops, which we represent by the variables $\chi_{k+1}^{di}$. The variable $\chi_{k+1}^{di}$ takes the value $\chi^{di}_{k+1}=-(d^i_{k+1}\!-\!d^i_{[k]})$, if $d^i_{k+1}\!-\!d^i_{[k]}$ is not received by the agent $N\!+\!1$; otherwise $\chi^{di}_{k+1}=0$. The agent $N\!+\!1$ updates its estimate of the average $\zeta_k$ according to the primal and dual variables that it has received at time $k$, that is,
\begin{align*}
\begin{aligned}
\hat{\zeta}_{k}=\hat{\zeta}_{k-1}&+\frac{1}{N}\sum_{i\in \Cd_{k+1}} \left(d^i_{k+1}-d_{[k]}^i + \chi_{k+1}^{di}\right).
\end{aligned}%\label{eqn:zeta_update}
\end{align*}
\textbf{Randomized event-based:} The protocol makes a case distinction. If $| d^i_{k+1}\!-\!d^i_{[k]}|  \leq  \Delta^{d}$, a communication is randomly triggered with probability $p_\text{trig}$. If $| d^i_{k+1}\!-\!d^i_{[k]}|  > \Delta^{d}$, a communication is triggered with certainty. Randomized communication from agent $N+1$ to the other agents works in a similar way. If $| z_{k+1}\!-\!z_{[k]}|  \leq  \Delta^{z}$, then a communication is randomly triggered with probability $p_\text{trig}$ between agent $N+1$ and agent $i$.

We observed in our numerical experiments that \textbf{randomized event-based} often improves \textbf{vanilla event-based} in terms of the achieved communication versus solution accuracy trade-off.

The error caused by the event-based communication remains bounded at all times thanks to the communication protocol and the periodic resets. This is summarized with the next proposition, whose proof is included in App.~\ref{app:ProofProp2}:
\begin{proposition}\label{prop:boundederr}
The error $\hat{\zeta}_k\!-\!\zeta_k$ at iteration $k$ is bounded by $|\hat{\zeta}_k\!-\!\zeta_k |\! \leq\! \Delta^d\!+\! T \bar{\chi}^d$,
where $T$ denotes the reset period (see Alg.~\ref{alg:over_relaxed_consensus}) and $\bar{\chi}^d$ is a bound on the disturbance $\chi_k^{di}$. 
\end{proposition}
We now state the main convergence result for Alg.~\ref{alg:over_relaxed_consensus}. The result arises as a corollary from the convergence analysis in the more general distributed optimization setting (see Thm.~\ref{thm:symbolic_rate}). We also provide sublinear convergence rates in a nonconvex setting (see Thm~\ref{prop:nonconvex_result}).

\begin{corollary} \label{corr:learning_convex} Let $f=\sum_{i=1}^N f^i$ be $m$-strongly convex and $L$-smooth with $\kappa=L/m$, and $g$ be convex. Let the step-size be $\rho=(mL)^{\frac{1}{2}}\kappa^\epsilon$ with $\epsilon \in [0, \infty)$, and  $\alpha=1$. For large enough $\kappa$, we have
\begin{align*}
|z_{k}-z_\ast|^2 \leq 4\left(\!1\!-\!\frac{1 }{4\,\kappa ^{\epsilon\! +\frac{1}{2}}}\!\right)^{2k}\! D_0 \!+\! \frac{5}{N}\kappa^{2+2\epsilon}\Delta^2,
\end{align*} 
where $z_\ast$ is the optimal value for the consensus variable $z$, and $D_0$ represents the initial error, $D_0=|z_{0}\!-\!z_\ast|^2 +\frac{1}{N}\!\sum_{i=1}^N\! |u^i_{0}\!-\!u^i_\ast|^2$, with $u^i_\ast$ denoting the optimal values of the dual variables associated with each agent. Here, $\Delta=N\Delta^{d}\!+\!\Delta^{z}\!+\!T(N\Bar{\chi}^{d}\!+\!\Bar{\chi}^{z})$ captures the error arising from the event-based communication. 
\end{corollary}

The convergence result bounds the distance between the consensus variable $z_k$ and the optimal solution $z_\ast=x_\ast$ that minimizes \eqref{eqn:DL_problem}. The analysis models our event-based learning algorithm as a dynamical system, accounting for disturbances introduced by the event-based communication strategy. By design, these disturbances remain bounded under the communication protocol. The next section elaborates on the formulation of our algorithm as a dynamical system. 

The strong convexity assumption enables faster convergence rates compared to more general nonconvex scenarios. Specifically, under this assumption, the rate of convergence is linear, as shown in Cor.~\ref{corr:learning_convex} and accelerated. In contrast, without such assumptions, convergence rates are generally much slower, typically sublinear or achieving only asymptotic convergence. 

We note that the strong convexity assumption in Cor.~\ref{corr:learning_convex} only requires $f:=\sum_{i=1}^Nf^i$ to be strongly convex, without imposing the same condition on the individual components $f^i$. In addition, we present a convergence result for general nonconvex cases in Thm.~\ref{prop:nonconvex_result} leading to sublinear convergence rates. The proof is provided in App.~\ref{app:nonconvex}.

\begin{theorem} \label{prop:nonconvex_result} 
Let each $f^i:\mathbb{R}^n\rightarrow \mathbb{R}$ be smooth (potentially nonconvex) and let $g:\mathbb{R}^n \rightarrow \bar{\mathbb{R}}$ be a proper, closed convex function. Let the relaxation parameter be $\alpha=1$, and the communication threshold $\Delta_k$ decay as $\Delta_k={\Delta_0}/{(k+1)^2}$. Then, the gradients and residuals converge with a rate of $\mathcal{O}(1 / k)$, and the following bound holds: %
\begin{align*} \frac{1}{K+1}\!\sum_{k=0}^K\!\Bigg(\!\frac{2}{3N} \sum_{i=1}^N \! \big|r^i_{k+1}\big|^2 +\! \frac{1}{6 N} \Big| G_{k+1}\Big|^2 \!\Bigg)=\mathcal{O}\!\left(\frac{{1}}{K}\right), \end{align*} where $r^i_{k+1}=x^i_{k+1}- z_{k+1}$ are the residuals, and the gradient terms are given by 
\begin{align*}
    G_{k+1}\!\in\!\frac{1}{ \rho N}\left(\sum_{i=1}^N \!\nabla f^i(x^i_{k+1}) + \partial g(z_{k+1})\right).
\end{align*}
\end{theorem}

\section{Event-Based ADMM as a Dynamical System}\label{sec:ADMM_dynamics}
We introduce a more general problem formulation that encompasses the previous section as a special case in order to broaden the scope of our analysis. This leads to the following constrained minimization problem
\begin{align}
\underset{x\in \mathbb{R}^p,~z\in \mathbb{R}^q}{\min} f(x)+g(z), \quad  \text { subject to }~  A x+B z=c,\label{eqn:main_problem}
\end{align}
where $x\!\in\! \mathbb{R}^p$ and $z\! \in \!\mathbb{R}^q$ are decision variables, $A\! \in \!\mathbb{R}^{r \times p}, B\! \in\! \mathbb{R}^{r \times q}$, and $c\! \in\! \mathbb{R}^r$ are corresponding matrices, and the objective function is decomposed into a smooth part $f\!:\!\mathbb{R}^p\!\rightarrow\!\mathbb{R}$ and a nonsmooth part $g\!:\!\mathbb{R}^q\!\rightarrow\!\bar{\mathbb{R}}$. We will provide an analysis under the following standard assumptions in distributed optimization.
\begin{assumption}\label{ass:matrix}
    The matrix $A$ is invertible and $B$ is full column rank.
\end{assumption}
\begin{assumption}
\label{ass:func_f}\label{ass:func_g}
The function $f$ is $m$-strongly convex and $L$-smooth. The function $g$ is convex.
\end{assumption}

{It is important to emphasize that the assumption of strong convexity for $f$  is introduced to derive linear rates within a dynamical systems framework. This assumption does not limit the practical applicability of the algorithm and a corresponding nonconvex result is included in Thm.~\ref{prop:nonconvex_result}. We also note that Assumption~\ref{ass:func_g} allows for nonconvex $f^i$ (see \eqref{eqn:DL_problem}) that $\sum f^i$ is strongly convex.}

The formulation in \eqref{eqn:main_problem} accommodates a variety of distributed optimization problems, including consensus, resource sharing, and distributed model fitting, see for example \citep{Boyd_2010}. App.~\ref{app:comm} further highlights how the general formulation can be tailored and simplified to accommodate specific applications, such as the sharing problem or finding a consensus on a graph. 

Our event-based distributed learning method is summarized in Alg.~\ref{alg:relaxed_event}. The algorithm is based on an over-relaxed version of ADMM, where an event-based communication structure between different agents is introduced. The over-relaxation brings the additional parameter $\alpha$, which, as we will show, can be used to achieve faster convergence rates. The communication structure of the algorithm is shown in Fig.~\ref{fig:Communication_inpaper} and includes three agents that keep track of the individual quantities $r_k=A x_k$, $s_k= B z_k$, and the dual multiplier $u_k$. In the special case of the consensus problem, the updates of the primal variable $x_k$ and dual variable $u_k$ decompose further into local updates based on $x_k^i$ and $u_k^i$, which results in the communication structure shown in Fig.~\ref{fig:dist_lr_server}. 

Alg.~\ref{alg:relaxed_event} begins by initializing its variables, and over a series of iterations, agents alternate between sharing information and optimizing their local variables. Key steps include updating variables based on local objectives and residuals, and triggering communication events when individual residual changes exceed predefined thresholds. The algorithm leverages event-based communication to reduce the communication load, while still achieving convergence towards an optimal solution of \eqref{eqn:main_problem}, as we will show in the following section. The event-based communication proceeds as in Sec.~\ref{sec:problem}, that is, the $r$-agent, for example, triggers a communication with the other agents if $| r_{k+1}\!-\! r_{[k]}|\! >\! \Delta^r$, at which point it sends the difference $r_{k+1}\!-\! r_{[k]}$ to the other agents. We again model communication failures by introducing the variables $\chi^{ru}_{k+1}$ if the communication is not received by the $u$-agent at time $k\!+\!1$. The notation is analogous for the remaining agents and communication lines (see also Fig.~\ref{fig:Communication_inpaper}).

\begin{algorithm}
\caption{Event-Based Distributed Optimization with Over-Relaxed ADMM}
\label{alg:relaxed_event}
\begin{algorithmic}
\REQUIRE Functions $f$ and $g$, matrices $A$ and $B$, vector $c$, parameters $\rho$ and $\alpha$. Initial condition $x_0,z_0$
\STATE $r_0\!=\!\hat{r}^s_{0}\!=\!\hat{r}^u_{0}\!=\!Ax_0,$$s_0\!=\!\hat{s}^r_{0}\!=\!\hat{s}^u_{0}\!=\!Bz_0,$$u_0\!=\!\hat{u}^r_{0}\!=\!\hat{u}^s_{0}\!=\!0$
\FOR{$k=0$ to $t_{\mathrm{max}}$}
 \hspace*{-\fboxsep}{\colorbox{YellowGreen}{\parbox{\linewidth}{ \STATE $\hat{s}_k^r, \hat{u}_k^r \gets$ receive $s_{k+1}\!-\!s_{[k]},u_{k+1}\!-\!u_{[k]}$
    \STATE $ x_{k+1}=\arg\!\min_x f(x)+\frac{\rho}{2}|Ax+\hat{s}^r_k-c+\hat{u}^r_k|^2$  
    \STATE event-based send  $r_{k+1}\!-\!r_{[k]}$ where $r_{k+1}=A x_{k+1}$}}}
    \hspace*{-\fboxsep}{\colorbox{Lavender}{\parbox{\linewidth}{ \STATE $\hat{r}_{k+1}^s, \!\hat{u}_k^s\gets\!$ receive $r_{k+1}\!-\!r_{[k]},u_{k+1}\!-\!u_{[k]}$
    \STATE $ z_{k+1}\!\!=\!\!\underset{z}{\arg\!\min} g(z)\!+\!\!\frac{\rho}{2}|\alpha \hat{r}^s_{k+1}\!\!\!-\!(1\!-\!\alpha) Bz_k\!+\! Bz\!-\!\alpha c\!+\!\hat{u}^s_k |^2$ 
    \STATE event-based send $s_{k+1}\!-\!s_{[k]}$, where $s_{k+1}=B z_{k+1}$}}}
\hspace*{-\fboxsep}{\colorbox{GreenYellow}{\parbox{\linewidth}{\STATE $\hat{r}^u_{k+1},\!\hat{s}^u_{k+1}\!\gets\!$ receive $r_{k+1}-\!r_{[k]},s_{k+1}-\!s_{[k]}$
\STATE $u_{k+1}=u_k+\alpha \hat{r}^u_{k+1}\!-\!(1-\alpha) \hat{s}^u_k+\hat{s}^u_{k+1}\!-\!\alpha c$  
\STATE event-based send $u_{k+1}\!-\!u_{[k]}$}}}
    \IF{$\mathrm{mod}(k+1,T)=0$}
    \STATE reset $\rightarrow$ $\hat{r}^{u;s}_{k+1}=r_{k+1}, \hat{s}_{k+1}^{u;r}=s_{k+1}$, $u_{k+1}^{r;s}=u_{k+1}$
    \ENDIF
\ENDFOR
\end{algorithmic}
\end{algorithm}

Alg.~\ref{alg:relaxed_event} has three update steps that occur sequentially, whereby the first two involve optimization problems that can be replaced by their corresponding stationarity conditions. This yields the following implicit update equations:
\begin{align*}
\begin{aligned}
     &0=\nabla f(x_{k+1})+\rho A^\top ( Ax_{k+1}+\hat{s}^r_k-c+\hat{u}^r_k)\\
     &0\in\! \partial g(\!z_{k+1}\!)\!+\!\rho B^\top\! ( \alpha \hat{r}^s_{k+1}\!\!-\!(1\!-\!\alpha) B\!z_k\!+\! Bz_{k+1}\!-\!\!\alpha c\!+\!\hat{u}^s_k ) \\
     &u_{k+1}=u_k+\alpha \hat{r}^u_{k+1}-(1-\alpha) \hat{s}^u_k+\hat{s}^u_{k+1}-\alpha c,\end{aligned}%\label{eqn:variable_update}
\end{align*}
which can be expressed by the dynamical system shown in Fig.~\ref{fig:dynSys}. We note that the variable $x_{k+1}$ is uniquely determined by $\hat{s}^r_k$ and $\hat{u}_k^r$ and does not depend on $x_k$, which means that only $\xi_k\!:=\!(s_k,u_k)$ comprises the state of the dynamical system. We further note that the dynamical system includes a nonlinear component, which arises from the (sub)gradient evaluations $\nabla f$ and $\partial g$, and the system is subjected to external disturbances $e_k$ that arise from the event-based communication. The detailed derivation and the corresponding matrices for the dynamics in Fig.~\ref{fig:dynSys} are included in App.~\ref{app:dynamics}. Our convergence analysis will build on the dynamical systems model of Alg.~\ref{alg:relaxed_event}. While our analysis is inspired by earlier works, such as \citep{Nishihara_2015} and \citep{Lessard_2016}, the Lyapunov function that is used to prove convergence rates are different due to the external disturbances caused by the event-based communication.

 \begin{figure}[h]
   \begin{minipage}{0.5\textwidth}
  \centering
  \tikzstyle{block} = [draw, rectangle, minimum height=2em, minimum width=2em]
\tikzstyle{smallblock} = [draw, rectangle, minimum height=1em, minimum width=1em]
\tikzstyle{longblock} = [draw, rectangle, minimum height=5em, minimum width=1em]

\tikzstyle{sum} = [draw, circle, node distance=2cm]
\tikzstyle{input} = [coordinate]

\tikzstyle{output} = [coordinate]
\tikzstyle{pinstyle} = [pin edge={to-,thin,black}]

\begin{tikzpicture}[auto, node distance=2cm,>=latex',every node/.style={font=\small}]
    % Blocks

    \node [block, fill={Lavender}] (aux) {$s$};  
    \node [block,  right=5cm of $(aux)$, fill={GreenYellow}] (dual) {$u$};
    \node [block,  below=1cm of $(aux)!0.5!(dual)$, fill=YellowGreen] (x) {$r$};

    \draw [->,dashed] (aux) --  (dual) node[midway, below]{$s_{k+1}-s_{[k]}$};
    \draw [->,dashed] ([xshift=-0.2cm]aux.south) --  (x.west) node[midway, below, sloped]{$s_{k+1}-s_{[k]}$};

    \draw [->,dashed] (x) --  ([xshift=0.2cm]aux.south) node[midway, above,sloped]{$r_{k+1}-r_{[k]}$};
    
    \draw [->,dashed] (x.east) --  ([xshift=0.2cm]dual.south) node[midway, below,sloped]{$r_{k+1}-r_{[k]}$};

    \draw [->,dashed] ([yshift=0.25cm]dual.west) --  ([yshift=0.25cm]aux.east) node[midway, above]{$u_{k+1}-u_{[k]}$};
    \draw [->,dashed] ([xshift=-0.2cm]dual.south) --  (x) node[midway, above,sloped]{$u_{k+1}-u_{[k]}$};

\end{tikzpicture}  \\ \vspace{0.3cm} \tikzstyle{block} = [draw, rectangle, minimum height=2.5em, minimum width=2.5em]
\tikzstyle{largeblock} = [draw, rectangle, minimum height=1em, minimum width=5em]
\tikzstyle{smallblock} = [draw, rectangle, minimum height=0.5em, minimum width=0.5em]
\tikzstyle{longblock} = [draw, rectangle, minimum height=5em, minimum width=1em]

\tikzstyle{sum} = [draw, circle, node distance=2cm]
\tikzstyle{input} = [coordinate]
\tikzstyle{elbow} = [coordinate]

\tikzstyle{pinstyle} = [pin edge={to-,thin,black}]

\begin{tikzpicture}[every node/.style={font=\small}]

% Blocks
\node [input](V) {};
%\node [sum, right=1cm of $(input)$] (sum) {$+$};
\node [block, right= 0.5 cm of $(V)$] (linear) {$\begin{array}{c|cc}
    \hat{A} & \hat{E}&\hat{B} \\   \hline
       \hat{C} & \hat{E}_y&\hat{D}  
\end{array}$};

\node [smallblock, below= 0.7 cm of $(linear)$] (nonlinear) {$\phi$};

\node [input, right= 1.8 cm of $(linear)$] (Error) {$e$};

\draw [->] (Error) -- (linear.east);

\node[right] at (Error) {$e$};

% Connections
\draw [-] (V) --  (linear);
\draw [<-] ([yshift=-0.25cm]linear.east) -| + (0.5,0) |- (nonlinear);

\draw [<-] (nonlinear) -| node[near end, left] {\scriptsize$\left[\begin{array}{c}
     r_{k+1}-c \\ s_{k+1} \end{array}\right]$} (V);

\node [smallblock, above= 0.8 cm of $(linear)$] (lag) {Delay};

\draw [->](lag) -| + (1.75,-0.75)  |- node[near start, right, yshift=0.4cm] {\scriptsize$\left[\begin{array}{c}
     s_{k} \\ u_{k} \end{array}\right]$} ([yshift=0.25cm]linear.east);

\draw [<-](lag) -- + (-1.5,0.0)  |- node[midway, left, yshift=0.5cm] {\scriptsize$\left[\begin{array}{c}
     s_{k+1} \\ u_{k+1} \end{array}\right]$} ([yshift=0.25cm]linear.west);

\end{tikzpicture}
     \end{minipage}
    \caption{The figure visualizes the event-based communication structure of Alg.~\ref{alg:relaxed_event} at the top and a discrete-time dynamical system which represents the sequence generated by the event-based ADMM algorithm on the bottom. The function $\phi$ is nonlinear and represents the evaluation of (sub)gradients.}
    \label{fig:Communication_inpaper} \label{fig:dynSys}
\end{figure}

\section{Convergence Analysis}\label{sec:convergence}\label{sec:symbolic_rates}

This section provides convergence guarantees for the event-based learning algorithm (Alg.~\ref{alg:relaxed_event}). The detailed proof for Thm.~\ref{thm:symbolic_rate} is provided in App.~\ref{app:convergence}.

\begin{theorem}\label{thm:symbolic_rate}
    Let Assumption~\ref{ass:matrix} and \ref{ass:func_g} be satisfied and let the step-size for Alg.~\ref{alg:relaxed_event} be $\rho\!=\!\kappa^\epsilon\! \sqrt{mL}\!/\!(\underline{\sigma}(\!A) \bar{\sigma}(\!A))$, for some $\epsilon\!\geq\!0$ and $\alpha\!\in\! (0.675,1\!+\!\sqrt{1\!-\!1/\sqrt{\kappa}})$, $\kappa\!=\!L\bar{\sigma}^2(A)/(m\underaccent{\bar}{\sigma}^2(A))$, where $\underaccent{\bar}{\sigma}$ and $\bar{\sigma}$ denote the minimum and maximum singular value of a matrix, respectively. Then, for large enough $\kappa$, the following bound holds: 
\begin{align*}
         |\xi_{k}\!-\!\xi_\ast|^2 \!\leq\! {\kappa_P} |\xi_{0}\!-\!\xi_\ast|^2\!\left(1\!-\!\frac{\alpha }{4\,\kappa ^{\epsilon +\frac{1}{2}}}\right)^{2k} \!\!\! + \!\frac{60\kappa^{2+2\epsilon}}{\alpha(1\!-\!|\alpha\!-\!1|)} \Delta^2,
\end{align*}
with $\xi_k=(s_k,u_k)$, and where $s_k=Bz_k$, $u_k$ is the dual variable, and $\xi_\ast$ the optimizer corresponding to \eqref{eqn:main_problem}. Furthermore, $\Delta\!=\!\Delta^{r}\!+\!\Delta^{s}\!+\!\Delta^{u}\!+\!T(\Bar{\chi}^r\!+\Bar{\chi}^s\!+\Bar{\chi}^u)$ represents the error arising from the event-based communication and 
$\kappa_P\!=\!{(2\sqrt{\kappa}\!-\!1\!+\!\sqrt{4\kappa(\alpha\!-\!1)^2\!+\!1})}/{(2\sqrt{\kappa}\!-\!1\!-\!\sqrt{4\kappa(\alpha\!-\!1)^2\!+\!1})}$. \end{theorem}

We conclude the section by highlighting a few important points. 

\textbf{(i)} For $\epsilon\!=\!0$, $\alpha\!=\!1$, the bound is considerably simplified to 
\begin{align*}
|\xi_{k}-\xi_\ast|^2 \leq 2 |\xi_{0}-\xi_\ast|^2\left(1-\frac{1}{4\sqrt{\kappa}}\right)^{2k} + 60\kappa^{2} \Delta^2,
\end{align*}
which shows that the convergence rate scales with $1\!/\!\sqrt{\kappa}$ and is therefore accelerated. This also highlights that the same convergence rate (up to constants) can be achieved with the event-based learning algorithm stated in Alg.~\ref{alg:over_relaxed_consensus} compared to a standard ADMM algorithm. As we will show in the numerical experiments, our event-based algorithm reduces communication without any significant reduction in accuracy.

\textbf{(ii)} The bound from Thm.~\ref{thm:symbolic_rate} also highlights how the communication thresholds $\Delta$ affect the solution accuracy. In the simplified scenario with $\epsilon=0$, $\alpha=1$ (the more general scenario follows the same rationale), the solution accuracy is bounded by $|\xi_{k}\!-\!\xi_\ast| \leq 8\kappa \Delta$, for large enough $k$. This means that the solution accuracy of Alg.~\ref{alg:relaxed_event} is proportional to the condition number $\kappa$ and $\Delta$.

\textbf{(iii)} We can therefore easily ensure convergence, by choosing a time-varying $\Delta=\Delta_k$ such that $\Delta_k\rightarrow 0$. The formal statement is included and derived in App.~\ref{app:dec_delta}. We also obtain precise nonasymptotic bounds. For example, if $\Delta_k=\Delta_0/(k+1)^t$ for any $t>0$, we conclude that the error converges with $\mathcal{O}(1/k^t)$ (see again App.~\ref{app:dec_delta}).

\textbf{(iv)} If $f$ fails to be strongly convex, we can include a small regularizer, for example of the type $m |x|^2/2$. Choosing a diminishing regularizer with $m=\mathcal{O}(1/k^2)$ and a diminishing threshold $\Delta_k=\mathcal{O}(1/k^4)$ can be shown to result in an accelerated convergence rate of $\mathcal{O}(1/k^2)$.

\textbf{(v)} The topology of the communication network, represented by the matrix $A$, directly influences the convergence rate, through the condition number $\kappa=L\bar{\sigma}^2(A)/(m\underaccent{\bar}{\sigma}^2(A))$. This formulation allows us to generalize our convergence results beyond simple client-server architectures. See App.~\ref{app:graph} for a detailed discussion on how agent network topology is encoded in the matrix $A$. 

\section{Numerical Experiments}\label{sec:results}

This section discusses the performance of Alg.~\ref{alg:over_relaxed_consensus} in numerical experiments, highlighting that Alg.~\ref{alg:over_relaxed_consensus} achieves fast convergence while reducing communication. Numerical experiments with the more general version (Alg.~\ref{alg:relaxed_event}) are included in App.~\ref{app:add_experiments}, where distributed training over a network of agents is explored. We also investigate the trade-off between communication load and solution accuracy achieved by selecting different communication thresholds. The communication load is calculated by counting the number of triggered communications for $T_{\mathrm{max}}$ number of steps and normalizing according to the full communication case of one data package per round.

\begin{figure}[htb]
    \centering
    \definecolor{mycolor1}{rgb}{0.00000,0.44700,0.74100}%
\definecolor{mycolor2}{rgb}{0.85000,0.32500,0.09800}%
\definecolor{mycolor3}{rgb}{0.92900,0.69400,0.12500}%
%\definecolor{mycolor4}{rgb}{0.49400,0.18400,0.55600}%
\definecolor{mycolor4}{rgb}{0.00000, 0.60000, 0.40000}% Adjusted to a teal green for better contrast

\definecolor{mycolor5}{rgb}{0.46600,0.67400,0.18800}%
\definecolor{mycolor6}{rgb}{0.30100,0.74500,0.93300}%
\definecolor{mycolor7}{rgb}{0.63500,0.07800,0.18400}%
\pgfplotsset{every tick label/.append style={font=\small}}
\pgfplotsset{every axis label/.append style={font=\small}}
\pgfplotsset{every axis plot/.append style={line width=1.0pt}} % Set the line width for the whole axis
\pgfplotsset{every mark plot/.append style={scale=0.05}}

\begin{tikzpicture}

\begin{axis}[
width=8.2cm, 
height=5.1cm,%6.2cm,
axis background/.style={fill=white},
axis line style={black},
legend cell align={left},
legend style={font=\small, anchor=south east, nodes={scale=0.7, transform shape}, 
at={(1.0,0.0)},
  fill opacity=0.8,
  draw opacity=1,
  text opacity=1,
  fill=white
},
legend columns=1,
x grid style={gray!50},y grid style={gray!50},
xmajorgrids,xminorgrids,
xmin=-1, xmax=150,
ymajorgrids,yminorgrids,
ymin=0.08, ymax=0.9,
tick align=inside,
xtick = {0,50,100,149},
xticklabels = {0,50,100,150},
ytick = {0.1,0.2,0.3,0.4,0.5,0.6,0.7,0.80,0.9},
yticklabels = {10,20,30,40,50,60,70,80,90},
ytick pos=left,
xtick pos=bottom,
xlabel={\textcolor{black}{Round}}, % Use standard color
ylabel={\textcolor{black}{Validation Accuracy \%}},%ymode=log
xlabel near ticks
]

\addplot[thick, mycolor1] table[x=step,y=FedEvent_Vanilla_val_acc_avg] {fedevent_data.dat};\addlegendentry{Alg.~\ref{alg:over_relaxed_consensus}, $(\Delta=1.75)$};
\addplot [name path=upper,draw=none,forget plot] table[x=step,y expr=\thisrow{FedEvent_Vanilla_val_acc_avg}+\thisrow{FedEvent_Vanilla_val_acc_std}] {fedevent_data.dat};
\addplot [name path=lower,draw=none,forget plot] table[x=step,y expr=\thisrow{FedEvent_Vanilla_val_acc_avg}-\thisrow{FedEvent_Vanilla_val_acc_std},forget plot] {fedevent_data.dat};
\addplot [fill=mycolor1!20, forget plot] fill between[of=upper and lower];

\addplot[thick, mycolor4] table[x=step,y=FedEvent_Rand_val_acc_avg] {fedevent_data.dat};\addlegendentry{Alg.~\ref{alg:over_relaxed_consensus}-Rand $(\Delta=3.75,p_{\mathrm{trig}}=0.7)$};
\addplot [name path=upper,draw=none,forget plot] table[x=step,y expr=\thisrow{FedEvent_Rand_val_acc_avg}+\thisrow{FedEvent_Rand_val_acc_std}] {fedevent_data.dat};
\addplot [name path=lower,draw=none,forget plot] table[x=step,y expr=\thisrow{FedEvent_Rand_val_acc_avg}-\thisrow{FedEvent_Rand_val_acc_std},forget plot] {fedevent_data.dat};
\addplot [fill=mycolor4!20, forget plot] fill between[of=upper and lower];

\addplot[thick, mycolor3] table[x=step,y=FedAvg_val_acc_avg] {data.dat};\addlegendentry{FedAvg - $part\_rate=0.7$};
\addplot [name path=upper,draw=none,forget plot] table[x=step,y expr=\thisrow{FedAvg_val_acc_avg}+\thisrow{FedAvg_val_acc_std}] {data.dat};
\addplot [name path=lower,draw=none,forget plot] table[x=step,y expr=\thisrow{FedAvg_val_acc_avg}-\thisrow{FedAvg_val_acc_std}] {data.dat};
\addplot [fill=mycolor3!20,forget plot] fill between[of=upper and lower];

\addplot[thick, mycolor7] table[x=step,y=FedProx_val_acc_avg] {data.dat};\addlegendentry{FedProx - $part\_rate=0.7$};
\addplot [name path=upper,draw=none,forget plot] table[x=step,y expr=\thisrow{FedProx_val_acc_avg}+\thisrow{FedProx_val_acc_std}] {data.dat};
\addplot [name path=lower,draw=none,forget plot] table[x=step,y expr=\thisrow{FedProx_val_acc_avg}-\thisrow{FedProx_val_acc_std}] {data.dat};
\addplot [fill=mycolor7!20,forget plot] fill between[of=upper and lower];

\addplot[thick, mycolor2] table[x=step,y=FedADMM_val_acc_avg] {data.dat};\addlegendentry{FedADMM - $part\_rate=0.9$};
\addplot [name path=upper,draw=none,forget plot] table[x=step,y expr=\thisrow{FedADMM_val_acc_avg}+\thisrow{FedADMM_val_acc_std}] {data.dat};
\addplot [name path=lower,draw=none,forget plot] table[x=step,y expr=\thisrow{FedADMM_val_acc_avg}-\thisrow{FedADMM_val_acc_std}] {data.dat};
\addplot [fill=mycolor2!20,forget plot] fill between[of=upper and lower];

\addplot[thick, mycolor5] table[x=step,y=SCAFFOLD_val_acc_avg] {data.dat};\addlegendentry{SCAFFOLD - $part\_rate=0.7$};
\addplot [name path=upper,draw=none,forget plot] table[x=step,y expr=\thisrow{SCAFFOLD_val_acc_avg}+\thisrow{SCAFFOLD_val_acc_std}] {data.dat};
\addplot [name path=lower,draw=none,forget plot] table[x=step,y expr=\thisrow{SCAFFOLD_val_acc_avg}-\thisrow{SCAFFOLD_val_acc_std}] {data.dat};
\addplot [fill=mycolor5!20,forget plot] fill between[of=upper and lower];

% Add labels inside the plot
\node at (110, 0.82) {\scriptsize{Alg.\ref{alg:over_relaxed_consensus}, Alg.\ref{alg:over_relaxed_consensus}-Rand, FedADMM}}; % Label at (20, 145)

\node at (110, 0.63) {\scriptsize{FedAvg, FedProx, SCAFFOLD}}; % Label at (140, 145)

\end{axis}

\end{tikzpicture}
\definecolor{mycolor1}{rgb}{0.00000,0.44700,0.74100}%
\definecolor{mycolor2}{rgb}{0.85000,0.32500,0.09800}%
\definecolor{mycolor3}{rgb}{0.92900,0.69400,0.12500}%
%\definecolor{mycolor4}{rgb}{0.49400,0.18400,0.55600}%
\definecolor{mycolor4}{rgb}{0.00000, 0.60000, 0.40000}% Adjusted to a teal green for better contrast
\definecolor{mycolor5}{rgb}{0.46600,0.67400,0.18800}%
\definecolor{mycolor6}{rgb}{0.30100,0.74500,0.93300}%
\definecolor{mycolor7}{rgb}{0.63500,0.07800,0.18400}%
\pgfplotsset{every tick label/.append style={font=\small}}
\pgfplotsset{every axis label/.append style={font=\small}}
\pgfplotsset{every axis plot/.append style={line width=1.0pt}} % Set the line width for the whole axis
\pgfplotsset{every mark plot/.append style={scale=0.05}}

\begin{tikzpicture}

\begin{axis}[
width=8.2cm, 
height=4.7cm,
axis background/.style={fill=white},
axis line style={black},
legend cell align={left},
legend style={font=\small, anchor=south east, nodes={scale=0.7, transform shape}, 
at={(1.0,0.0)},
  fill opacity=0.8,
  draw opacity=1,
  text opacity=1,
  fill=white
},
legend columns=1,
x grid style={gray!50},y grid style={gray!50},
xmajorgrids,xminorgrids,
xmin=-1, xmax=150,
ymajorgrids,yminorgrids,
ymin=0.0, ymax=200,
tick align=inside,
xtick = {0,50,100,149},
xticklabels = {0,50,100,150},
ytick = {0,40,80,120,160,200},
yticklabels = {0,20,40,60,80,100},
ytick pos=left,
xtick pos=bottom,
xlabel={\textcolor{black}{Round}}, % Use standard color
ylabel={\textcolor{black}{Communication Load \%}},
]
\addplot[smooth, semithick, mycolor1] table[x=step,y=FedEvent_Vanilla_total_run_load_avg_smooth] {fedevent_data.dat};\addlegendentry{Alg.~\ref{alg:over_relaxed_consensus}  $(\Delta=1.75)$};
\addplot [name path=upper,draw=none,forget plot] table[x=step,y expr=\thisrow{FedEvent_Vanilla_total_run_load_avg_smooth}+\thisrow{FedEvent_Vanilla_total_run_load_std_smooth}] {fedevent_data.dat};
\addplot [name path=lower,draw=none,forget plot] table[x=step,y expr=\thisrow{FedEvent_Vanilla_total_run_load_avg_smooth}-\thisrow{FedEvent_Vanilla_total_run_load_std_smooth},forget plot] {fedevent_data.dat};
\addplot [fill=mycolor1!20, forget plot] fill between[of=upper and lower];

\addplot[smooth, semithick, mycolor4] table[x=step,y=FedEvent_Rand_total_run_load_avg_smooth] {fedevent_data.dat};\addlegendentry{Alg.~\ref{alg:over_relaxed_consensus}-Rand  $(\Delta=3.75,p_{\mathrm{trig}}=0.7)$};
\addplot [name path=upper,draw=none,forget plot] table[x=step,y expr=\thisrow{FedEvent_Rand_total_run_load_avg_smooth}+\thisrow{FedEvent_Rand_total_run_load_std_smooth}] {fedevent_data.dat};
\addplot [name path=lower,draw=none,forget plot] table[x=step,y expr=\thisrow{FedEvent_Rand_total_run_load_avg_smooth}-\thisrow{FedEvent_Rand_total_run_load_std_smooth},forget plot] {fedevent_data.dat};
\addplot [fill=mycolor4!20, forget plot] fill between[of=upper and lower];

%Horizontal line at y = 0.7
\addplot[mycolor3, thick] coordinates {(0,140) (150,140)};       \addlegendentry{FedAvg/FedProx/SCAFFOLD - $part\_rate=0.7$}

%Horizontal line at y = 0.9
\addplot[mycolor2, thick] coordinates {(0,180) (150,180)};        \addlegendentry{FedADMM - $part\_rate=0.9$}

\end{axis}

\end{tikzpicture}
    \caption[Accuracy and Communication Load Trends]{Validation accuracy (top) and communication load percentage (bottom) over 150 communication rounds for training a CIFAR-10 classifier. The results indicate that Alg.~\ref{alg:over_relaxed_consensus} achieves top accuracy at a lower communication rate. The plots compare the performance of various algorithms, including Alg.~\ref{alg:over_relaxed_consensus} with different parameter settings (Vanilla and randomized), FedAvg, FedProx, FedADMM, and SCAFFOLD. Notably, ADMM-based methods (Alg.~\ref{alg:over_relaxed_consensus}, Alg.~\ref{alg:over_relaxed_consensus}-Rand and FedADMM) demonstrate better convergence by reaching up to $78\%$ test accuracy, compared to other algorithms FedAvg, FedProx and SCAFFOLD, which reach only $70\%$ accuracy. Among ADMM-based methods, Alg.~\ref{alg:over_relaxed_consensus} and  Alg.~\ref{alg:over_relaxed_consensus}-Rand achieve the same accuracy with over 20\% less communication load.
 Communication load curves are smoothed using a window length of three for visualization purposes.}
 \vspace{0.5cm}
    \label{fig:time_cifar}
\end{figure}
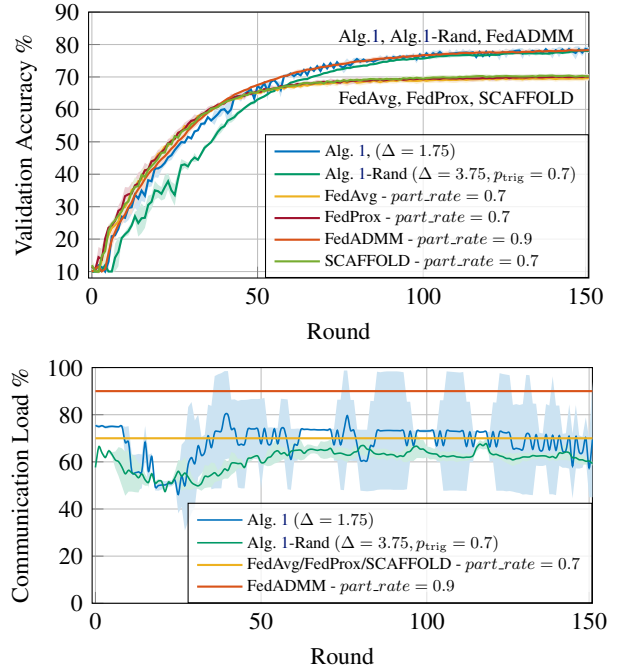

\begin{table*}[htb]
    \centering
    \begin{tabular}{c|p{1cm}p{1cm}p{1cm}|p{1cm}p{1cm}p{1cm}p{1cm}}
        \toprule
        \multirow{2}{*}{\textbf{Algorithm}} & \multicolumn{3}{c}{\textbf{MNIST Target Accuracy}} 
        & \multicolumn{4}{c}{\textbf{CIFAR-10 Target Accuracy}} \\
        \cmidrule{2-8}
        & \textbf{80\%} & \textbf{85\%} & \textbf{90\%}  
        & \textbf{70\%} & \textbf{75\%} & \textbf{77\%} & \textbf{78\%} \\
        \midrule
        \textbf{Alg.~\ref{alg:over_relaxed_consensus} - Randomized} 
        & \textbf{629} & \textbf{693} & \underline{1723} 
        & 12531 %0.2 4.5
        & \textbf{13422} & \underline{15008} %0.2	3.5
 & \textbf{18376} \\
        \textbf{Alg.~\ref{alg:over_relaxed_consensus} - Vanilla} 
        & 816 & {1285} & \textbf{1710}  
        & {12214} & \underline{14780} %0	3.25
        & \textbf{14780} & \underline{20690}\\
        \textbf{FedADMM} \citep{Zhou_Li_2023} 
        & \underline{800} & \underline{1200} & $>$2000 
        & {12000} & 15000 & 21000 & 27000 \\
        \textbf{FedAvg} \citep{mcmahan_communicationefficient_2017}
        & \underline{800} & 2000 & N/A  
        & \textbf{3000} & N/A & N/A & N/A \\
        \textbf{FedProx} \citep{li_federated_2020}
        & 1000 & 2000 & N/A  
        & \underline{6000} & N/A & N/A & N/A \\
        \textbf{SCAFFOLD} \citep{Karimireddy_SCAFFOLD_2020}
        & 1600  & 2000 & 3200  
        & 12000 & N/A & N/A & N/A \\
        \bottomrule
    \end{tabular}
    \caption[Load]{The total number of communication events required by each algorithm to achieve the target accuracies for the MNIST and CIFAR-10 classifiers within 100 and 150 rounds, respectively. “N/A” indicates cases where the target accuracy was not reached within the specified rounds. Parameter choices for the algorithms are detailed in Appendix~\ref{app:add_experiments}. Reported values are averages over multiple experiments with different random seeds, with standard deviations below 2\%, making them negligible. The corresponding communication load and accuracy trends for the rightmost column of CIFAR-10 are shown in Fig.~\ref{fig:time_cifar}. The results emphasize the trade-off between achieving higher validation accuracy and maintaining communication efficiency across various algorithm configurations.}
    \label{tab:num_events} 
\end{table*}

Due to space limitations, we present two examples in this section. Further experiments (including linear regression, LASSO, and distributed training over a network of agents) are presented in App.~\ref{app:add_experiments}. App.~\ref{app:add_experiments} also includes hyperparameters for the experiments, model details and discusses the effect of communication failures.

We start by evaluating the performance of Alg.~\ref{alg:over_relaxed_consensus} on MNIST \citep{deng2012mnist} and CIFAR-10 \citep{cifar10}. Tab.~\ref{tab:num_events} reports the total number of communication events required by each algorithm to achieve the target accuracies for the MNIST and CIFAR-10 classifiers. Our event-based algorithm consistently requires fewer communication events to achieve high accuracies compared to baseline methods. This reduction is attributed to the selective triggering mechanism, which prevents unnecessary communication while ensuring convergence. For instance, on the CIFAR-10 dataset, our approach achieved 78\% accuracy with a cost of 18,376 communication events, compared to 27,000 for FedADMM.

The comparison with other federated learning methods emphasizes the challenges associated with non-i.i.d. data distribution and communication overhead. FedAvg, as highlighted in \citep{Li_Yang_2020,glasgow2022sharp}, experiences slowdowns in the presence of non-i.i.d. data, and increasing participation does not necessarily alleviate this issue. FedProx has the same issue and is unable to converge to a classifier that generalizes across all digits. FedADMM and SCAFFOLD can indeed cope with non-i.i.d. data, in general, both achieving high classification accuracies. However, FedADMM has disadvantages arising from the random sampling mechanism and SCAFFOLD suffers from an additional communication load to communicate two variables (client drift and local model). Notably, all baselines employ a random selection of agents, which, in non-i.i.d. scenarios, misses crucial changes and results in a waste of communication resources. Our method addresses these challenges by adopting an event-based agent selection approach and outperforms all baselines by yielding uniformly better trade-off curves.

\section{Conclusion}\label{sec:concl}
We introduce an event-based distributed learning approach that effectively reduces communication overhead by triggering events only when local models undergo significant changes. The method, based on over-relaxed ADMM, exhibits accelerated convergence rates in convex settings, demonstrates robustness to communication failures, and outperforms common baselines such as FedAvg, FedProx, SCAFFOLD and FedADMM in our experiments, which include an MNIST and CIFAR-10 learning task. The experiments highlight that savings of more than 35\% are possible without significantly degrading the solution accuracy (less than $1$\%). Our method allows for explicit trade-offs between communication load and solution accuracy, making it promising for large-scale learning systems with heterogeneous data and communication constraints.

\textbf{Limitations:} While our approach offers significant improvements in communication efficiency, it has not yet accounted for adversarial attacks, such as gradient poisoning, or unreliable nodes in the network. These factors could potentially degrade the robustness of the method. {Nevertheless, our event-based methodology can be integrated with robust aggregation or anomaly detection methods \cite{Pillutla_robust_2023,Yin_bzyn_20218} to improve security without compromising communication efficiency.} Additionally, this method has not been specifically analyzed with respect to differential privacy and does not address privacy concerns in the current formulation. 

\textbf{Discussion:} In this article, we focused on communication efficiency and convergence relationship under some constraints. However, practical implementations often face additional challenges, such as limited bandwidth, high latency, and network failures. Although we have not explicitly named these factors, they can be effectively modeled through our packet drop framework, which provides a foundation for handling such constraints. 

Furthermore, while we initially framed our algorithm as a synchronous method, event-based communication can also be adapted to function in asynchronous systems. This flexibility allows the method to accommodate real-world scenarios with varying degrees of network reliability and synchronization. 

Finally, our method is compatible with compression and quantization techniques, which can further reduce the size of the models exchanged during communication events and improve communication efficiency. This can also help minimize the amount of data stored in the agents, contributing to more efficient memory usage and reducing the overall communication load.

\bibliography{sample}
\bibliographystyle{arxiv_bib}

\onecolumn

\newpage

\appendix
\onecolumn

\newpage
\section{Communication Structure}\label{app:comm}
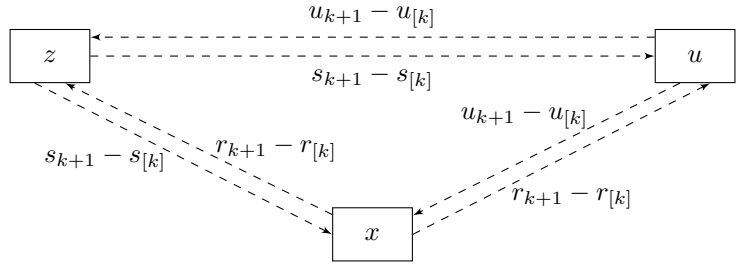
\begin{figure}[H]
    \centering
    \tikzstyle{block} = [draw, rectangle, minimum height=2em, minimum width=3em]
\tikzstyle{largeblock} = [draw, rectangle, minimum height=1em, minimum width=5em]
\tikzstyle{smallblock} = [draw, rectangle, minimum height=1em, minimum width=1em]
\tikzstyle{longblock} = [draw, rectangle, minimum height=5em, minimum width=1em]

\tikzstyle{sum} = [draw, circle, node distance=2cm]
\tikzstyle{input} = [coordinate]

\tikzstyle{output} = [coordinate]
\tikzstyle{pinstyle} = [pin edge={to-,thin,black}]

\begin{tikzpicture}[auto, node distance=2cm,>=latex']
    % Blocks

    \node [block] (aux) {$z$};  
    \node [block,  right=8cm of $(aux)$] (dual) {$u$};
    \node [block,  below=2cm of $(aux)!0.5!(dual)$] (x) {$x$};

    \draw [->,dashed] (aux) --  (dual) node[midway, below]{$s_{k+1}-s_{[k]}$};
    \draw [->,dashed] ([xshift=-0.2cm]aux.south) --  (x.west) node[midway, left, xshift=-0.1cm]{$s_{k+1}-s_{[k]}$};

    \draw [->,dashed] (x) --  ([xshift=0.2cm]aux.south) node[midway, right, xshift=0.1cm]{$r_{k+1}-r_{[k]}$};
    
    \draw [->,dashed] (x.east) --  ([xshift=0.2cm]dual.south) node[near start, right, xshift=0.2cm]{$r_{k+1}-r_{[k]}$};

    \draw [->,dashed] ([yshift=0.25cm]dual.west) --  ([yshift=0.25cm]aux.east) node[midway, above]{$u_{k+1}-u_{[k]}$};
    
    \draw [->,dashed] ([xshift=-0.2cm]dual.south) --  ([yshift=0.25cm]x.east) node[near start, left, xshift=-0.2cm]{$u_{k+1}-u_{[k]}$};

\end{tikzpicture}
    \caption{The communication structure that arises from Alg.~\ref{alg:relaxed_event}, where $s:=Bz$, $r:=Ax$, and $u$ denotes the dual variable.}
    \label{fig:general}
\end{figure}

This section discusses the sharing problem and consensus reaching over graphs as two special cases of the more general constrained minimization problem 
\begin{align}
\underset{x\in \mathbb{R}^p,~z\in \mathbb{R}^q}{\min} f(x)+g(z), \qquad  \text { subject to }~  A x+B z=c,\label{eqn:app_general_form}
\end{align}
with variables $x \in \mathbb{R}^p$ and $z \in \mathbb{R}^q$ and constant matrices $A \in \mathbb{R}^{r \times p}, B \in \mathbb{R}^{r \times q}$, and $c \in \mathbb{R}^r$. The objective function is decomposed into a smooth part $f:\mathbb{R}^p\rightarrow\mathbb{R}$ and nonsmooth part $g:\mathbb{R}^q\rightarrow\mathbb{R}$. The communication structure of the problem formulation \eqref{eqn:app_general_form} is shown in Fig.~\ref{fig:general}, where primal, dual and auxiliary variables are treated as different communication nodes.

\subsection{Sharing Problem}

We will show that the event-based communication structure introduced in Fig.~\ref{fig:general} simplifies considerably for the sharing problem. The sharing problem takes the following form,
\begin{align*}
    \min_{x^1, \ldots, x^N \in \mathbb{R}^p}& \quad \sum_{i=1}^N~f^i(x^i)+g\left(\sum_{i=1}^N x^i\right),
\end{align*}
and arises as a special case from \eqref{eqn:app_general_form} when choosing $f(x)=\sum_{i=1}^N~f^i(x^i)$, $x=(x^1,x^2,\dots,x^N)\in \mathbb{R}^{Np}$, $A=I_{Np}$, $B=-(I_p,I_p,\dots,I_p)$, $c=0$. The problem can be solved via the following updates, by agents $i=1,\ldots,N$:
\begin{align}
\begin{aligned}
& x^i_{k+1}=\underset{x^i \in \mathbb{R}^p}{\operatorname{argmin}} f^i\left(x^i\right)+\frac{\rho}{2}\left|x^i-x^i_k+\hat{h}_{k}\right|^2, \end{aligned}\end{align}
and by agent $N+1$:
\begin{align}
\begin{aligned}
&\bar{x}_{k+1} = \frac{1}{N}\sum_{i=1}^N \hat{x}^i_{k+1}\\
& z^{k+1}=\underset{z \in \mathbb{R}^p}{\operatorname{argmin}}~g(N z)+\frac{N \rho}{2}\left|{z}-\bar{x}_{k+1}-\frac{1}{\rho} {u}^k\right|^2 \\
& u_{k+1}=u_k+\rho\left(\bar{x}_{k+1}-{z}_{k+1}\right) \\
& h_{k+1}= \bar{x}_{k+1}-z_{k+1}+\frac{1}{\rho} u_{k+1}.
\end{aligned}
\end{align}

\begin{figure}[H]
    \centering
   \tikzstyle{block} = [draw, rectangle, minimum height=2em, minimum width=3em]
\tikzstyle{largeblock} = [draw, rectangle, minimum height=1em, minimum width=5em]

\tikzstyle{block} = [draw, rectangle, minimum height=2em, minimum width=2em]
\tikzstyle{thickblock} = [draw, rectangle, minimum height=1em, minimum width=1em, thick]
\tikzstyle{smallblock} = [draw, rectangle, minimum height=1em, minimum width=1em]
\tikzstyle{longblock} = [draw, rectangle, minimum height=1em, minimum width=30em, thick]
\tikzstyle{input} = [coordinate]

\tikzstyle{output} = [coordinate]
\tikzstyle{pinstyle} = [pin edge={to-,thin,black}]

\begin{tikzpicture}[auto, node distance=2cm,>=latex',every node/.style={font=\small}]
    % Blocks
    \node [input, name=input1] {};
    \node [input, name=input2, below of=input1] {};
     % Summation
    \node [block, right of=input1, node distance=1cm] (A1) {Agent 1};

    \node [block, right of=A1, node distance=3cm] (A2) {Agent 2};
    
     \node [block, right of=A2, node distance=3cm] (A3) {Agent 3};
       \node [block, right of=A3, node distance=3cm] (A4) {Agent 4};

    \node [longblock,  below=1.2cm of $(A2)!0.5!(A3)$] (bus) {Agent 5};

    \draw [->,dashed] (A1.south) -- ([xshift=-4.5cm]bus.north)  node[near start, left, font=\tiny]{$x^1_{k+1}-x^1_{[k]}$};

    \draw [->,dashed] (A2.south) -- ([xshift=-1.5cm]bus.north) node[near start, left, font=\tiny]{$x^2_{k+1}-x^2_{[k]}$};
    \draw [->,dashed]  (A3.south)  -- ([xshift=1.5cm]bus.north) node[near start, left, font=\tiny]{$x^3_{k+1}-x^3_{[k]}$} ;
    \draw [->,dashed] (A4.south) --  ([xshift=4.5cm]bus.north) node[near start, left, font=\tiny]{$x^4_{k+1}-x^4_{[k]}$};
    \draw [->,dashed] ([xshift=-4.2cm]bus.north) -- ([xshift=0.3cm]A1.south) node[near start, right, font=\tiny]{$h_{k+1}-h_{[k]}$ };
    \draw [->,dashed] ([xshift=-1.2cm]bus.north) -- ([xshift=0.3cm]A2.south) node[near start, right, font=\tiny]{$h_{k+1}-h_{[k]}$ };
    \draw [->,dashed] ([xshift=1.8cm]bus.north) -- ([xshift=0.3cm]A3.south) node[near start, right, font=\tiny]{$h_{k+1}-h_{[k]}$ };
    \draw [->,dashed] ([xshift=4.8cm]bus.north) --  ([xshift=0.3cm]A4.south) node[near start, right, font=\tiny]{$h_{k+1}-h_{[k]}$ };

\end{tikzpicture}
    \caption{The diagram visualizes the communication structure for the sharing problem for $N=4$ agents.}
    \label{fig:sharing}
\end{figure}
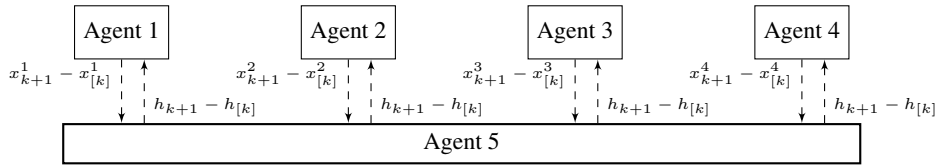

For the sharing problem, the general communication scheme in Fig.~\ref{fig:general} reduces to the diagram in Fig.~\ref{fig:sharing}, where each node communicates their local variable in an event-based manner.

\subsection{Consensus over a Graph}\label{app:graph}

As another example, we will show that \eqref{eqn:app_general_form} also generalizes to distributed learning scenarios over graphs. We consider a network topology, captured by an undirected connected graph $\mathcal{G}=(\mathcal{V}, \mathcal{E})$, where $\mathcal{V}=\{1, \ldots, N\}$ is the set of vertices and $\mathcal{E} \subseteq \mathcal{V} \times \mathcal{V}$ is the set of edges. Each agent (vertex) has a local data distribution, and the aim is to train a model without a central server to aggregate the collected information. The problem can be formulated as follows:
\begin{align*}
\underset{{x}^i\in \mathbb{R}^p, {z}^{i j}\in \mathbb{R}^p}{\min} ~ \sum_{i=1}^N f^i\left({x}^i\right), \qquad \text { subject to }~ & {x}^i={z}^{i j}, ~{x}^j={z}^{i j}, \quad \forall(i, j) \in \mathcal{E}.
\end{align*}

Similar to the formulation in \citep{Yu_Freris_2023_CE}, we define transmitter and receiver matrices $\hat{A}_t, \hat{A}_r \in$ $\mathbb{R}^{|\mathcal{E}| \times N}$ for all edges, i.e.,
\begin{align*}
\left[\hat{A}_t\right]_{e i}=\left[\hat{A}_r\right]_{e j}=\begin{cases}
    1 & (i, j) \in \mathcal{E} \\
    0 & \text{otherwise}
\end{cases}, \quad \forall e \in  \mathcal{E}. \end{align*}  
By stacking ${x}^i, {z}^{i j} \in \mathbb{R}^{p}$ into column vectors ${x} \in \mathbb{R}^{Np}, {z} \in \mathbb{R}^{|\mathcal{E}|p}$, respectively, we conclude that distributed learning over graphs is indeed a special case of \eqref{eqn:app_general_form},
\begin{align*}
\underset{x\in \mathbb{R}^{Np},~z\in \mathbb{R}^{|\mathcal{E}|p}}{\min} ~f(x), \qquad
\text { subject to } & \left[\begin{array}{c}
\hat{A}_t \otimes I_p \\
\hat{A}_r \otimes I_p
\end{array}\right] {x}=\left[\begin{array}{c}
I_{|\mathcal{E}| p} \\
I_{|\mathcal{E}| p}
\end{array}\right] {z},
\end{align*}
where $\otimes$ denotes the Kronecker product and $I_p$ the identity matrix. Thus, the matrices $A$ and $B$ encode the topology of the communication graph, which will affect the convergence rates as highlighted with our main result Thm.~\ref{thm:symbolic_rate} where the convergence rate is dictated by the value $\kappa=\bar{\sigma}(A)L/({\underaccent{\bar}{\sigma}(A)}{m})$.

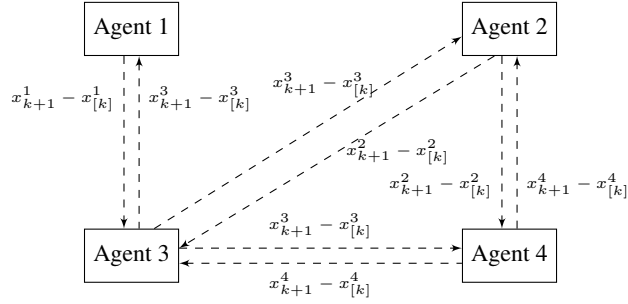
\begin{figure}
    \centering
    \tikzstyle{block} = [draw, rectangle, minimum height=2em, minimum width=3em]
\tikzstyle{largeblock} = [draw, rectangle, minimum height=1em, minimum width=5em]

\tikzstyle{block} = [draw, rectangle, minimum height=2em, minimum width=2em]
\tikzstyle{thickblock} = [draw, rectangle, minimum height=1em, minimum width=1em, thick]
\tikzstyle{smallblock} = [draw, rectangle, minimum height=1em, minimum width=1em]
\tikzstyle{longblock} = [draw, rectangle, minimum height=1em, minimum width=30em, thick]
\tikzstyle{input} = [coordinate]

\tikzstyle{output} = [coordinate]
\tikzstyle{pinstyle} = [pin edge={to-,thin,black}]

\begin{tikzpicture}[auto, node distance=2cm,>=latex',every node/.style={font=\small}]
  
    % Blocks
    \node [block, right of=input1] (A1) {Agent 1};
    \node [block, right of=A1, node distance=5cm] (A2) {Agent 2};
    \node [block, below of=A1, node distance=3cm] (A3) {Agent 3};
    \node [block, right of=A3, node distance=5cm] (A4) {Agent 4};

    % Arrows
    
    \draw [->,dashed] ([xshift=-0.1cm]A1.south) -- ([xshift=-0.1cm]A3.north)  node[near start, left, font=\tiny]{$x^1_{k+1}-x^1_{[k]}$};
    \draw [<-,dashed] ([xshift=0.1cm]A1.south) -- ([xshift=0.1cm]A3.north)  node[near start, right, font=\tiny]{$x^3_{k+1}-x^3_{[k]}$};
   
    \draw [->,dashed] ([yshift=0.1cm]A3.east) -- ([yshift=0.1cm]A4.west) node[midway, above, font=\tiny]{$x^3_{k+1}-x^3_{[k]}$} ;
    \draw [<-,dashed] ([yshift=-0.1cm]A3.east) -- ([yshift=-0.1cm]A4.west) node[midway, below, font=\tiny]{$x^4_{k+1}-x^4_{[k]}$} ;
    
    \draw [->,dashed] ([xshift=0.1cm]A4.north) -- ([xshift=0.1cm]A2.south) node[near start, right, font=\tiny]{$x^4_{k+1}-x^4_{[k]}$};
    \draw [->,dashed] ([xshift=-0.1cm]A2.south) -- ([xshift=-0.1cm]A4.north) node[near end, left, font=\tiny]{$x^2_{k+1}-x^2_{[k]}$};

    \draw [->,dashed] ([xshift=-0.2cm]A2.south) -- ([yshift=0.1cm]A3.east) node[midway, right, font=\tiny]{$x^2_{k+1}-x^2_{[k]}$};
    \draw [->,dashed] ([xshift=0.3cm]A3.north) -- ([yshift=-0.1cm]A2.west) node[near end, left, font=\tiny]{$x^3_{k+1}-x^3_{[k]}$};

\end{tikzpicture}
    \caption{The diagram visualizes the communication structure for a distributed learning problem over a graph that connects four agents with four edges. }
    \label{fig:graph_structure}
\end{figure}

The resulting instance of Alg.~\ref{alg:over_relaxed_consensus} takes the following form:
\begin{align}\begin{aligned}
x^i_{k+1} & =\underset{x^i \in \mathbb{R}^p}{\operatorname{argmin}} f_i\left(x_i\right)+\frac{|\mathcal{N}_i| \rho}{2}\left|x^i-\frac{1}{2}\left(x_i^k-\bar{x}^i_k\right)+\frac{1}{\rho} p^i_k\right|^2\\
\bar{x}^i_{k+1} & =\frac{1}{|\mathcal{N}_i|} \sum_{j \in \mathcal{N}_i} \hat{x}^j_{k+1} \\
p^i_{k+1} & =p^i_k+\frac{\rho}{2}\left(x^i_{k+1}-\bar{x}^i_{k+1}\right), 
\end{aligned}
\end{align}
where $\mathcal{N}_i$ represents the set containing the neighbors of the agent $i$ and $|\mathcal{N}_i|$ is the number of vertices. In the event based-communication setting, an agent transmits its local model ($x^i_{k+1}$) to the neighbors only if there has been a significant change in the local model. Fig.~\ref{fig:graph_structure} shows an example with four agents, each communicating local variables only.

\newpage
\section{Proof of Thm.~\ref{prop:nonconvex_result} }\label{app:nonconvex}
\begin{proof}
We will establish the convergence rate by analyzing the behavior of a carefully chosen Lyapunov function. Let us begin by formulating the augmented Lagrangian for \eqref{eqn:DL_problem}:
\begin{align}
    \mathcal{L}_\rho(x, z, y) = \sum_{i=1}^N f^i(x^i) + g(z) + \sum_{i=1}^N (y^{i})^\top (x^i - z) + \frac{\rho}{2} \sum_{i=1}^N |x^i - z|_2^2,\label{eqn:aug_lag}
\end{align}
where $x=(x^1,\ldots,x^N)$ and $y=(y^1,\ldots,y^N)$ represent the primal and dual variables respectively, and $\rho > 0$ is the penalty parameter.

We express the ADMM updates in Alg.~\ref{alg:over_relaxed_consensus} for $\alpha=1$ as follows by introducing the scaled dual variable $u^i=\frac{1}{\rho}y^i$:
\begin{align}
x^i_{k+1} &= \arg \min_{x^i} \left( f^i(x^i) + \frac{\rho }{2} |x^i - \hat{z}_k + u^i_k|_2^2 \right), \quad &\forall i \in \{1, \ldots, N\},\label{eqn:x_up}\\
 z_{k+1} &= \arg \min_z \left( g(z) + \frac{N\rho}{2} \sum_{i=1}^N |\hat{x}^i_{k+1} - z + \hat{u}^i_k|_2^2 \right),\label{eqn:z_up}\\
u^i_{k+1} &= u^i_k + x^i_{k+1} - \hat{z}_{k+1}, \quad &\forall i \in \{1, \ldots, N\}.\label{eqn:u_up}
\end{align}Here, we use the notation $\hat{z}_{k+1}={z}_{k+1}+\varepsilon^z_{k+1}$, $\hat{x}^i_{k+1}={x}^i_{k+1}+\varepsilon^{x,i}_{k+1}$, and $\hat{u}^i_{k+1}={u}^i_{k+1}+\varepsilon^{u,i}_{k+1}$ to account for errors emerging from event-based communication.

From \eqref{eqn:x_up} and \eqref{eqn:z_up}, we derive the following first-order optimality conditions for $ x^i_{k+1} $ and  $ z_{k+1} $
\begin{align}
    0 = &\nabla f^i(x^i_{k+1}) + \rho (x^i_{k+1} - \hat{z}_k + u^i_k)\label{eqn:f_optimality}\\
    0 \in &\partial g(z_{k+1}) + \rho  \sum_{i=1}^N (z_{k+1} - \hat{x}^i_{k+1} - \hat{u}^i_k).\label{eqn:g_optimality}
\end{align}
We then define the Lyapunov function:
\begin{align}
    V_k = |z_k - z_\ast|_2^2 + \frac{1}{N} \sum_{i=1}^N |u^i_k - u^i_\ast|_2^2 ,\label{eqn:lyap_non}
\end{align}
where $(u^i_\ast, z_\ast)$ denotes the optimal dual and consensus variables. Our goal is to demonstrate that this Lyapunov function is monotonically decreasing.

The optimality condition in \eqref{eqn:f_optimality} implies that $x^i_{k+1}$ minimizes,
\begin{align*}
f^i(x)+\rho\left(u^i_{k+1}+z_{k+1}-z_k\right)^\top   x +\rho (\varepsilon_{k+1}^z-\varepsilon^z_{k})^\top  x.
\end{align*}

From this minimization, we can derive the following inequality,
\begin{align}\begin{aligned}
    f^i(x_{k+1}^i)-f^i(x^i_\ast)\leq  \rho\left(u^i_{k+1}+z_{k+1}-z_k\right)^\top  (x^i_\ast-x^i_{k+1}) +\rho (\varepsilon_{k+1}^z-\varepsilon^z_{k})^\top  (x^i_\ast-x^i_{k+1}).\end{aligned}\label{eqn:f_diff}
\end{align}

Similarly, the optimality condition \eqref{eqn:g_optimality} indicates that $z_{k+1}$ minimizes,\
\begin{align}
    g\left(z\right)-\rho\sum_{i=1}^N( u^i_{k+1} + \varepsilon^{d,i}_{k+1} +\varepsilon^z_{k+1})^\top z,\label{eqn:g_min}
\end{align}
where  $\varepsilon^{d,i}_{k+1}=\varepsilon^{x,i}_{k+1} + \varepsilon^{u,i}_k.$  This minimization leads to,
\begin{align}
    g\left(z_{k+1}\right)-g\left(z_\ast\right) \leq\rho \sum_{i=1}^N ( u^i_{k+1})^\top (z_{k+1}-z_\ast) - \rho\sum_{i=1}^N(\varepsilon^{d,i}_{k+1} +\varepsilon^z_{k+1})^\top (z_{k+1}-z_\ast). \label{eqn:g_diff}
\end{align}

By adding  \eqref{eqn:g_diff} and the sum over $i$ of  \eqref{eqn:f_diff}, and applying the conditions $x^i_\ast-z_\ast=0$ along with the relation $x^i_\ast-x^i_{k+1}=-r^i_{k+1}-(z_{k+1}-z_\ast)$, we obtain,
\begin{align}
\begin{aligned}
g(z_{k+1})\!-\!g(z_\ast)\!+\!\sum_{i=1}^N \!\left(f^i(x^i_{k+1})\!-\!f^i(x^i_\ast)\right)\! \leq&\!-\!\rho\!\sum_{i=1}^N\! \left(u^i_{k+1}\right)^\top  r^i_{k+1}\!-\!\rho\!\sum_{i=1}^N \! \left(z_{k+1}-z_k\right)^\top   (r^i_{k+1}+(z_{k+1}-z_\ast)) \\\quad&\!+\!\rho\!\sum_{i=1}^N\!  (\varepsilon_{k+1}^z-\varepsilon^z_{k})^\top  (x^i_\ast-x^i_{k+1})\!+\! \rho\!\sum_{i=1}^N \! (\varepsilon^{d,i}_{k+1} +\varepsilon^z_{k+1})^\top (z_{k+1}\!-\!z_\ast),\end{aligned}\label{eqn:inter1}
\end{align} where the residual is defined as $r^i_k:=x^i_{k}-z_k$.

Since $(x^i_\ast,z_\ast,u_\ast^i)$ is a saddle point of $\mathcal{L}_0$ in \eqref{eqn:aug_lag}, i.e., $  \mathcal{L}_0(x_\ast,z_\ast,u_\ast)\leq  \mathcal{L}_0(x_{k+1},z_{k+1},u_\ast)$, we have,
\begin{align}
    -g(z_{k+1})+g(z_\ast)-\sum_{i=1}^N \left(f^i(x^i_{k+1})-f^i(x^i_\ast)\right)\leq \sum_{i=1}^N {u_\ast^i}^\top (x^i_{k+1}-z_{k+1})=\sum_{i=1}^N {u^i_\ast}^\top r^i_{k+1}.\label{eqn:saddle1}
\end{align}

By adding \eqref{eqn:inter1} and \eqref{eqn:saddle1} and multiplying by $\frac{2}{\rho}$, we arrive at 
\begin{align}
\begin{aligned}
0 \geq& \underbrace{2\sum_{i=1}^N \left(u^i_{k+1}-u^i_\ast\right)^\top  r^i_{k+1}}_{\mathrm{(I)}}+\underbrace{2\sum_{i=1}^N  \left(z_{k+1}-z_k\right)^\top   (r^i_{k+1}+(z_{k+1}-z_\ast))}_{\mathrm{(II)}} \\&\qquad-2\sum_{i=1}^N  (\varepsilon_{k+1}^z-\varepsilon^z_{k})^\top  (x^i_\ast-x^i_{k+1}) - 2\sum_{i=1}^N  (\varepsilon^{d,i}_{k+1} {+}\varepsilon^z_{k+1})^\top (z_{k+1}-z_\ast).\end{aligned}\label{eqn:inter2}
\end{align}

Here $\left(u^i_{k+1}-u^i_\ast\right)^\top  r^i_{k+1}$ can be written as $\left(u^i_{k+1}-u^i_\ast\right)^\top (u^i_{k+1}-u^i_k+\varepsilon^z_{k+1})$ which splits into \begin{align*}\left(u^i_{k+1}-u^i_\ast\right)^\top  r^i_{k+1}&=\frac{1}{2}\left(u^i_{k+1}-u^i_k\right)^\top\varepsilon^z_{k+1}+ \frac{1}{2}|u^i_{k+1}-u^i_k|^2+\frac{1}{2}\left(u^i_{k+1}-u^i_k\right)^\top (u^i_{k+1}-u^i_k+\varepsilon^z_{k+1})\\&\qquad+\left(u^i_{k}-u^i_\ast\right)^\top (u^i_{k+1}-u^i_\ast)+\left(u^i_{k}-u^i_\ast\right)^\top\varepsilon^z_{k+1}- |u^i_{k}-u^i_\ast|^2.\end{align*}

Next, we substitute $u_{k+1}^i-u_k^i=r_{k+1}^i-\varepsilon^z_{k+1}$
and use the following squared norm identity,
\begin{align*}
     \frac{1}{2}|u^i_{k+1}-u^i_k|^2+\left(u^i_{k}-u^i_\ast\right)^\top (u^i_{k+1}-u^i_\ast)= \frac{1}{2}|u^i_{k+1}-u^i_\ast|^2+\frac{1}{2}|u^i_{k}-u^i_\ast|^2.
\end{align*}

Consequently, we can expand terms (I) and (II) as follows:
\begin{align*}
    \mathrm{(I)}&=\sum_{i=1}^N \left(|u^i_{k+1}-u^i_\ast|^2-|u^i_{k}-u^i_\ast|^2 +2(u^i_{k+1}-u^i_\ast)^\top\varepsilon^z_{k+1} +|\varepsilon_{k+1}^z|^2+|r^i_{k+1}|^2-2(\varepsilon^z_{k+1})^\top r^i_{k+1}    \right)\\
    \mathrm{(II)}&=\sum_{i=1}^N  \left(|r^i_{k+1}+\left(z_{k+1}-z_k\right)|^2+ |z_{k+1}-z_\ast|^2 - |z_{k}-z_\ast|^2 -|r^i_{k+1}|^2 \right).
\end{align*}

Substituting these expansions back into \eqref{eqn:inter2} and using the definition of our Lyapunov function from \eqref{eqn:lyap_non}, we can express the decrease in the Lyapunov function as,

\begin{align*}
\begin{aligned}
0 \geq &N (V_{k+1}-V_k)+\sum_{i=1}^N \left(2(u^i_{k+1}-u^i_\ast)^\top\varepsilon^z_{k+1} +|\varepsilon_{k+1}^z|^2-2(\varepsilon^z_{k+1})^\top r^i_{k+1}    \right)\\\quad&+\sum_{i=1}^N  \left(|r^i_{k+1}+\left(z_{k+1}-z_k\right)|^2 \right) +2\sum_{i=1}^N  (\varepsilon_{k+1}^z-\varepsilon^z_{k})^\top  (x^i_{k+1}-x^i_\ast) - 2\sum_{i=1}^N  (\varepsilon^{d,i}_{k+1} +\varepsilon^z_{k+1})^\top (z_{k+1}-z_\ast).\end{aligned}
\end{align*} 

Modifying the error terms using $x^i_{k+1}-x_\ast^i=r^i_{k+1}+(z_{k+1}-z_\ast)$ and $r^i_{k+1}=u^i_{k+1}-u^i_k+\varepsilon^z_{k+1}$, we get,
%\di{fit! (latex work)}
\begin{align}
\begin{aligned}
N (V_{k+1}-V_k)\leq -\sum_{i=1}^N  |r^i_{k+1}&+\left(z_{k+1}-z_k\right)|^2 + \sum_{i=1}^N \big(-2 (\varepsilon^z_{k+1}-\varepsilon^z_{k})^\top (u^i_{k+1}-u^i_\ast) - |\varepsilon^z_{k+1}|^2 \\&\quad+ 2 (\varepsilon^{d,i}_{k+1}+\varepsilon^z_{k})^\top (z_{k+1}-z_\ast) -2 (\varepsilon^z_{k})^\top  \left(u^i_{k} -u^i_\ast\right)-2 (\varepsilon^z_{k})^\top  \varepsilon^z_{k+1}\big). \end{aligned}\label{eqn:inter3}
\end{align} 

Furthermore, we expand the squared norm term:
\begin{align}
  -\sum_{i=1}^N|r^i_{k+1}+\left(z_{k+1}-z_k\right)|^2=  -\sum_{i=1}^N|r^i_{k+1}|^2 -2 \sum_{i=1}^N(r^i_{k+1})^\top \left(z_{k+1}-z_k\right) -\sum_{i=1}^N|z_{k+1}-z_k|^2.\label{eqn:sqr_bound}
\end{align}

Next we will establish a bound for the cross term  $-2 \sum_{i=1}^N (r^i_{k+1})^\top  \left(z_{k+1}-z_k\right) $.

We recall that \eqref{eqn:g_optimality} implies \eqref{eqn:g_min}. This minimization property leads to the following pair of inequalities:
\begin{align*}
    g\left(z_{k+1}\right)- g\left(z_k\right)&\leq \rho\sum_{i=1}^N( u^i_{k+1} + \varepsilon^{d,i}_{k+1} +\varepsilon^z_{k+1})^\top (z_{k+1}-z_k)\\
    g\left(z_{k}\right)- g\left(z_{k+1}\right)&\leq -\rho\sum_{i=1}^N( u^i_{k} + \varepsilon^{d,i}_{k} +\varepsilon^z_{k})^\top (z_{k+1}-z_k).
\end{align*}

By adding these inequalities and rearranging terms, we obtain,
\begin{align*}
    0\leq \sum_{i=1}^N( u^i_{k+1}-u^i_k + \varepsilon^{d,i}_{k+1}-\varepsilon^{d,i}_{k} +\varepsilon^z_{k+1}-\varepsilon^z_{k})^\top (z_{k+1}-z_k)  =&\sum_{i=1}^N( r^i_{k+1}+ \varepsilon^{d,i}_{k+1}-\varepsilon^{d,i}_{k} -\varepsilon^z_{k})^\top (z_{k+1}-z_k).\end{align*}
   This yields the desired bound on the cross term $-2 \sum_{i=1}^N (r^i_{k+1})^\top  \left(z_{k+1}-z_k\right)$ which takes the form \begin{align}
  -2\sum_{i=1}^N( r^i_{k+1})^\top (z_{k+1}-z_k)\leq&2\sum_{i=1}^N(\varepsilon^{d,i}_{k+1}-\varepsilon^{d,i}_{k} -\varepsilon^z_{k})^\top (z_{k+1}-z_k).\label{eqn:cr_bound}
\end{align}

As the next step, we rewrite $|z_{k+1} - z_k|$ in terms of the gradients of $f^i$ and $g$. This will be important for deriving the desired convergence result.
From the first-order optimality condition of the $ z $-update \eqref{eqn:g_optimality}, 

and rearranging for $ z_{k+1} $, we get,
\begin{align}
    z_{k+1} \in \frac{1}{N}\sum_{i=1}^N \left( x^i_{k+1} + u^i_k+\varepsilon^{d,i}_{k+1} \right) - \frac{1}{\rho N} \partial g(z_{k+1}).\label{eqn:z_kp1}
\end{align}

The optimality condition for $x^i_{k+1}$ (see \eqref{eqn:f_optimality}) results in,
\begin{align}
 - z_k   =\varepsilon^z_{k}+\frac{1}{N}\sum_{i=1}^N\left( -x^i_{k+1}  - u^i_k-\frac{1}{\rho} \nabla f^i(x^i_{k+1})\right).\label{eqn:z_k}
\end{align}

We combine \eqref{eqn:z_kp1} and \eqref{eqn:z_k} to express the update for $z_{k+1}-z_k$ as follows,
\begin{align*}
    z_{k+1} - z_k \in -\frac{1}{\rho N}\left( \sum_{i=1}^N \nabla f^i(x^i_{k+1}) + \partial g(z_{k+1})\right) + \varepsilon^z_{k}+\frac{1}{N}\sum_{i=1}^N \varepsilon^{d,i}_{k+1} .
\end{align*}

Thus, $ z_{k+1} - z_k $ depends on the averaged gradients $ \nabla f^i $ and $ \partial g $, scaled by the penalty parameter $ \rho $. We now take the square and apply Young's inequality on the cross term with $\gamma'=2$, which yields,
\begin{align}
    -|z_{k+1} - z_k|^2 \leq -\frac{1}{2} \left|\frac{1}{\rho N}\left( \sum_{i=1}^N \nabla f^i(x^i_{k+1}) + \grg_{k+1}\right)\right|^2 + \left|\varepsilon^z_{k}+\frac{1}{N}\sum_{i=1}^N \varepsilon^{d,i}_{k+1} \right|^2,\label{eqn:z_k_dif}
\end{align} for any $\grg_{k+1} \in  \partial g(z_{k+1})$.

By substituting \eqref{eqn:sqr_bound}, \eqref{eqn:cr_bound},  and \eqref{eqn:z_k_dif} in \eqref{eqn:inter3}, and applying Young's inequality to cross terms, we derive the following inequality:
\begin{align*}
\begin{aligned}
   N( V_{k+1}-V_k) \leq&-\sum_{i=1}^N  |r^i_{k+1}|^2-\frac{1}{2} \left|\frac{1}{\rho N}\left( \sum_{i=1}^N \nabla f^i(x^i_{k+1}) + \grg_{k+1}\right)\right|^2 + \left|\varepsilon^z_{k}+\frac{1}{N}\sum_{i=1}^N \varepsilon^{d,i}_{k+1} \right|^2 \\&\quad+\sum_{i=1}^N \Big( \gamma_4| \varepsilon^z_{k+1}-\varepsilon^z_{k}|^2+\frac{1}{\gamma_4}|r^i_{k+1}|^2+ \gamma_1 |\varepsilon^z_{k+1}|^2+\frac{1}{\gamma_1}|u^{k}-u^\ast|^2 - |\varepsilon^z_{k+1}|^2+\gamma_2|2\varepsilon^{d,i}_{k+1}  -  \varepsilon^{d,i}_k|^2\\&\qquad\qquad\quad+\frac{1}{\gamma_2}|z_{k+1}-z_k|^2 + \gamma_3|\varepsilon^{d,i}_{k+1}+\varepsilon^z_{k}|^2+\frac{1}{\gamma_3}|z_k-z_\ast|^2 +2 (\varepsilon^z_{k+1}-2\varepsilon^z_{k})^\top \varepsilon^z_{k+1} \Big),\end{aligned}
\end{align*}for any $\grg_{k+1} \in  \partial g(z_{k+1})$.

We can further simplify the expression by choosing the values as $\gamma_1=\gamma_3=\gamma_k$ and $\gamma_2=\gamma_4=3$,

\begin{align*}
\begin{aligned}
   N( V_{k+1}-V_k) \leq&\frac{N}{\gamma_k}V_k-\frac{2}{3}\sum_{i=1}^N  |r^i_{k+1}|^2-\frac{1}{6} \left|\frac{1}{\rho N}\left( \sum_{i=1}^N \nabla f^i(x^i_{k+1}) + \grg_{k+1}\right)\right|^2 + \left|\varepsilon^z_{k}+\frac{1}{N}\sum_{i=1}^N \varepsilon^{d,i}_{k+1} \right|^2 \\&\quad+\sum_{i=1}^N \Big( 3| \varepsilon^z_{k+1}-\varepsilon^z_{k}|^2+ \gamma_k |\varepsilon^z_{k+1}|^2 - |\varepsilon^z_{k+1}|^2+3|2\varepsilon^{d,i}_{k+1}  -  \varepsilon^{d,i}_k|^2 + \gamma_k|\varepsilon^{d,i}_{k+1}+\varepsilon^z_{k}|^2 \\&\qquad\qquad\quad+4|\varepsilon^z_{k+1}-2\varepsilon^z_{k}|^2+4|\varepsilon^z_{k+1}|^2 \Big),\end{aligned}
\end{align*}for any $\grg_{k+1} \in  \partial g(z_{k+1})$.

The error values arising from event-based communication are bounded by the communication thresholds, $|\varepsilon^{d,i}_k|\leq \Delta^d_k$, $|\varepsilon^{z}_k|\leq {\Delta^z_k}$. This leads to the following inequality,
\begin{align*}
\begin{aligned}
    V_{k+1} \leq\left(1+\frac{1}{\gamma_k}\right)V_k -\frac{2}{3N}\sum_{i=1}^N |r^i_{k+1}|^2 -\frac{1}{6N}& \left|\frac{1}{\rho N}\left( \sum_{i=1}^N \nabla f^i(x^i_{k+1}) + \grg_{k+1}\right)\right|^2\\&\qquad+ (3\gamma_k+51+\frac{8}{3N}){\Delta^z_k}^2+ (2\gamma_k+30+\frac{8}{3N}){\Delta^d_k}^2,\end{aligned}
\end{align*}for any $\grg_{k+1} \in  \partial g(z_{k+1})$, where $r_{k+1}^i=x_{k+1}^i-z_{k+1}$ represents the residuals at step $k+1$.
Simplifying the expression, we obtain:
\begin{align}
    V_{k+1} \leq \left(1 + \frac{1}{\gamma_k}\right) V_k  -\frac{2}{3N}\sum_{i=1}^N |x^i_{k+1}-z_{k+1}|^2 -\frac{1}{6N} \left|\frac{1}{\rho N}\left( \sum_{i=1}^N \nabla f^i(x^i_{k+1}) + \grg_{k+1}\right)\right|^2 +  \mathcal{O}(\gamma_k\cdot\Delta^2_k),
\end{align}for any $\grg_{k+1} \in  \partial g(z_{k+1})$,
where $ \Delta_k$ represents the time-varying communication threshold, i.e., chosen bound for the perturbation term.
This inequality suggests a relationship between $V_k$ and $V_{k+1}$ at consecutive steps.

To analyze the convergence of the sequence $V_k$, we apply Polyak's Lemma (Lemma 2 in \citep[Chapter 2.2]{Poliak_1987}), which establishes convergence under certain additional assumptions.  Polyak's Lemma states that if a sequence $V_k$  satisfies an inequality of the form,
\begin{align*}
    V_{k+1} \leq \left(1 + \frac{1}{\gamma_k}\right) V_k - c^-_k + c^+_k,
\end{align*} where $c_k^-$ and $c_k^+$ are sequences of non-negative terms, then the sequence $V_k$ is bounded above provided that $\sum_{k=0}^{\infty}\frac{1}{\gamma_k}< \infty$, and $\sum_{k=0}^{\infty}c_k^+< \infty$. 

We choose $\gamma_k=(k+1)^p$ for some $p>1$ to ensure convergence.
Substituting this choice into the recurrence relation, we obtain:
\begin{align*}
    V_{k+1} \leq \left(1 + \frac{1}{(k+1)^p}\right) V_k  -\frac{2}{3N}\sum_{i=1}^N |x^i_{k+1}-z_{k+1}|^2 -\frac{1}{6N} \left|\frac{1}{\rho N}\left( \sum_{i=1}^N \nabla f^i(x^i_{k+1}) + \grg_{k+1}\right)\right|^2 +  \mathcal{O}(k^p\cdot\Delta^2_k),
\end{align*}for any $\grg_{k+1} \in  \partial g(z_{k+1})$.

By assumption the communication threshold $\Delta_k$ decays as $\Delta_k \sim \mathcal{O}\left(1 / k^t \right)$. Under this assumption, the perturbation term $\mathcal{O}(k^p\cdot\Delta^2_k)$ scales as $\mathcal{O}(k^{p-2t}\cdot \Delta_0^2)$, which satisfies $\sum_{k=0}^{\infty}\gamma_k \Delta^2_k< \infty$ if $p>1$ and $t>\frac{1+p}{2}$. These conditions ensure that the perturbation term decays sufficiently fast, ensuring boundedness of $V_k$.

Finally, by summing over $k=0$ to $K$ and dividing by $K+1$, we obtain the following bound for the average of the residuals and gradient terms:
\begin{align}
    \frac{1}{K+1}\sum_{k=0}^K \left( \frac{2}{3N}\sum_{i=1}^N |x^i_{k+1} - z_{k+1}|^2 + \frac{1}{6 N}\left| \frac{1}{\rho N}\left(\sum_{i=1}^N  \nabla f^i(x^i_{k+1}) + \grg_{k+1} \right)\right|^2\right) \leq \mathcal{O}\left(\frac{1}{K}\right),
\end{align} for any $\grg_{k+1} \in  \partial g(z_{k+1})$, where communication threshold decays $\Delta_k\leq\frac{\Delta_0}{(k+1)^{2}}$.
This result establishes a sublinear convergence rate for both the residuals 
and the gradient terms. 
The rate of decay of the communication threshold ensures that these error terms decrease at a rate proportional to $ \mathcal{O}\left(\frac{1}{K}\right)$, which yields the desired result.
\end{proof}

\newpage
\section{Derivation of Alg.~\ref{alg:relaxed_event} as a Dynamical System}\label{app:dynamics}
In this section, we represent  Alg.~\ref{alg:relaxed_event} as a dynamical system that consists of linear dynamics with a nonlinear feedback interconnection. The communication structure is summarized with Fig.~\ref{fig:Communication_inpaper}.

For the convenience of the reader, we start by restating Alg.~\ref{alg:relaxed_event}, which is based on an over-relaxed ADMM algorithm. 

\begin{algorithm}[h]
\caption{Event-Based Distributed Optimization with Over-Relaxed ADMM}
\label{alg:relaxed_event_app}
\begin{algorithmic}
\REQUIRE Functions $f$ and $g$, matrices $A$ and $B$, vector $c$, parameters $\rho$ and $\alpha$
\REQUIRE Initial condition $x_0,z_0$
\STATE $r_0=\hat{r}^s_{0}=\hat{r}^u_{0}=Ax_0,\quad$  $s_0=\hat{s}^r_{0}=\hat{s}^u_{0}=Bz_0,\quad$  $u_0=\hat{u}^r_{0}=\hat{u}^s_{0}=0$
\FOR{$k=0$ to $t_{\mathrm{max}}$}
    \STATE $\hat{s}_k^r, \hat{u}_k^r \gets$ event-based receive of $s_{k+1}-s_{[k]},u_{k+1}-u_{[k]}$
    \STATE $ x_{k+1}=\arg \min_x f(x)+\frac{\rho}{2}|Ax+\hat{s}^r_k-c+\hat{u}^r_k|^2$  \hfill \COMMENT{r-agent}
    \STATE event-based send of $r_{k+1}-r_{[k]}$ where $r_{k+1}=A x_{k+1}$
    \STATE
    \STATE $\hat{r}_{k+1}^s, \hat{u}_k^s\gets$ event-based receive of $r_{k+1}-r_{[k]},u_{k+1}-u_{[k]}$
    \STATE $ z_{k+1}=\arg \min_z g(z)+\frac{\rho}{2} | \alpha \hat{r}^s_{k+1}-(1-\alpha) Bz_k+ Bz-\alpha c+\hat{u}^s_k |^2 $ \hfill \COMMENT{s-agent}
    \STATE event-based send of $s_{k+1}-s_{[k]}$ where $s_{k+1}=B z_{k+1}$
    \STATE
       \STATE $\hat{r}^u_{k+1},\hat{s}^u_{k+1} \gets$ event-based receive of $r_{k+1}-r_{[k]},s_{k+1}-s_{[k]}$
    \STATE $u_{k+1}=u_k+\alpha \hat{r}^u_{k+1}-(1-\alpha) \hat{s}^u_k+\hat{s}^u_{k+1}-\alpha c$  \hfill \COMMENT{u-agent}  
       \STATE event-based send of $u_{k+1}-u_{[k]}$
       \STATE
    \IF{$\mathrm{mod}(k+1,T)=0$}
    \STATE reset $\rightarrow$ $\hat{r}^{u;s}_{k+1}=r_{k+1}, \hat{s}_{k+1}^{u;r}=s_{k+1}$, $u_{k+1}^{r;s}=u_{k+1}$
    \ENDIF
\ENDFOR
\end{algorithmic}
\end{algorithm}

The following definitions will be useful for simplifying the updates of the iterates:

\begin{definition}\label{def:func_f}
Let Assumption~\ref{ass:matrix} and \ref{ass:func_f} hold. We define the function $\hat{f}: \mathbb{R}^n\rightarrow \mathbb{R}$ as follows,
\begin{align}
     \hat{f}=(\rho^{-1}f) \circ A^{-1},
\end{align}
where $\rho$ is the step-size of Alg.~\ref{alg:relaxed_event}. The function is $\hat{m}:=m/(\rho \bar{\sigma}^2(A))$-strongly convex and $\hat{L}:=L/(\rho\underaccent{\bar}{\sigma}^2(A))$-smooth, and has therefore the condition number
\begin{align*}
\kappa:=\frac{\hat{L}}{\hat{m}}=\frac{L}{m}\frac{\bar{\sigma}^2(A)}{\underaccent{\bar}{\sigma}^2(A)}.
\end{align*}
\end{definition}

\begin{definition}\label{def:func_g}
 Let Assumption~\ref{ass:matrix} and \ref{ass:func_g} hold. The function $\hat{g}: \mathbb{R}^m\rightarrow \mathbb{\bar{R}}$ is defined as
\begin{align}
      \hat{g}=(\rho^{-1}g) \circ B^{\dagger} + \psi_{\text{im}(B)},
\end{align}
where $B^\dagger$ is the Moore-Penrose inverse of $B$, $\psi_{\text{im}(B)}$ is the indicator function of the image of $B$, and $\rho$ is the step-size of Alg.~\ref{alg:relaxed_event}.
\end{definition}

We proceed by summarizing the notation that will be used subsequently. The sequences $r_k$ and $s_k$ are defined as $r_k:=A x_k$ and $s_k:=B z_k$. 
We introduced the variable $\hat{r}_k^s$, for example, which models agent $s$'s estimate of the variable $r_k$. The variables $\hat{r}^u_k, \hat{s}^r_k, \hat{s}^u_k$, etc., are defined analogously and follow the notational convention
\begin{align*}
\widehat{\text{variable}}_k^\text{receiving\_agent} := \text{receiving\_agent's~estimate~of~variable~at~time $k$}.
\end{align*}
As a result of the event-based communication, the local estimates $\hat{r}_k^s, \hat{r}_k^u, \hat{s}^r_k$, etc., differ from $r_k,s_k$, etc. These differences will be captured by the variable $\varepsilon$ for which we introduce the following notational convention:
\begin{align*}
\varepsilon_{k}^{\text{variable},\text{receiving\_agent}}:&=\widehat{\text{variable}}^{\text{receiving\_agent}}_{k}-{\text{variable}}_{k}\\&=\text{receiving\_agent's~estimation error~of~variable~at time $k$.}
\end{align*}
We further introduce the error 
\begin{align}
    e_k:=(\varepsilon^{rs}_{k+1},\varepsilon^{ru}_{k+1}, \varepsilon^{su}_{k+1}, \varepsilon^{sr}_{k}, \varepsilon^{su}_{k}, \varepsilon^{ur}_k, \varepsilon^{us}_{k}),\label{eqn:app_error_state_definition}
\end{align}
that collects the estimation errors of the different agents. By virtue of the event-based communication mechanism and the reset mechanism, the error $e_k$ is bounded by the communication threshold $\Delta$. We finally introduce the notation for the corresponding communication thresholds, $\Delta^{rs},\Delta^{sr}$, etc. (see Fig.~\ref{fig:Communication_inpaper}), according to the same rationale:
\begin{align*}
    \Delta^{\text{variable},\text{receiving\_agent}}:= \text{threshold~for triggering~a~communication~of~variable~to~receiving\_agent.}
\end{align*}
To sum up, the vector $e_k$ contains the errors on the communication lines shown on Fig.~\ref{fig:Communication_inpaper}. For example, $\varepsilon^{rs}_{k+1}$ stands for the difference between the actual value of state $r_{k+1}$ and agent $s$'s estimate $\hat{r}^s_{k+1}$, that is, $\varepsilon^{rs}_{k+1} = \hat{r}^s_{k+1}-r_{k+1}$ at time step $k+1$.  

If the value $r_{k+1}$ has deviated more than $\Delta^{rs}$ amount since the time-step $[k]$, where the last value $r_{[k]}$ has been communicated to the agent $s$, a communication is triggered. This means $rs \in \mathcal{C}_{k+1}$. 
\begin{align*}
|  r_{k+1}-r_{l^{rs}_{k}}|  > \Delta^{rs} \iff rs \in \mathcal{C}_{k+1} \iff [{k+1}]=k+1.\end{align*}
The set $\mathcal{D}_{k+1}$, which is a subset of $\mathcal{C}_{k+1}$, collects indices of failed transmission lines at time step $k+1$. We further introduce that the superscript $c$ to denote the complement of a set.
If the communication does not fail, that is, $rs \in \mathcal{D}^c_{k+1} $, then agent $s$'s estimate of $r_{k+1}$ is updated as follows.
\begin{align*}
&rs \in \mathcal{C}_{k+1} \land  rs \in \mathcal{D}^c_{k+1} \iff \hat{r}^s_{k+1}=\hat{r}^s_k +( r_{k+1}-r_{[{k}]}).\end{align*}
To incorporate the effect of communication drops, we introduce the variable $\chi^{rs}_{k+1}$, which represents the disturbance that results from dropped communications,
\begin{align}
   rs \in \mathcal{D}_{k+1}\Rightarrow\chi^{rs}_{k+1}=-(r_{k+1}-r_{[k]}).  \label{eqn:chi_def} 
\end{align}
Therefore, the dynamics of $\hat{r}_{k+1}^s$ are expressed as follows
\begin{align}
    \hat{r}_{k+1}^s=r_{[{k+1}]} + \sum_{l=1}^{k+1} \chi^{rs}_{l}.\label{eqn:hatr_dyn}
\end{align}
When deriving the previous equation, we have exploited the fact that 
\begin{align*}
rs\in \mathcal{C}_{k+1} &\Rightarrow   [{k+1}]=k+1\\
rs\in \mathcal{C}^c_{k+1} &\Rightarrow   [{k+1}]=[k].
\end{align*}
To summarize, in the case of communication drop, the agent $s$ updates the image of $r$ with a disturbed value. 

We now express the different minimization steps in Alg.~\ref{alg:relaxed_event_app} by their corresponding stationarity conditions, and simplify the corresponding expressions. The minimization step for the primal variable $x$ can be rewritten as follows
\begin{align*}
\begin{aligned}
 x_{k+1} &=\arg \min_{x \in \mathbb{R}^p} f(x)+\frac{\rho}{2}\left|Ax+\hat{s}^r_k-c+\hat{u}^r_k\right|^2\\ &= A^{-1} \arg \min_{r \in \mathbb{R}^n} f(A^{-1}r)+\frac{\rho}{2}\left|r+\hat{s}^r_k-c+\hat{u}^r_k\right|^2,\end{aligned}
\end{align*}
due to the fact that $A$ is invertible, which yields
\begin{align*}
r_{k+1} = \arg\min_{r \in \mathbb{R}^n} \hat{f}(r)+\frac{1}{2}\left|r+\hat{s}^r_k-c+\hat{u}^r_k\right|^2.
\end{align*}
The variable $r_{k+1}$ satisfies therefore the following stationarity condition
\begin{align*}
0=\nabla \hat{f}(r_{k+1})+r_{k+1}+\hat{s}^r_k-c+\hat{u}^r_k,
\end{align*}
which can be rearranged to
\begin{align}
    r_{k+1}-c=-\nabla \hat{f}(r_{k+1})-s_k-\varepsilon^{sr}_k-u_k-\varepsilon^{ur}_{k},\label{eqn:r_exp}
\end{align}
where $\hat{s}^r_k$ is replaced by $s_k+\varepsilon^{sr}_{k}$ and $\hat{u}^r_k$ by $u_k+\varepsilon^{ur}_k$.

Similarly, the update step of the auxiliary variable $z$ can be reformulated as
\begin{align*}
     z_{k+1} &= \arg \min_{z \in \mathbb{R}^n} g(z)+\frac{\rho}{2} | \alpha \hat{r}^s_{k+1}-(1-\alpha) s_k+ B z-\alpha c+\hat{u}^s_k |^2 \\
    &= B^\dagger \arg \min_{s \in \mathbb{R}^m} g(B^\dagger s)+\psi_{\text{im}(B)}(s)+\frac{\rho}{2} | \alpha \hat{r}^s_{k+1}-(1-\alpha) s_k+ s-\alpha c+\hat{u}^s_k |^2,
    \end{align*}
since the matrix $B$ has full column rank and therefore possesses the left inverse $B^\dagger$. This yields
\begin{align*}
          s_{k+1} = \arg \min_{s \in \mathbb{R}^m} \hat{g}(s)+\frac{1}{2} | \alpha \hat{r}^s_{k+1}-(1-\alpha) s_k+ s-\alpha c+\hat{u}^s_k |^2,
\end{align*}
and implies the following stationarity condition for $s_{k+1}$
\begin{align}
    0&\in \partial \hat{g}(s_{k+1}) + \alpha\hat{r}^s_{k+1} - (1-\alpha)s_k + s_{k+1}-\alpha c +\hat{u}^s_k.\label{eqn:s_subdif}
\end{align}
This stationarity condition can be reformulated as
\begin{align}
 s_{k+1}=s_k +(\alpha-1){u}_k+\alpha\nabla \hat{f}(r_{k+1})-\gamma_{k+1}-\varepsilon^{us}_k + \alpha\varepsilon^{sr}_k +\alpha\varepsilon^{ur}_k-\alpha\varepsilon^{rs}_{k+1},\label{eqn:s_exp}
\end{align}
for some $\gamma_{k+1}\in \partial \hat{g}(s_{k+1})$, and where we have expressed $\hat{r}^s_{k+1}$ as $r_{k+1}+\varepsilon^{rs}_{k+1}$ and $\hat{u}^s_k$ as $u_k+\varepsilon^{us}_k$. We have further replaced $r_{k+1}-c$ by the expression given in \eqref{eqn:r_exp}.

The update of the dual variables $u_k$ evolve according to the following dynamics:
\begin{align*}
    u_{k+1}&=u_k+\alpha \hat{r}^u_{k+1}-(1-\alpha)\hat{s}^u_k + \hat{s}^u_{k+1}-\alpha c\\&=u_k+\alpha (r_{k+1}+\varepsilon^{ru}_{k+1})-(1-\alpha)(s_k+\varepsilon_k^{su}) + (s_{k+1}+\varepsilon^{su}_{k+1})-\alpha c.
\end{align*}
The dynamics can be further simplified by replacing $s_{k+1}$ with the help of \eqref{eqn:s_subdif}, which yields: 
\begin{align}
    u_{k+1}&=-\gamma_{k+1}-\alpha\varepsilon^{rs}_{k+1}+\alpha\varepsilon^{ru}_{k+1}+\varepsilon^{su}_{k+1}+(\alpha-1)\varepsilon^{su}_{k}-\varepsilon^{us}_{k}.\label{eqn:u_exp}
\end{align}
As a result of these simplifications, we note that $r_{k+1}$ is uniquely determined by $s_k$, $u_k$ and the corresponding errors $\varepsilon_k^{sr}$ and $\varepsilon_k^{ur}$. We further note that according to \eqref{eqn:s_exp} and \eqref{eqn:u_exp} the iterates of Alg.~\ref{alg:relaxed_event_app} can be represented as an interconnection between a linear dynamical system, with a nonlinear feedback interconnection that models the evaluation of the gradient $\nabla \hat{f}$ and $\partial \hat{g}$. The state of the dynamical system is therefore chosen as $\xi_k:=(s_k,u_k)$, the output as $y_k:=(r_{k+1}-c,s_{k+1})$, and the input as $v_k:=(\nabla \hat{f}(r_{k+1}),\gamma_{k+1})$, where $\gamma_{k+1}\in \partial \hat{g}(s_{k+1})$. We also define output variables $ w^1_k:=(r_{k+1}-c,\nabla \hat{f}(r_{k+1}))$, $ w^2_k:=(s_{k+1}, \gamma_{k+1})$, which will be employed for the convergence analysis.

\begin{figure}
     \centering
    \tikzstyle{block} = [draw, rectangle, minimum height=2.5em, minimum width=2.5em]
\tikzstyle{largeblock} = [draw, rectangle, minimum height=1em, minimum width=5em]
\tikzstyle{smallblock} = [draw, rectangle, minimum height=0.5em, minimum width=0.5em]
\tikzstyle{longblock} = [draw, rectangle, minimum height=5em, minimum width=1em]

\tikzstyle{sum} = [draw, circle, node distance=2cm]
\tikzstyle{input} = [coordinate]
\tikzstyle{elbow} = [coordinate]

\tikzstyle{pinstyle} = [pin edge={to-,thin,black}]

\begin{tikzpicture}[every node/.style={font=\small}]

% Blocks
\node [input](V) {};
\node [block, right= 0.5 cm of $(V)$] (linear) {$\begin{array}{c|cc}
    \hat{A} & \hat{E}&\hat{B} \\   \hline
       \hat{C} & \hat{E}_y&\hat{D}  
\end{array}$};

\node [smallblock, below= 0.7 cm of $(linear)$] (nonlinear) {$\phi$};

\node [input, right= 1.8 cm of $(linear)$] (Error) {$e$};

\draw [->] (Error) -- (linear.east);% <-- this solve your problem

\node[right] at (Error) {$e$};

% Connections
\draw [-] (V) --  (linear);
\draw [<-] ([yshift=-0.25cm]linear.east) -| node[midway, left, xshift=2.25cm, yshift=-0.4cm] {\small$\left[\begin{array}{c}
     \nabla \hat{f}(r_{k+1}) \\ \gamma_{k+1} \end{array}\right]$} + (0.5,0) |- (nonlinear);

\draw [<-] (nonlinear) -| node[near end, left] {\small$\left[\begin{array}{c}
     r_{k+1}-c \\ s_{k+1} \end{array}\right]$} (V);

\node [smallblock, above= 0.8 cm of $(linear)$] (lag) {Delay};

\draw [->](lag) -| + (1.75,-0.75)  |- node[near start, right, yshift=0.4cm] {\small$\left[\begin{array}{c}
     s_{k} \\ u_{k} \end{array}\right]$} ([yshift=0.25cm]linear.east);

\draw [<-](lag) -- + (-1.5,0.0)  |- node[midway, left, yshift=0.5cm] {\small$\left[\begin{array}{c}
     s_{k+1} \\ u_{k+1} \end{array}\right]$} ([yshift=0.25cm]linear.west);

\end{tikzpicture}
  \caption{The dynamical system following \eqref{eqn:dynsys2} is visualized.}
    \label{fig:Lures}
\end{figure}
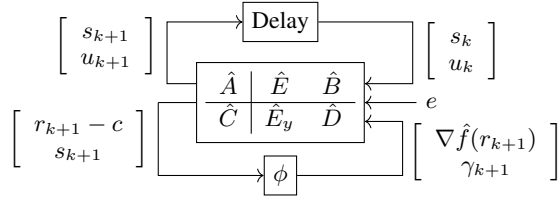

According to these definitions, we can express the iterates of Alg.~\ref{alg:relaxed_event_app} as trajectories of the following nonlinear dynamical system,
\begingroup % keep the change local
\setlength\arraycolsep{2pt}
\begin{align}
&\begin{aligned}
       \xi_{k+1}&=\underbrace{\begin{bmatrix}
         1&\alpha-1\\0&0
     \end{bmatrix}}_{:=\hat{A}}\xi_k +\underbrace{\begin{bmatrix}
         \alpha& -1\\0&-1\end{bmatrix}}_{:=\hat{B}}v_k+ \underbrace{\begin{bmatrix}
       -\alpha& 0&0&\alpha&0& \alpha&-1\\-\alpha&\alpha&1&0&\alpha-1&0&-1
     \end{bmatrix}}_{:=\hat{E}} e_k, \quad v_k=\phi(y_k),\\
   y_k&=\underbrace{\begin{bmatrix}
         -1&-1\\1&\alpha-1
     \end{bmatrix}}_{:=\hat{C}}\xi_k+\underbrace{\begin{bmatrix}
         -1& 0\\\alpha&-1\end{bmatrix}}_{:=\hat{D}}v_k+ \underbrace{\begin{bmatrix}
       0& 0&0&-1&0&-1&0\\-\alpha&0&0&\alpha&0&\alpha&-1
     \end{bmatrix}}_{:=\hat{E}^y}e_k, 
    \label{eqn:dynsys2}\end{aligned}\\
&~~\begin{aligned}
     w^1_k&=\underbrace{\begin{bmatrix}
         -1&-1\\0&0
     \end{bmatrix}}_{:=\hat{C}^1}\xi_k +\underbrace{\begin{bmatrix}
         -1& 0\\1&0\end{bmatrix}}_{:=\hat{D}^1} v_k+ \underbrace{\begin{bmatrix}
0&0&0&-1&0&-1&0\\0&0&0&0&0&0&0
     \end{bmatrix}}_{:=\hat{E}^1}e_k,\\   w^2_k&=\underbrace{\begin{bmatrix}
         1&\alpha-1\\0&0
     \end{bmatrix}}_{:=\hat{C}^2}\xi_k+\underbrace{\begin{bmatrix}
         \alpha& -1\\0&1\end{bmatrix}}_{:=\hat{D}^2}v_k+ \underbrace{\begin{bmatrix}
   -\alpha& 0&0&\alpha&0& \alpha&-1\\0&0&0&0&0&0&0
     \end{bmatrix}}_{:=\hat{E}^2}e_k,\end{aligned}
\end{align}
\endgroup
where $\phi$ denotes the nonlinear feedback interconnection that captures the evaluation of the gradients $\nabla \hat{f}$ and $\partial \hat{g}$.
Fig.~\ref{fig:Lures} provides a graphical representation of the time-invariant dynamics determined by the matrices $\hat{A},\hat{B},\hat{C},\hat{D}$.

We close the section with the following proposition that shows that $|e_k|$ is bounded.
\begin{proposition}
\label{prop:boundederr2}
The error $e_k$ at iteration $k$ is bounded by
\begin{align*}
|e_k| \leq \Delta, \quad \Delta:=\sum_{l\in \{ rs,ru,su,sr,su,ur,us \}} \Delta^l + T \bar{\chi}^l,
\end{align*}
where the variable $\bar{\chi}^l$ is an upper bound on the communication drops.
\end{proposition}
\begin{proof}
The proof is analogous to Prop.~\ref{prop:boundederr}. The error resulting from the event-based communication structure is given by
\begin{align}
    \varepsilon_{k+1}^{rs}&=\hat{r}^s_{k+1}-r_{k+1}=\underbrace{r_{[k+1]}-r_{k+1}}_{\mathrm{I}}+\underbrace{\sum_{l=1}^{k+1} \chi^{rs}_{l}}_{\mathrm{II}}.
\end{align}
We further note that the first term is bounded by $\Delta^{rs}$ by virtue of the communication rule
\begin{align*}
    |r_{k+1}-r_{[k+1]}| \leq \Delta^{rs}.
\end{align*}
Through the assumption $|\chi^{rs}_{l}|\leq\Bar{\chi}^{rs}$, the second part is bounded by $T\Bar{\chi}^{rs}$, where $T$ is the reset period. Therefore, we conclude that $|e_{k+1}^{rs}|$ is bounded by $\Delta^{rs}+T\Bar{\chi}^{rs}$. Similarly, the other elements of the vector $e_k$ are bounded by  $\Delta^{ru}+T\Bar{\chi}^{ru}$,  $\Delta^{su}+T\Bar{\chi}^{su}$, etc. Hence, $|e_k|$ is bounded by $\Delta$ where
$$\Delta=\sum_{l\in \{ rs,ru,su,sr,su,ur,us\}} \Delta^{l}+T\Bar{\chi}^{l}.$$\end{proof}

The analysis indicates that a periodic reset with a period $T$ is required to achieve a bounded error. If no reset is included, Alg.~\ref{alg:relaxed_event} may not converge, which could result in a large error that accumulates over time. The dependence of $\Delta$ on the period $T$ highlights how the reset period $T$ affects the error (where smaller $T$ leads to a smaller error bound). If there are no communication failures, there is also no need for a reset ($\Bar{\chi}=0$), and $\Delta$ reduces to the collection of communication thresholds.

\newpage
\section{Convergence Analysis}\label{app:convergence}
We first start by proving the following intermediate lemmas.
\begin{lemma}
\label{lemma:M1}
Let Assumption~\ref{ass:func_f} be satisfied.
Then, the following holds,
\begingroup
\setlength\arraycolsep{1pt}
    \begin{align}
        \begin{bmatrix}(r_1-r_2)^\top& (\nabla \hat{f}(r_1)-\nabla \hat{f}(r_2))^\top\end{bmatrix} \left( \begin{bmatrix}
            -2\hat{m}\hat{L}&(\hat{m}+\hat{L})\\(\hat{m}+\hat{L})&-2
        \end{bmatrix}\!\otimes\!I_n \right)\begin{bmatrix}r_1-r_2\\ \nabla \hat{f}(r_1)-\nabla \hat{f}(r_2)\end{bmatrix} \geq 0,
    \end{align}
    \endgroup
    for all $r_1,r_2\in \mathbb{R}^n$.
\begin{proof}
We define the auxiliary function $\tilde{f}(r):=\hat{f}(r)-\frac{\hat{m}}{2}|r|^2 $, which is $\hat{L}-\hat{m}$-smooth and convex by the properties of $\hat{f}$. Then the following inequality holds,
\begin{align*}
 (\nabla \tilde{f}(r_1) - \nabla \tilde{f}(r_2))^\top (r_1-r_2 )\geq \frac{1}{\hat{L}-\hat{m}}| \nabla \tilde{f}( r_1)-\nabla \tilde{f} (r_2)|^2,\end{align*}
for any $r_1,r_2\in \mathbb{R}^n$. Substituting $\tilde{f}(r):=\hat{f}(r)-\frac{\hat{m}}{2}|r|^2 $ and $\nabla \tilde{f}(r)=\nabla \hat{f}(r)-\hat{m}r$, we get
\begin{align*}
(\hat{m}+\hat{L})(r_1-r_2)^\top(\nabla \hat{f}(r_1)-\nabla \hat{f}(r_2))\geq \hat{m}\hat{L}| r_1-r_2|^2+|\nabla \hat{f}(r_1)-\nabla \hat{f}(r_2)|^2,
\end{align*}
which yields the desired result.\end{proof}
\end{lemma}

\begin{lemma}
\label{lemma:M2}
Let Assumption~\ref{ass:func_f} be satisfied. Then, the following holds, 
\begin{align}
         \begin{bmatrix}(s_1-s_2)^\top& (\gamma_1-\gamma_2)^\top\end{bmatrix}  \left(\begin{bmatrix}
            0&1\\1&0
        \end{bmatrix}\otimes I_m\right)\begin{bmatrix}s_1-s_2\\ \gamma_1-\gamma_2\end{bmatrix}\geq 0,
\end{align}
where $\gamma_1\in \partial \hat{g}(s_1)$ and $\gamma_2\in \partial \hat{g}(s_2)$ and for any $s_1,s_2\in \mathbb{R}^m$.
\begin{proof}
The subdifferential of a convex function is a monotone operator, and therefore
\begin{align*}
            (s_1-s_2)^\top (\gamma_1-\gamma_2) \geq 0.
\end{align*}\end{proof}
\end{lemma}

\begin{lemma}\label{lemma:w}
    Let $x_*,z_*$ denote the minimizer of \eqref{eqn:main_problem} and define $r_*:=A x_*$, $s_*:=B z_*$, $\beta_*:=\nabla \hat{f}(r_*)$, and $\gamma_*\in \partial \hat{g}(s_*)$. Then, the iterates of Alg.~\ref{alg:relaxed_event_app} with step-size $\rho=\rho_0 (\hat{m}\hat{L})^{\frac{1}{2}}$ satisfy
    \begin{align*}
         (w^i_k-w^i_\ast)^\top M^i (w^i_k-w^i_\ast)\geq 0, \quad \forall i \in \{1,2\}, \quad \forall k\geq 0,
\end{align*}
with
\begin{align*}
M^1:=\left[\begin{array}{cc}
-2 \rho_0^{-2} & \rho_0^{-1}\left(\kappa^{-\frac{1}{2}}+\kappa^{\frac{1}{2}}\right) \\
\rho_0^{-1}\left(\kappa^{-\frac{1}{2}}+\kappa^{\frac{1}{2}}\right) & -2
\end{array}\right]\otimes I_n, \quad M^2:=\begin{bmatrix}
            0&1\\1&0
        \end{bmatrix}\otimes I_m,
\end{align*}
where 
\begin{align*}
    w^1_k:=\left[\begin{array}{c} r_{k+1}-c  \\\beta_{k+1} \end{array}\right], \quad w^2_k:=\left[\begin{array}{c} s_{k+1} \\\gamma_{k+1} \end{array}\right], \quad w_\ast^1:=\left[\begin{array}{c} r_\ast-c \\ \beta_\ast\end{array}\right],\quad w_\ast^2:=\left[\begin{array}{c} s_\ast \\ \gamma_\ast \end{array}\right].
\end{align*}
\end{lemma}
\begin{proof} The proof follows directly from Lemma~\ref{lemma:M1} and Lemma~\ref{lemma:M2}.
\end{proof}

\begin{lemma} \label{Vlemma}
Let the sequence $V_k \geq 0$ satisfy
\begin{align}
    V_{k+1}\leq V_k (1-\tilde\alpha) +\tilde\beta \tilde\alpha,
\end{align}
for all $k\geq0$, where the parameters $\tilde \alpha$, $\tilde \beta$ satisfy $0<\tilde\alpha<1$ and $0\leq\tilde\beta$. Then, the following holds for all $k\geq0$:
\begin{align}
    V_k \leq V_0 (1-\tilde\alpha)^k +\tilde\beta.
\end{align}
\end{lemma}
\begin{proof} We prove the lemma by induction.

The claim holds for $k=0$. We therefore assume that the claim holds for $k$ and show that, as a result, the claim holds for $k+1$. More precisely,
\begin{align}
\begin{aligned}
  V_{k+1}&\leq V_k (1-\tilde\alpha) +\tilde\beta \tilde\alpha\\
  &\leq V_0 (1-\tilde\alpha)^{k+1} +(1-\tilde\alpha)\tilde\beta+\tilde\beta \tilde\alpha\\
  &\leq V_0 (1-\tilde\alpha)^{k+1} +\tilde\beta,\end{aligned}\end{align} which completes the induction argument.
\end{proof}

In App.~\ref{app:dynamics}, we expressed the iterates of Alg.~\ref{alg:relaxed_event_app} as the trajectories of a dynamical system. The dynamical system was given as a linear time-invariant system that was interconnected in feedback with a nonlinear function $\phi$. We now arrive at the main result that will be used to show convergence of Alg.~\ref{alg:relaxed_event_app}.

\begin{theorem}\label{prop:main_result}
Let Assumption~\ref{ass:func_f} be satisfied, let the step-size for Alg.~\ref{alg:relaxed_event_app} be $\rho=\rho_0(\hat{m}\hat{L})^{\frac{1}{2}}$, and let $\xi_\ast=(B z_\ast,u_\ast),$ where $(x_\ast,z_\ast)$ is the minimizer of \eqref{eqn:main_problem} and $u_\ast$ the corresponding dual variable.
    
Suppose there exists a positive definite matrix $P\succ 0$, $0<\tau<1$, and nonnegative constants $\lambda^1$, $\lambda^2$, $\gamma^1$, $\gamma^2$, $\gamma^3$ and $\gamma^4$ such that the following linear matrix inequality 
\begingroup % keep the change local
\setlength\arraycolsep{1pt}
\begin{align}
    \begin{aligned}    
0 \succeq &\left[\begin{array}{cc}
(1\!+\!\gamma^1)\hat{A}^{\top} P \hat{A}\!-\!\tau^2 P & \hat{A}^{\top} P \hat{B} \\
\hat{B}^{\top} P \hat{A} &(1\!+\!\gamma^2)\hat{B}^{\top} P \hat{B} 
\end{array}\right]+\left[\begin{array}{ll}
\hat{C}^1 & \hat{D}^1 \\
\hat{C}^2 & \hat{D}^2
\end{array}\right]^{\top}\left[\begin{array}{cc}
\Lambda^1 M^1 & 0 \\
0 & \Lambda^2 M^2 
\end{array}\right]\left[\begin{array}{ll}
\hat{C}^1 & \hat{D}^1 \\
\hat{C}^2 & \hat{D}^2
\end{array}\right]  \label{eqn:statmentLMI}\end{aligned}
\end{align} \endgroup
is satisfied, where $\Lambda^1=\lambda^1(1+\gamma^3) $, $\Lambda^2=\lambda^2(1+\gamma^4)$. Then, for all $k \geq 0$, we have
\begin{equation}
    |\xi_{k}-\xi_\ast|^2 \leq {\kappa_P} |\xi_{0}-\xi_\ast|^2\tau^{2k} + \frac{\bar{\sigma}( Q)\Delta^2}{\underaccent{\bar}{\sigma}(P)(1-\tau^2)},
\end{equation}
where $\kappa_P={\Bar{\sigma}(P)}/{\underaccent{\bar}{\sigma}(P)}$ denotes the condition number of the matrix $P$, $\Delta$ is a bound on the error $e_k$ (see Prop.~\ref{prop:boundederr2}), and \begin{align}
&\begin{aligned}
Q=&\left(1+ \frac{1}{\gamma^1}+\frac{1}{\gamma^2} \right)  \hat{E}^\top P \hat{E} + \left(1 + \frac{1}{\gamma^3}+ \frac{1}{\gamma^4}\right) \sum_{i=1}^2 \lambda^i {\hat{E}^{i\top}} M^i \hat{E}^i.\end{aligned}\label{eqn:Q}
\end{align}
\end{theorem}

\begin{proof} We consider the following quadratic storage function,
\begin{align*}
    V_{k}=(\xi_k-\xi_\ast)^\top P (\xi_k-\xi_\ast),
\end{align*}
and claim that the following inequality holds for the iterates of Alg.~\ref{alg:relaxed_event_app}:
\begin{align*}
\begin{aligned}
V_{k+1}- \tau^2 V_{k} + \sum_{i=1}^2 \lambda^i & (w^i-w^i_\star)^\top  M^i (w^i-w^i_\star)\leq \\ 
&e_k^\top\Big(\left(1+ \frac{1}{\gamma^1}+\frac{1}{\gamma^2} \right) E^\top P  E + \sum_{i=1}^2 \lambda^i \left(1 + \frac{1}{\gamma^3}+ \frac{1}{\gamma^4}\right) {E^{i\top}} M^i E^i \Big)e_{k}.
\end{aligned}
\end{align*}

\textit{Proof of the claim:} 
We insert the system dynamics stated in App.~\ref{app:dynamics} into the expression on the left-hand side, which yields
\begin{align}\begin{aligned}
&V_{k+1}-\tau^2V_{k} + \sum_{i=1}^2 \lambda^i (w_k^i-w^i_\star)^\top M^i (w_k^i-w^i_\star)\\=&(\mathbf{\xi}_{k+1}-\mathbf{\xi}_{\star})^T P (\mathbf{\xi}_{k+1}-\mathbf{\xi}_{\star})-\tau^2(\mathbf{\xi}_{k}-\mathbf{\xi}_{\star})^T P (\mathbf{\xi}_{k}-\mathbf{\xi}_{\star}) + \sum_{i=1}^2 \lambda^i (w^i_k-w^i_\star)^\top M^i (w^i_k-w^i_\star)\\=&\tilde{\xi}_k^\top \left( A^\top P A -\tau^2 P\right)\tilde{\xi}_k +\tilde{\upsilon}_{k}^\top \hat{B}^\top P\hat{B}\tilde{\upsilon}_{k} + e_{k}^\top E^\top P Ee_{k} \\&+ 2 \left( \tilde{\upsilon}_{k}^\top \hat{B}^\top  P \hat{A}\tilde{\xi}_k  + e_{k}^\top E^\top P \hat{A}\tilde{\xi}_k +  \tilde{\upsilon}_{k}^\top \hat{B}^\top P Ee_{k}
\right)\\
&+ \sum_{i=1}^2 \lambda^i \left( \tilde{\xi}_k^\top  {\hat{C}^{i\top}} M^i {\hat{C}^i} \tilde{\xi}_k +\tilde{\upsilon}_{k}^\top {\hat{D}^{i\top}} M^i {\hat{D}^i}\tilde{\upsilon}_{k} + e_{k}^\top {E^{i\top}} M^i E^ie_{k}\right) \\&+ 2  \sum_{i=1}^2 \lambda^i \left( \tilde{\upsilon}_{k}^\top {\hat{D}^{i\top}}  M^i {\hat{C}^i}\tilde{\xi}_k  + e_{k}^\top {E^{i\top}}M^i {\hat{C}^i}\tilde{\xi}_k +  \tilde{\upsilon}_{k}^\top {\hat{D}^{i\top}} M^i {E^i}e_{k} \right),  \end{aligned}\label{eq:Lyap_IQC}
\end{align}
where $\tilde{\xi}_k=\mathbf{\xi}_{k}-\mathbf{\xi}_{\star}$, and $\tilde{\upsilon}_k=\mathbf{\upsilon}_{k}-\mathbf{\upsilon}_{\star}$, for simplicity. We now apply Young's inequality on the cross terms in \eqref{eq:Lyap_IQC}, which yields
\begin{align}\begin{aligned}
    &V_{k+1}- \tau^2 V_{k} + \sum_{i=1}^2 \lambda^i (w^i_k-w^i_\star)^\top M^i (w^i_k-w^i_\star) \\ & \leq \tilde{\xi}_{k}^\top \left( A^\top P A -\tau^2 P\right)\tilde{\xi}_{k} +\tilde{\upsilon}_{k}^\top \hat{B}^\top P\hat{B}\tilde{\upsilon}_{k} + e_{k}^\top E^\top P Ee_{k} +2\tilde{\upsilon}_{k}^\top \hat{B}^\top  P \hat{A}\tilde{\xi}_{k}\\ & + \gamma^2 ( \tilde{\upsilon}_{k}^\top \hat{B}^\top P\hat{B}\tilde{\upsilon}_{k} ) + \frac{1}{\gamma^2} (e_k^\top E^\top P E e_k)+ \gamma^1 ( \tilde{\xi}_{k}^\top \hat{A}^\top P \hat{A}\tilde{\xi}_{k} ) + \frac{1}{\gamma^1} (e_k^\top E^\top P E e_k)\\&+ \sum_{i=1}^2 \lambda^i \left( \tilde{\xi}_{k}^\top  {\hat{C}^{i\top}} M^i {\hat{C}^i} \tilde{\xi}_{k} +\tilde{\upsilon}_{k}^\top {\hat{D}^{i\top}} M^i {\hat{D}^i}\tilde{\upsilon}_{k} + e_{k}^\top {E^{i\top}} M^i E^ie_{k} + 2  \left( \tilde{\upsilon}_{k}^\top {\hat{D}^{i\top}}  M^i {\hat{C}^i}\tilde{\xi}_{k} \right) \right) \\ &+\sum_{i=1}^2 \lambda^i\left(\gamma^4  \tilde{\upsilon}_{k}^\top \hat{D}^{i\top} M^i \hat{D}^i\tilde{\upsilon}_{k}  +\frac{1}{\gamma^4} e_k^\top E^{i\top} M^i E^i e_k+ \gamma^3 \tilde{\xi}_{k}^\top \hat{C}^{i\top} M^i \hat{C}^i\tilde{\xi}_{k} + \frac{1}{\gamma^3} e_k^\top E^{i\top} M^i E^i e_k\right).\end{aligned}\label{eq:IQC_lyap}\end{align}
If we rearrange the right-hand side of the inequality in matrix form, we obtain,
\begingroup % keep the change local
\setlength\arraycolsep{2pt}
\begin{align}
\begin{aligned}
V_{k+1}- \tau^2 V_{k} + &\sum_{i=1}^2 \lambda^i (w^i_k-w^i_\star)^\top  M^i (w^i_k-w^i_\star)\leq\\
    \left[\begin{array}{cc} \tilde{\xi}^\top_k &\tilde{\upsilon}^\top_k\end{array}\right] \Bigg(&
\left[\begin{array}{cc}
(1+\gamma^1)\hat{A}^{\top} P \hat{A}-\tau^2 P & \hat{A}^{\top} P \hat{B} \\
\hat{B}^{\top} P \hat{A} & (1+\gamma^2 )\hat{B}^{\top} P \hat{B} 
\end{array}\right]\\&+\left[\begin{array}{ll}
\hat{C}^1 & \hat{D}^1 \\
\hat{C}^2 & \hat{D}^2
\end{array}\right]^{\top}\left[\begin{array}{cc}
\lambda^1 M^1 (1+\gamma^3) & 0 \\
0 & \lambda^2 M^2 (1+\gamma^4)
\end{array}\right]\left[\begin{array}{ll}
\hat{C}^1 & \hat{D}^1 \\
\hat{C}^2 & \hat{D}^2
\end{array}\right] \Bigg)\left[\begin{array}{c} \tilde{\xi}_{k} \\\tilde{\upsilon}_k\end{array}\right]\\& +e_k^\top\left(\left(1+ \frac{1}{\gamma^1}+\frac{1}{\gamma^2} \right)  E^\top P  E + \sum_{i=1}^2 \lambda^i \left(1 + \frac{1}{\gamma^3}+ \frac{1}{\gamma^4}\right) {E^{i\top}} M^i E^i \right)e_{k}.
\end{aligned}
\end{align}
\endgroup
The fact that the linear matrix inequality \eqref{eqn:statmentLMI} is satisfied proves the claim. Furthermore, we conclude that $  \sum_{i=1}^2 \lambda^i (w^i_k-w^i_\star)^\top  M^i (w^i_k-w^i_\star) \geq 0
$ from Lemma~\ref{lemma:M1} and \ref{lemma:M2}. This simplifies the previous expression to
\begin{align*}
\begin{aligned}
V_{k+1}\leq \tau^2 V_{k}+ e_k^\top Q e_{k},
\end{aligned}\end{align*}
where we have also inserted the definition of the matrix $Q$. The right-hand side can further be bounded by virtue of the reset mechanism and the event-based communication, which results in
\begin{align}
\begin{aligned}
V_{k+1}\leq \tau^2 V_{k}+ \bar{\sigma}(Q)\Delta^2.
\end{aligned}\label{eqn:lyapunov_ineq}
\end{align}
We are now in a position where we can apply Lemma~\ref{Vlemma}, which concludes
\begin{align*}
\begin{aligned}
V_{k}\leq \tau^{2k} V_{0}+ \frac{\bar{\sigma}(Q)\Delta^2}{1-\tau^2}. \end{aligned}
\end{align*}
By definition of the quadratic storage function we conclude
\begin{align*}
\begin{aligned}
 |\xi_{k}-\xi_\ast|^2 \leq \tau^{2k}\frac{\Bar{\sigma}(P)}{\underaccent{\bar}{\sigma}(P)} |\xi_{0}-\xi_\ast|^2+ \frac{\bar{\sigma}(Q)\Delta^2}{\underaccent{\bar}{\sigma}(P)(1-\tau^2)}, \end{aligned}
\end{align*}
which implies the result of Thm.~\ref{prop:main_result}. \end{proof}

We are now ready to prove our main result in Thm.~\ref{thm:symbolic_rate}.

\textit{Proof of Thm.~\ref{thm:symbolic_rate}.}
The proof is based on Thm.~\ref{prop:main_result} which shows that if the matrix inequality

    \begin{align}\label{eqn:sym_LMI}
    \begin{aligned}    
0 \succeq \left[\begin{array}{cc}
(1+\gamma^1)\hat{A}^{\top} P \hat{A}-\tau^2 P & \hat{A}^{\top} P \hat{B} \\
\hat{B}^{\top} P \hat{A} &(1+\gamma^2)\hat{B}^{\top} P \hat{B} 
\end{array}\right]+\left[\begin{array}{ll}
\hat{C}^1 & \hat{D}^1 \\
\hat{C}^2 & \hat{D}^2
\end{array}\right]^{\top}\left[\begin{array}{cc}
\Lambda^1 M^1 & 0 \\
0 & \Lambda^2 M^2 
\end{array}\right]\left[\begin{array}{ll}
\hat{C}^1 & \hat{D}^1 \\
\hat{C}^2 & \hat{D}^2
\end{array}\right]  \end{aligned}
\end{align} 

is satisfied for a symmetric positive definite matrix $P$ and for positive constants $\Lambda^1, \Lambda^2,\gamma^1,\gamma^2$, the following bound holds 
\begin{align*}
  |\xi_{k}-\xi_\ast|^2 \leq {\kappa_P} |\xi_{0}-\xi_\ast|^2\tau^{2k} + \frac{\bar{\sigma}( Q)\Delta^2}{\underaccent{\bar}{\sigma}(P)(1-\tau^2)},
\end{align*}
where $\kappa_P$ denotes the condition number of $P$ and $Q$ is defined in App.~\ref{app:convergence}. In fact, the following set of parameters satisfies the linear matrix inequality \eqref{eqn:sym_LMI},
\begin{align*}
\begin{aligned}
    P=\begin{bmatrix} 1 & \alpha -1\\ \alpha -1 & 1-\frac{1}{\sqrt{\kappa}} \end{bmatrix},\quad\tau=1-\frac{\alpha }{4\,\kappa ^{\epsilon +\frac{1}{2}}},\quad\Lambda^1=\alpha \,\kappa ^{\epsilon -\frac{1}{2}},\quad\Lambda^2=\alpha,\quad\gamma^1=\frac{\alpha }{\kappa ^{\epsilon +\frac{3}{2}}},\quad\gamma^2=\frac{1}{\kappa}. \end{aligned}%\label{eqn:gamma_choice}
\end{align*}

This can be checked as follows: The matrix on the right-hand side of \eqref{eqn:sym_LMI} can be expressed as $-\frac{1}{4}\kappa^{-2}\mathbb{L}$, where $\mathbb{L}$ is a symmetric $4\times4$ matrix (compared to earlier analyses \citep{Nishihara_2015}, the last row and last column is not zero). We now prove that $\mathbb{L}$ is positive semidefinite for all sufficiently large $\kappa$ by checking the leading principle minors, which can be expressed as polynomials in $\kappa$. If the leading terms of the principle minors have positive coefficients, it means that for large enough $\kappa$, the principle minor will indeed be positive.

The leading term for the first principle minor is given by $6\kappa^{\frac{3}{2}-\epsilon}$ and is therefore positive. Likewise, the second principle minor is dominated by the positive term $24 (2-\alpha) \kappa^{\frac{7}{2}-\epsilon}$. For the third leading principle minor, there are two different cases. If $\epsilon=0$, the leading term of the third leading principle minor is $16\kappa^{5}(\alpha^4 - 4\alpha^3 - 4\alpha^2 +22\alpha - 12)/\alpha$, which is positive for $\alpha\in(0.675,2)$. If $\epsilon>0$, the leading term of the third principle minor becomes $192\kappa^5(2-\alpha)$, which is positive for $\alpha\in(0,2)$. Finally, for the fourth principle minor, there are also two different cases. If $\epsilon=0$, the leading term is $ 64\kappa^{\frac{13}{2}}(\alpha^4 - 4\alpha^3 - 4\alpha^2 +22\alpha - 12)/\alpha^2$, which is positive for $\alpha\in(0.675,2) $. If $\epsilon>0$, the leading term of the fourth principal minor becomes $768\kappa^{\frac{13}{2}}(2-\alpha)/\alpha$, which is positive for $\alpha\in(0,2)$. In conclusion, for all sufficiently large $\kappa$, all four leading principle minors are positive, which implies that $\mathbb{L}$ is positive definite.

It remains to bound the second term $\bar{\sigma}(Q)/(\underaccent{\bar}{\sigma}(P)(1-\tau^2))$. We again investigate the symbolic expression, and conclude that the term is always bounded by $60\kappa^{2+2\epsilon}/(\alpha(1-|\alpha-1|))$ for large enough $\kappa$. This concludes the proof.\qed

\newpage
\section{Bound on Event-Based Error Variables}\label{app:ProofProp2}
We restate Prop.~\ref{prop:boundederr} and present its proof.

\begin{proposition*}
    The error $\hat{\zeta}_k\!-\!\zeta_k$ at iteration $k$ is bounded by $|\hat{\zeta}_k\!-\!\zeta_k |\! \leq\! \Delta^d\!+\! T \bar{\chi}^d$,
where $T$ denotes the reset period (see Alg.~\ref{alg:over_relaxed_consensus}) and $\bar{\chi}^d$ is a bound on the disturbance $\chi_k^{di}$. 
\end{proposition*}

\begin{proof}
We note that the error $\hat{\zeta}_k-\zeta_k$ can be expressed as
\begin{align*}
   \hat{\zeta}_k-\zeta_k= \frac{1}{N}  \sum_{i=1}^N  \underbrace{( d_{[k+1]}^i -d^i_{k+1})}_{\mathrm{I}}+\underbrace{ \sum_{l=T_{[k]}}^k\chi^{di}_{l+1}}_{\mathrm{II}},
\end{align*}
where $T_{[k]}$ denotes the last time instant where a reset has been performed. The terms $\mathrm{I}$ and $\mathrm{II}$ have each a clear interpretation: In the absence of communication failures, $\hat{\zeta}$ is an average over the primal and dual variables, $x_{[k+1]}^i$ and $u_{[k]}^i$, that were last communicated, which leads to the term $\mathrm{I}$. The term $\mathrm{II}$ captures the dropped information through failures. The communication protocol ensures that $|d^i_{k+1}\!-\!d^i_{[k+1]}| \!\leq\!\Delta^{d}$, for all $k\geq 0$, which means that the term $\mathrm{I}$ is bounded by $\Delta^d$. The bound for the term $\mathrm{II}$ arises from the triangle inequality, which yields, $\bar{\chi}^d$ and concludes the proof. 
\end{proof}

The previous proposition required the variable $\chi_{k+1}^{di}$ to be bounded. Prop.~\ref{prop:2} establishes such a bound under standard conditions on $f$ and $g$.

\begin{proposition}\label{prop:2}
Let $f$ be $L$-smooth and convex and let $\{z\in \mathbb{R}^n~|~g(z)< \infty\}$ be contained in a ball of radius $R$. Then, the disturbances $\chi_k^{di}$ and $\chi_k^{zi}$ are bounded by 
\begin{align*}
|\chi_k^{zi}|\leq 2R, ~~|\chi_k^{di}|\leq (\alpha+1) \frac{2(\rho+L)}{\rho} |x^{i}_\ast| + 2R, 
\end{align*}
for all $i\!=\!1,\dots,N$ and all $k\!\geq\ 0$, where $x^{i}_\ast:=\arg\min_{x\in \mathbb{R}^n} f^i(x)\!+\!\rho |x|^2/2$, and where $\chi_k^{zi}$ denotes the communication drops when communicating $z_k$ between agent $N+1$ and agent $i$.\end{proposition}
\begin{proof} 

Due to the assumption that the domain of $g$ is contained in a ball of radius $R$, we conclude $|z_{k}|\leq R$ for all $k\geq 0$. This also implies that $|\hat{z}_k^i|\leq R$ and concludes the bound on $\bar{\chi}^z$. For obtaining the remaining two bounds, we analyze the $x^i,u^i$ dynamics of agent $i$, where we introduce the convex conjugate
\begin{align*}
\bar{f}^{i}(u,z)=\sup_{x\in \mathbb{R}^n} u^\mathrm{T} x - f^i(x)-\frac{\rho}{2} |x-z|^2.
\end{align*}
We note that the supremum is attained for $x^i_{k+1}$, if $u=-\rho u_k^i$ and $z=\hat{z}_k^i$ in the previous equation. In addition, due to the properties of the convex conjugate, we conclude that $\bar{f}^i(\cdot,z)$ is $1/\rho$-smooth and $1/(\rho+L)$-strongly convex. The conjugate subgradient theorem implies,
\begin{align*}
\nabla_u \bar{f}^i(-\rho u_k^i,\hat{z}_k^i) = x_{k+1}^i,
\end{align*}
which means that the updates for $u_k^i$ can now be expressed as:
\begin{align*}
    u_{k+1}^i = u_k^i+\nabla_u \bar{f}^i(-\rho u_k^i,\hat{z}_k^i)-\hat{z}_k^i.
\end{align*}
By applying Taylor's theorem, we obtain
\begin{align*}
u_{k+1}^i =  u_k^i+\nabla^2_u \bar{f}^i(\nu_k,\hat{z}_k^i) (-\rho u_k^i)-\hat{z}_k^i+\nabla_u \bar{f}^i(0,\hat{z}_k^i),
\end{align*}
for some $\nu_k\in \mathbb{R}^n$. By leveraging the fact that, as a result of strong convexity and smoothness of $\bar{f}^i(\cdot,z)$, the Hessian $\nabla_u^2 \bar{f}^i(\cdot,z)$ is upper and lower bounded by $1/\rho$ and $1/(\rho+L)$, respectively, we conclude
\begin{align}
    |u_{k+1}^i| \leq \left(1-\frac{\rho}{\rho+L}\right) |u_k^i| + | \hat{z}_k^i-\nabla_u \bar{f}^i(0,\hat{z}_k^i)|. \label{eq:rectmp}
\end{align}
By a similar argument based on Taylor's theorem, we can bound the last term in the previous equation by
\begin{align*}
| \hat{z}_k^i-\nabla_u \bar{f}^i(0,\hat{z}_k^i)|\leq |x^{i}_\ast| + \frac{L}{\rho+L} |\hat{z}_k^i|.
\end{align*}
Unrolling the recursion in \eqref{eq:rectmp} and exploiting the fact that $u_{0}^i=0$ yields
\begin{align*}
|u_k^i| \leq \frac{\rho+L}{\rho} |x^{i}_\ast| + \sup_{k\geq 0} | \hat{z}_k^i|,
\end{align*}
where the last term is bounded by $R$. Finally, due to the $1/\rho$-smoothness of $\bar{f}^i(\cdot,z)$, we conclude $|x_{k+1}^i|\leq |u_k^i|$, which yields the desired result as follows,
\begin{align*}
    |\chi^{di}|=|\alpha x^i_{k+1} +u^i_k |&\leq \alpha | x^i_{k+1} | +|u^i_k| \leq (\alpha+1) |u^i_k| \leq (\alpha+1)\left(\frac{\rho+L}{\rho} |x^{i}_\ast| + \sup_{k\geq 0} | \hat{z}_k^i| \right)\\&\leq (\alpha+1)\left(\frac{\rho+L}{\rho} |x^{i}_\ast| + R\right).
\end{align*}
\end{proof}

\newpage
\section{Diminishing Communication Threshold}\label{app:dec_delta}

In the main text, we focused our presentation on fixed communication thresholds. However, it is important to note that our approach and our analysis can be easily extended to the case where communication thresholds $\Delta$ are varied as a function of the number of iterations. For example, it is straightforward to show that for any vanishing sequence $\Delta_k$, our iterates indeed converge to the minimizer of \eqref{eqn:main_problem}. 

\begin{corollary}\label{lemma:modified_2}
Let the assumptions of Thm.~\ref{prop:main_result} be satisfied and let $\Delta_k \geq 0$ be such that $\Delta_k \rightarrow 0$ for $k \rightarrow \infty$. Then ${\lim _{k \rightarrow \infty}} |\xi_{k}-\xi_\ast|^2 \rightarrow 0$.
\end{corollary}
\begin{proof}
    According to Thm.~\ref{prop:main_result}, the following holds,
    \begin{align*}
V_{k+1}\leq \tau^2 V_k + \Bar{\sigma}(Q)\Delta_k^2,
\end{align*}
see \eqref{eqn:lyapunov_ineq}. We now apply Lemma 3 in Sec.~2.2 in \citep{Poliak_1987}, which yields the desired result.
\end{proof}

We can also derive an explicit convergence rate. In fact, the following corollary proves that if $\Delta_k^2$ is the form $q/(k+1)^t$, where $q>0$ and $t>0$ are constants and $k$ is the iteration number, $| \xi_{k}-\xi_\ast|^2$ converges at a rate of $\mathcal{O}(1/k^t)$.

\begin{corollary}
\label{lemma:chosen_delta}
Let the assumptions of Thm.~\ref{prop:main_result} be satisfied and let $\Delta_k^2\leq\frac{q}{(k+1)^t}$, $\forall k\geq 0$, $t>0$. Then, the following holds for all $k\geq 0$:
\begin{align*}
    |\xi_k-\xi_\ast|^2 \leq \frac{1}{\underaccent{\bar}{\sigma}(P)} \left(\frac{k_0}{k+k_0}\right)^t c_0,
\end{align*}
where $k_0=\frac{1}{\left(\frac{2}{1+\tau^2}\right)^t-1}$ and $c_0=\mathrm{max}\left\{\frac{2\Bar{\sigma}(Q)q}{1-\tau^2},\bar{\sigma}(P)|\xi_0-\xi_\ast|^2\right\}$.
\end{corollary}

\begin{proof} According to \eqref{eqn:lyapunov_ineq}, the following holds,
\begin{align*}
    V_{k+1}&\leq \tau^2V_k  + \bar{\sigma}(Q)\frac{q}{(k+1)^t}.
\end{align*}
We make the following claim:
\begin{align*}
    V_{k}&\leq c_0\left(\frac{k_0}{k+k_0}\right)^t, \quad \forall k \geq 0.
\end{align*}

We prove the claim by induction. The claim holds for $k=0$ due to the fact that $c_0\geq \bar{\sigma}(P)|\xi_0-\xi_\ast|^2$. We therefore assume that the claim holds for $k$ and show that this implies that the claim holds for $k+1$. This yields
\begin{align*}
V_{k+1}&\leq \tau^2V_k  + \bar{\sigma}(Q)\frac{q}{(k+1)^t}\\
&\leq \tau^2c_0\left(\frac{k_0}{k+k_0}\right)^t +\bar{\sigma}(Q) \frac{q}{(k+1)^t}\\
&\leq c_0\left(\frac{k_0}{k+k_0+1}\right)^t \left( \tau^2 \left(\frac{k+k_0+1}{k+k_0}\right)^t +\frac{\bar{\sigma}(Q) q}{c_0k_0^t}\left(\frac{k+k_0+1}{k+1}\right)^t \right)\\
&\leq c_0\left(\frac{k_0}{k+k_0+1}\right)^t \left( \tau^2 \left(\frac{k_0+1}{k_0}\right)^t +\frac{\bar{\sigma}(Q) q}{c_0}\left(\frac{k_0+1}{k_0}\right)^t \right)\\
&\leq c_0\left(\frac{k_0}{k+k_0+1}\right)^t,
\end{align*}
and completes the induction argument.

\end{proof}

\newpage
\section{Additional Experiments and Hyperparameters}\label{app:add_experiments}\label{app:hyperparameters}
We ran various experiments in order to assess the performance of the event-based distributed learning algorithm (Alg.~\ref{alg:over_relaxed_consensus}). In the comparative studies, we choose FedAvg \citep{mcmahan_communicationefficient_2017}, FedProx \citep{li_federated_2020}, SCAFFOLD \citep{Karimireddy_SCAFFOLD_2020} and FedADMM \citep{Zhou_Li_2023} as baselines, since these methods have been developed to address challenges such as data heterogeneity and communication efficiency. For a fair comparison in terms of computation resources in all setups, each of the agents are run for the same number of local gradient steps. 

We include the first example in Sec.~\ref{sec:results}, which showcases how two image classifiers (for MNIST and CIFAR-10 datasets) can be trained in a distributed and communication efficient way. Our setup included  $N\!=\!10$ agents for MNIST, each storing data for a single digit, resulting in the most extreme non-i.i.d. distribution of data among agents. For a CIFAR-10 classifier, the data are distributed among $N\!=\!100$ agents according to a Dirichlet distribution, i.e., we sample $p_a\! \sim \!\mathrm{Dir}N(\beta)$, where $N$ is the number of agents and $\beta\! =\! 0.5$. We then assign a $p_{a,j}$ proportion of the training data of class $a$ to agent $j$.

We applied our implementation to train a fully connected neural network on the MNIST dataset and a convolutional network on the CIFAR-10 dataset. The classifier model has 4 convolutional layers, each with $3\!\times\!3$ kernels and 32, 64, 128, and 256 filters, respectively, followed by three fully connected layers with ReLU activation functions. After each set of convolutions, a $2\!\times\!2$ max pooling layer is applied, followed by a ReLU activation. 
We train the MNIST classifier model using Alg.~\ref{alg:over_relaxed_consensus}, where we replace the full minimization step of each local objective with five steps of stochastic gradient descent with a learning rate of $l_\text{r}=10^{-1}$, and the CIFAR-10 model with 3 epochs of stochastic gradient descent (batch size 20, learning rate $l_\text{r}=10^{-3}$). Further hyperparameters are listed in Tabs.~\ref{tab:hyperparameters_mnist} and ~\ref{tab:hyperparameters_cifar}.

Tab.~\ref{tab:num_events} in Sec.~\ref{sec:results} summarizes the main result of this paper, by comparing the performance of different methods. From this table, it is clear that Alg.~\ref{alg:over_relaxed_consensus} achieves the same test accuracy with less communication cost. The communication configurations for Tab.~\ref{tab:num_events} are summarized in Tab.~\ref{tab:configs}.

\begin{table*}[b]
    \centering
    \begin{tabular}{c|ccc|cccc}
        \toprule
        \multirow{2}{*}{\textbf{Algorithm}} & \multicolumn{3}{c}{\textbf{MNIST Target Accuracy}} 
        & \multicolumn{4}{c}{\textbf{CIFAR-10 Target Accuracy}} \\
        \cmidrule{2-8}
        & \textbf{80\%} & \textbf{85\%} & \textbf{90\%}  
        & \textbf{70\%} & \textbf{75\%} & \textbf{77\%} & \textbf{78\%} \\
        \midrule
        \textbf{Alg.~\ref{alg:over_relaxed_consensus}-randomized} ($p_{\mathrm{trig}},\Delta^d$)
        & $(0.1,5)$ & $(0.1,4)$ & $(0.1,1)$
        &$(0.2,4.5)$
        & $(0.1,3.75)$ & $(0.2,3.5)$
 & $(0.7,3.75)$ \\
\textbf{Alg.~\ref{alg:over_relaxed_consensus}-Vanilla} ($\Delta^d$) 
        & $(3)$  & $(2)$  &$(1)$   
        &$(4.25)$ & $(3.25)$
        & $(3.25)$ & $(1.75)$\\
        \textbf{FedADMM} ($part\_rate$)
        & 0.4 & 0.6 & 1.0
        & 0.4 & 0.5 & 0.7 & 0.9 \\
        \textbf{FedAvg} ($part\_rate$)
        & 0.4 & 1.0 & - 
        & 0.1&-&-&- \\
        \textbf{FedProx} ($part\_rate$)
        & 0.5 & 1.0 & -  
        & 0.2 &-&-&-  \\
        \textbf{SCAFFOLD} ($part\_rate\times 2$)
        & $0.4 \times 2$  & $0.5 \times 2$ & $0.8 \times 2$  
        & $0.2 \times 2$ & - & - & - \\
        \bottomrule
    \end{tabular}
    \caption[Load]{ Communication configurations across algorithms. Values represent the probability of communication for baseline methods. SCAFFOLD values are doubled due to double package transmission per round.}
    \label{tab:configs} 
\end{table*}

\begin{figure}[h] 
\definecolor{mycolor1}{rgb}{0.00000,0.44700,0.74100}%
\definecolor{mycolor2}{rgb}{0.85000,0.32500,0.09800}%
\definecolor{mycolor3}{rgb}{0.92900,0.69400,0.12500}%
\definecolor{mycolor4}{rgb}{0.49400,0.18400,0.55600}%
\definecolor{mycolor5}{rgb}{0.46600,0.67400,0.18800}%
\definecolor{mycolor6}{rgb}{0.9290, 0.6940, 0.1250}%
\definecolor{mycolor7}{rgb}{0.63500,0.07800,0.18400}%
\pgfplotsset{every tick label/.append style={font=\small}}
\pgfplotsset{every axis label/.append style={font=\small}}
\pgfplotsset{every axis plot/.append style={line width=1.0pt}} % Set the line width for the whole axis
\pgfplotsset{every mark plot/.append style={scale=0.05}}
\begin{tikzpicture}

\begin{axis}[%
width=2.0in,
height=1.2in,
at={(0.01in,0.01in)},
scale only axis,
xmode=linear,
xmin=70,
xmax=92,
xminorticks=true,
xtick={60,70,75,80,85,90,100},
xticklabels={60,70,75,80,85,90,100},
xlabel style={font=\color{white!15!black}},
xlabel={Classification Accuracy (\%)},
xlabel near ticks,
ymode=linear,
ymin=0.0,
ymax=1.0,
yminorticks=true,
ytick={0,0.2,0.4,0.6,0.8,1},
yticklabels={0,20,40,60,80,100},
yminorticks=true,
ylabel style={font=\color{white!15!black}},
ylabel={Total Comm. Load ($\%$)},
ylabel near ticks,
axis background/.style={fill=white},
xmajorgrids,
xminorgrids,
ymajorgrids,
yminorgrids,
title={MNIST}
]
\addplot [color=mycolor1, mark=square, mark size=1.8pt, mark options={solid, mycolor1}, forget plot]
  table[row sep=crcr]{%
91.575	1\\
91.5708333333333	0.954545454545455\\
90.9791666666667	0.855050505050505\\
88.8833333333333	0.665656565656566\\
86.7833333333333	0.642929292929293\\
84.4958333333333	0.509090909090909\\
82.075	0.408080808080808\\
78.5083333333333	0.333333333333333\\
73.925	0.256565656565657\\
67.1333333333333	0.188383838383838\\
63.2	0.149494949494949\\
59.6583333333333	0.132323232323232\\
49.4416666666667	0.10959595959596\\
50.425	0.1\\
47.875	0.0893939393939394\\
42.525	0.0828282828282828\\
43.825	0.0883838383838384\\
38.9	0.047979797979798\\
42.325	0.0606060606060606\\
35.775	0.048989898989899\\
23.7	0.0383838383838384\\
26.5	0.0383838383838384\\
};
\addplot [color=mycolor1, mark=triangle, mark options={solid, mycolor1}, forget plot]
  table[row sep=crcr]{%
91.575	1\\
91.6041666666667	0.958080808080808\\
91.0583333333333	0.861616161616162\\
89.4541666666667	0.704545454545455\\
88.4875	0.609090909090909\\
87.3625	0.496464646464646\\
86.2625	0.426262626262626\\
85.2583333333333	0.346969696969697\\
81.175	0.314646464646465\\
79.7166666666667	0.27979797979798\\
78.1333333333333	0.235858585858586\\
%80.0625	0.221717171717172\\
77.5333333333333	0.215656565656566\\
77.2791666666667	0.212121212121212\\
76.4625	0.206565656565657\\
75.6208333333333	0.196464646464646\\
73.6541666666667	0.178282828282828\\
73.6875	0.172727272727273\\
75.7583333333333	0.177777777777778\\
%75.5625	0.163636363636364\\
72.1166666666667	0.164141414141414\\
%71.2541666666667	0.164141414141414\\
};
\addplot [color=mycolor2, mark=asterisk, mark options={solid, mycolor2}, forget plot]
  table[row sep=crcr]{%
13.1	0\\
58.7333333333333	0.1\\
74.7333333333333	0.2\\
77.4	0.3\\
84.8333333333333	0.4\\
82.8333333333333	0.5\\
85.6166666666667	0.6\\
87.0166666666667	0.7\\
87.5	0.8\\
89.0833333333334	0.9\\
88.7666666666667	1\\
};
\addplot [color=mycolor6, mark=asterisk, mark options={solid, mycolor6}, forget plot]
  table[row sep=crcr]{%
47.3416666666667	0.1\\
71.1958333333333	0.2\\
75.4083333333333	0.3\\
79.0333333333333	0.4\\
83.4583333333333	0.5\\
83.3666666666667	0.6\\
83.9291666666667	0.7\\
83.25	0.8\\
82.6583333333333	0.9\\
85.8416666666667	1\\
};
\addplot [color=mycolor7, mark=asterisk, mark options={solid, mycolor7}, forget plot]
  table[row sep=crcr]{%
46.4875	0.1\\
71.8916666666667	0.2\\
79.4708333333333	0.3\\
79.7166666666667	0.4\\
83.5458333333333	0.5\\
83.3583333333333	0.6\\
80.8416666666667	0.7\\
84.2083333333333	0.8\\
83.8291666666667	0.9\\
85.5958333333333	1\\
};

\addplot [color=mycolor5, mark=asterisk, mark options={solid, mycolor5}]
  table[row sep=crcr]{%
32.5041666666667	0.2\\
62.4291666666667	0.4\\
75.8208333333333	0.6\\
83.9708333333333	0.8\\
84.8875	1\\
88.4833333333333	1.2\\
89.5708333333333	1.4\\
90.6916666666667	1.6\\
91.4875	1.8\\
91.9	2.0\\
};

\end{axis}
\end{tikzpicture}
\definecolor{mycolor1}{rgb}{0.00000,0.44700,0.74100}%
\definecolor{mycolor2}{rgb}{0.85000,0.32500,0.09800}%
\definecolor{mycolor3}{rgb}{0.92900,0.69400,0.12500}%
\definecolor{mycolor4}{rgb}{0.49400,0.18400,0.55600}%
\definecolor{mycolor5}{rgb}{0.46600,0.67400,0.18800}%
\definecolor{mycolor6}{rgb}{0.30100,0.74500,0.93300}%
\definecolor{mycolor7}{rgb}{0.63500,0.07800,0.18400}%
\pgfplotsset{every tick label/.append style={font=\small}}
\pgfplotsset{every axis label/.append style={font=\small}}
\pgfplotsset{every axis plot/.append style={line width=1.0pt}} % Set the line width for the whole axis
\pgfplotsset{every mark plot/.append style={scale=0.05}}
\begin{tikzpicture}

\begin{axis}[%
width=2.0in,
height=1.2in,
at={(0.01in,0.01in)},
scale only axis,
xmode=log,
xmin=0.65,
xmax=0.80,
xminorticks=true,
xtick={0.6,0.65,0.7,0.75,0.8,0.85,0.9,0.95,1},
xticklabels={60,65,70,75,80,85,90,95,100},
xlabel style={font=\color{white!15!black}},
xlabel={Classification Accuracy (\%)},
xlabel near ticks,
ymode=linear,
ymin=0,
ymax=30000,
yminorticks=true,
ytick={0,6000,12000,18000,24000,30000},
yticklabels={0,20,40,60,80,100},
scaled ticks=false,
yminorticks=true,
ylabel style={font=\color{white!15!black}},
ylabel={Total Comm. Load ($\%$)},
ylabel near ticks,
axis background/.style={fill=white},
xmajorgrids,
xminorgrids,
ymajorgrids,
yminorgrids,
legend style={font=\small, anchor=north east, nodes={scale=1, transform shape},at={(1.7,1)}},
legend cell align={left},
legend columns=1,
title={CIFAR-10}
]

\addplot[color=mycolor1, mark=square, mark size=1.8pt, mark options={solid, mycolor1}]
 table[row sep=crcr]{0.6694	11534.5\\
0.7314	12214\\%0.7171	12517\\
0.7428	12657.3333\\
0.7472	13806\\
0.7723	14779.6667\\%0.7542	16954\\
0.7687	17440.3333\\%0.753	18568.6667\\%0.7644	19650.6667\\
0.7829	20690\\
0.7812 30000\\
};\addlegendentry{Alg.~\ref{alg:over_relaxed_consensus}-Vanilla};

\addplot[color=mycolor1, mark=triangle, mark size=1.8pt, mark options={solid, mycolor1}]
 table[row sep=crcr]
{0.7385	12531.3333	\\
0.7483	13192.3333	\\
0.7637	14090	\\%0.7545	14268.5	\\
0.7718	15008.5	\\%0.7558	16779.5	\\
0.774	17568	\\%0.7632	18749.6667	\\%0.7662	19738.6667	\\%0.7636	20978.6667	\\
0.7752	21846.3333	\\
0.7814	22273	\\
};\addlegendentry{Alg.~\ref{alg:over_relaxed_consensus}-Rand};

\addplot[color=mycolor2, mark=asterisk, mark size=1.8pt, mark options={solid, mycolor2}]  
table[row sep=crcr]{
0.1435	600\\
0.298	1500\\
0.4147	3000\\
0.5943	6000\\
0.6812	9000\\
0.7244	12000\\
0.753	15000\\
0.769	18000\\
0.7726	21000\\
0.7789	24000\\
0.7827	27000\\
0.7837	30000\\
};\addlegendentry{FedADMM};

\addplot[color=mycolor7, mark=asterisk, mark size=1.8pt, mark options={solid, mycolor7}]  
table[row sep=crcr]{
0.4909	600\\
0.6837	1500\\
0.6901	3000\\
0.7014	6000\\
0.7024	9000\\
0.7047	12000\\
0.7019	15000\\
0.6999	18000\\
0.7005	21000\\
0.7027	24000\\
0.6981	27000\\
0.7005	30000\\
};\addlegendentry{FedProx};

\addplot[color=mycolor3, mark=asterisk, mark size=1.8pt, mark options={solid, mycolor3}]  
table[row sep=crcr]{
0.4488	600\\
0.6887	1500\\
0.7028	3000\\
0.6992	6000\\
0.7046	9000\\
0.7024	12000\\
0.6974	15000\\
0.7014	18000\\
0.6964	21000\\
0.6917	24000\\
0.6971	27000\\
0.6964	30000\\
};\addlegendentry{FedAvg};

\addplot[color=mycolor5, mark=asterisk, mark size=1.8pt, mark options={solid, mycolor5}]  
table[row sep=crcr]{
0.5071	1200\\
0.6749	3000\\
0.6821	6000\\
0.7079	12000\\
0.6987	18000\\
0.7007	24000\\
0.6987	30000\\
0.7021	36000\\
0.7025	42000\\
0.7011	48000\\
0.6992	54000\\
0.7001	60000\\
};\addlegendentry{SCAFFOLD};
\end{axis}

\end{tikzpicture}%
    \caption{The figure compares different federated learning methods on the MNIST and CIFAR-10 datasets with respect to the resulting trade-off between total communication load and classification accuracy on the test set. In the MNIST case (left), randomization includes agent-to-server communication with 0.1 probability. For CIFAR-10 (right), randomization incorporates server-to-agent communication with 0.2 probability. Points along each curve represent different $\Delta$ thresholds, demonstrating the relationship between communication reduction and model accuracy.}
    \label{fig:Method_comparison}
\end{figure}
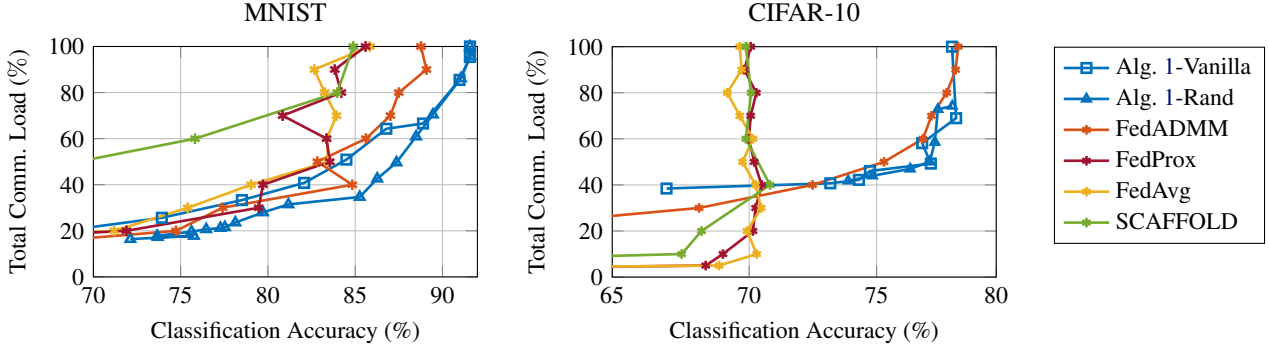

Fig.~\ref{fig:Method_comparison} illustrates the trade-off between accuracy and communication load.
The results demonstrate that our event-based approach consistently achieves higher accuracy with fewer communication events compared to baselines. Each point in Fig.~\ref{fig:Method_comparison} represents a different value of $\Delta$, where $\Delta$ monotonically increases along the curve, demonstrating that with our algorithm and a well-chosen $\Delta$ threshold, communication among agents can be reduced while still achieving a high classification accuracy. Our experimental results indicate that the approach can reduce communication costs by over 30\% without significant accuracy degradation. Notably, SCAFFOLD doubles the communication cost due to its dual-package communication protocol. These findings directly translate to the results in Tab.~\ref{tab:num_events} showing total communication events for target accuracies. The extensive experimentation across both small-scale (MNIST) and large-scale (CIFAR-10) scenarios demonstrates the scalability and effectiveness of our event-based communication strategy, particularly for large-scale distributed learning problems.

\begin{table}[h]
    \centering
    \caption{The table summarizes the hyperparameters used for distributed training of MNIST classifier (Fig.~\ref{fig:Method_comparison}, Tab.~\ref{tab:num_events})}
    \label{tab:hyperparameters_mnist}
    \begin{tabular}{l l}
        \toprule
        \textbf{Hyperparameter} & \textbf{Value} \\
        \midrule
        number of agents (\(N\))  & 10 \\  
        size of neural network layers & [400, 200, 10] \\
        learning rate (gradient descent step-size) & 0.1 \\ 
        number of iterations& 100\\  
        $\Delta^d=\Delta$, $\Delta^z=0.1\times \Delta$& range between $\left[0,10 \right]$ \\
        $\mu$ (FedProx) & $0.1$\\
      augmented lagrangian parameter ($\rho$)  (FedADMM, Alg.~\ref{alg:over_relaxed_consensus}) &  1\\
        $n_g$ (SCAFFOLD) & 1 \\
        \bottomrule
    \end{tabular}
\end{table}
\begin{table}[h!]
    \centering
    \caption{The table summarizes the hyperparameters used for the distributed training of CIFAR-10 classifier (Fig.~\ref{fig:Method_comparison}, Tab.~\ref{tab:num_events})}
    \label{tab:hyperparameters_cifar}
    \begin{tabular}{l l}
        \toprule
        \textbf{Hyperparameter} & \textbf{Value} \\
        \midrule
        number of agents (\(N\)) & 100 \\
        augmented lagrangian parameter ($\rho$) (FedADMM, Alg.~\ref{alg:over_relaxed_consensus}) & 0.01 \\
        learning rate & 0.01 \\
        momentum  & 0.9 \\
        number of iterations & 150\\
        number of local epochs & 3\\
        batch size & 20\\
        $\Delta^d=\Delta$, $\Delta^z=0.01\times \Delta$ & range between $[0,4]$\\
         $\mu$ (FedProx) & $0.1$\\
         $n_g$ (SCAFFOLD) & 1 \\
        \bottomrule
    \end{tabular}
\end{table}

%%%%%%%%%%%%%%%%%%%%%%%%%%%%%%%%%%%%%%%%%%%%%%%

The next sections provide additional numerical experiments. We first show an example based on LASSO where the local objectives are strongly convex (Sec.~\ref{app:lin_lasso} and \ref{app:drop_lasso}). In this setup, our theoretical results apply. Sec.~\ref{app:graph_exp} shows how our algorithm can train an MNIST classifier, when only local communications are allowed, as specified by a given communication graph. In such a setup, the baselines FedAvg, FedProx, SCAFFOLD and FedADMM are not applicable.

\subsection{Linear Regression and LASSO with Non-i.i.d. Data}\label{app:lin_lasso}

\begin{figure}[h]
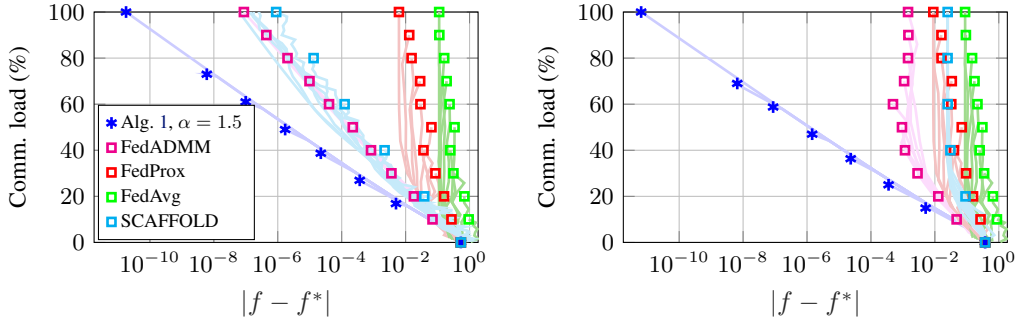

  \centering
   \input{figs/experiments/Noniid_LS}  \input{figs/experiments/Noniid_lasso_eps}
    \caption{The figure shows the communication load versus accuracy trade-off for the different methods applied to two distinct problems derived from \eqref{eqn:lasso}: linear regression ($\lambda=0$, left panel), and on the right, LASSO ($\lambda=0.1$, right panel).} 
\label{fig:LS}
\end{figure}

We conduct numerical experiments based on the following distributed learning problem:
\begin{align}
\begin{aligned}
&\underset{x\in \mathbb{R}^n,z\in \mathbb{R}^n}{\min} \sum_{i=1}^N \frac{1}{2}|A^i x^i-b^i|^2 +\lambda|z|_1,\\
&\text {subject to }~ x^i-z=0, \quad i=1, \ldots, N,
\label{eqn:lasso}
\end{aligned}
\end{align}
where \( A^i \in \mathbb{R}^{m\times n} \), \( b^i \in \mathbb{R}^{m} \). In the data generation process, we generate samples from three different distributions: a standard normal distribution, a Student's t distribution with one degree of freedom, and a uniform distribution in the range \([-5, 5]\). These samples are concatenated to form a single dataset, which is then partitioned into subsets for each agent \( i \) to obtain \( (A^i, b^i) \). Finally, we normalize the feature vectors and target values for each agent to prepare the data for the learning problem.
In this non-i.i.d. setting, local optimal points of individual agents $x^i_\ast$ are far away from each other, and their average $\sum_{i=1}^N x^i_\ast/N$ is also far away from the global optima $x_\ast$. The experiments were run for $T_\mathrm{max}=50$ steps, which are required for Alg.~\ref{alg:over_relaxed_consensus} to converge to the global optimal point with high accuracy. Fig.~\ref{fig:LS} illustrates the communication load against the absolute difference between the objective function value $f$ and the optimal value $f^*$, where the communication load is defined as the number of communications accumulated over time.

In the first scenario, we set $\lambda=0$ to obtain a linear regression problem, where the proposed algorithm with relaxation parameter $\alpha=1.5$ clearly outperforms baseline methods by a large margin. We note that the gap between the global and local optimal points prevents FedAvg and FedProx from converging to the optimal point $f^\ast$.

For the second case, we set $\lambda=0.1$ to solve the LASSO problem. By assumption, FedAvg, SCAFFOLD and FedADMM require the local objective functions to be smooth. However, we allow handling nonsmooth local objective functions, which is relevant to the distributed learning problems with $\ell^1$ regularization. To avoid a noncontinuous gradient for the local minimization for SCAFFOLD, FedADMM, FedAvg and FedProx, the local update step is carried out by the following local gradient,
\begin{align}   
\nabla_{x^i}\Tilde{f}^i(x^i)=A^{i\top}(A^i x^i-b^i) + \frac{\lambda}{N}\begin{cases}
   \mathrm{sgn}(x^i) & |x^i|>\delta\\
    \frac{1}{\delta}x^i  & |x^i|\leq\delta\\
\end{cases},
 \end{align}
where $\delta$ can be chosen as small as $1e-16$ (double precision machine epsilon). However, we found that the results are largely unaffected by the choice of $\delta$.

\begin{table}[h]
    \centering
    \caption{The table summarizes the hyperparameters used for the distributed linear regression and LASSO experiments (Fig.~\ref{fig:LS}).}
    \label{tab:hyperparameters_lasso}
    \begin{tabular}{l l}
        \toprule
        \textbf{Hyperparameter} & \textbf{Value} \\
        \midrule
        number of agents (\(N\)) & 50 \\
        augmented lagrangian parameter (\(\rho\)) & 1 \\
        gradient descent step-size & 1 \\
        number of iterations & 50\\
        $\Delta^d=\Delta^z=\Delta$ & range between $[0,10^{-2}]$\\
        \bottomrule
    \end{tabular}
\end{table}

\subsection{Effect of Communication Drops} \label{app:drop_lasso}

\begin{figure}[H]
    \centering
  \input{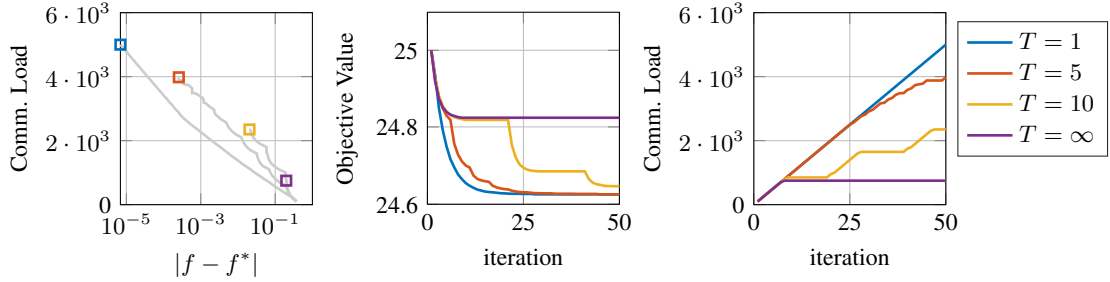}
    \caption{The left panel presents the trajectory of communication load versus suboptimality of the objective function value. The panel in the center shows the evolution of the objective function for different values of the reset period for a drop rate of 0.3 and for $\Delta=10^{-3}$, whereas the right panel shows the cumulative communication load over time in addition to the reset communication at each $T$ step.}
    \label{fig:LASSO_exp_drop}
\end{figure}

To observe the effect of communication drops, we repeated the same LASSO experiment in \eqref{eqn:lasso} with hyperparameters in Tab.~\ref{tab:hyperparameters_drop}, but this time, we allow the transmission of information from the agents to the server to fail with a probability of $0.3$. As seen in the second panel of Fig.~\ref{fig:LASSO_exp_drop}, if we have no reset, i.e., $T=\infty$, the algorithm cannot converge and a significant error remains. On the left panel, the trade-off between communication load and suboptimality is presented. More frequent reset operations lead to a faster convergence and less error, in exchange for additional communication cost that comes with the reset. 

\begin{table}[h]
    \centering
    \caption{The table summarizes the hyperparameters used for the distributed LASSO experiment against communication drops (Fig.~\ref{fig:LASSO_exp_drop}).}
    \label{tab:hyperparameters_drop}
    \begin{tabular}{l l}
        \toprule
        \textbf{Hyperparameter} & \textbf{Value} \\
        \midrule
        number of agents (\(N\)) & 50 \\
        L1 regularization parameter (\(\lambda\)) & 0.1 \\
        augmented lagrangian parameter (\(\rho\)) & 1 \\
        relaxation parameter (\(\alpha\)) & 1 \\
        gradient descent step-size & 1 \\
        number of iterations& 50\\
        $\Delta^d=\Delta^z=\Delta$ & $10^{-3}$\\
        \bottomrule
    \end{tabular}
\end{table}

\subsection{Distributed Training on a Graph}\label{app:graph_exp}
 
Our distributed learning algorithm, Alg.~\ref{alg:relaxed_event}, is general enough to train a machine learning classifier over a network of agents; the network structure can be encoded by a proper selection of the linear constraint matrices $A$ and $B$ (see App.~\ref{app:comm} for further details). Our framework therefore generalizes well beyond server-client structures, and our theoretical analysis also captures the influence of the network structure on the resulting convergence rate.

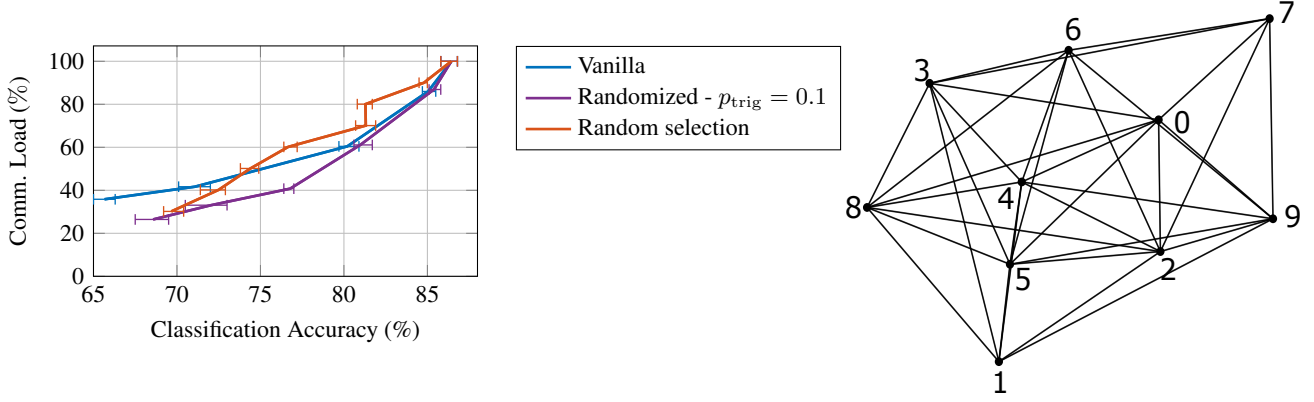
\begin{figure}[h]
    \begin{minipage}{0.65\textwidth}
    \centering

\definecolor{mycolor1}{rgb}{0.00000,0.44700,0.74100}%
\definecolor{mycolor2}{rgb}{0.85000,0.32500,0.09800}%
\definecolor{mycolor3}{rgb}{0.92900,0.69400,0.12500}%
\definecolor{mycolor4}{rgb}{0.49400,0.18400,0.55600}%
\definecolor{mycolor5}{rgb}{0.46600,0.67400,0.18800}%
\definecolor{mycolor6}{rgb}{0.30100,0.74500,0.93300}%
\definecolor{mycolor7}{rgb}{0.63500,0.07800,0.18400}%

\pgfplotsset{every tick label/.append style={font=\small}}
\pgfplotsset{every axis label/.append style={font=\small}}
\pgfplotsset{every axis plot/.append style={line width=1.0pt}} % Set the line width for the whole axis
\pgfplotsset{every mark plot/.append style={scale=0.05}}

\begin{tikzpicture}

\begin{axis}[%
width=2.0in,
height=1.2in,
at={(0.001in,0.001in)},
scale only axis,
xmode=linear,
xmin=65,
xmax=88,
xtick={60,65,70,75,80,85},
xticklabels={60,65,70,75,80,85},
xlabel near ticks,
xminorticks=true,
xlabel={Classification Accuracy (\%)},
ytick={0,14,28,42,56,70},
yticklabels={0,20,40,60,80,100},
ymin=0,
ymax=75,
ylabel={Comm. Load ($\%$)},
ylabel style={at={(-0.00,0.5)}, font=\small},
ylabel near ticks,
axis background/.style={fill=white},
title style={font=\bfseries},
xmajorgrids,
xminorgrids,
ymajorgrids,
yminorgrids,
legend style={font=\small, anchor=north west, nodes={scale=1, transform shape},at={(1.1,1)}},
legend cell align={left}
]
\addplot [color=mycolor1]
  table[row sep=crcr]{%
86.43	70.000\\
85.04	60.170\\
80.2	42.302\\
71.07	29.166\\
65.66	25.079\\
};\addlegendentry{Vanilla};
\addplot [color=mycolor1, forget plot]
 plot [error bars/.cd, x dir=both, x explicit, error bar style={line width=0.5pt}, error mark options={line width=0.5pt, mark size=2.0pt, rotate=90}]
 table[row sep=crcr, x error plus index=2, x error minus index=3]{%
86.43	70.000	0.370000000000005	0.629999999999995\\
85.04	60.170	0.460000000000008	0.339999999999989\\
80.2	42.302	0.700000000000003	0.5\\
71.07	29.166	0.930000000000007	0.969999999999999\\
65.66	25.079	0.640000000000001	0.659999999999997\\
};
\addplot [color=mycolor4]
  table[row sep=crcr]{%
86.43	70.000\\
85.39	60.815\\
80.98	42.768\\
76.78	28.534\\
72.04	23.152\\
68.57	18.514\\
};\addlegendentry{Randomized - $p_{\mathrm{trig}}=0.1$};
\addplot [color=mycolor4, forget plot]
 plot [error bars/.cd, x dir=both, x explicit, error bar style={line width=0.5pt}, error mark options={line width=0.5pt, mark size=2.0pt, rotate=90}]
 table[row sep=crcr, x error plus index=2, x error minus index=3]{%
86.43	70.000	0.370000000000005	0.629999999999995\\
85.39	60.815	0.409999999999997	0.290000000000006\\
80.98	42.768	0.719999999999985	0.379999999999995\\
76.78	28.534	0.219999999999999	0.379999999999995\\
72.04	23.152	0.960000000000008	1.53999999999999\\
68.57	18.514	0.929999999999993	1.07000000000001\\
};
\addplot [color=mycolor2]
  table[row sep=crcr]{%
69.66	21.125\\
72.49	28.130\\
74.4	35.110\\
76.66	42.127\\
81.31	49.069\\
81.29	56.045\\
84.79	63.006\\
86.43	70.000\\
};\addlegendentry{Random selection};
\addplot [color=mycolor2, forget plot]
 plot [error bars/.cd, x dir=both, x explicit, error bar style={line width=0.5pt}, error mark options={line width=0.5pt, mark size=2.0pt, rotate=90}]
 table[row sep=crcr, x error plus index=2, x error minus index=3]{%
69.66	21.125	0.739999999999995	0.460000000000008\\
72.49	28.130	0.410000000000011	1.08999999999999\\
74.4	35.110	0.5	0.600000000000009\\
76.66	42.127	0.540000000000006	0.259999999999991\\
81.31	49.069	0.590000000000003	0.609999999999985\\
81.29	56.045	0.409999999999997	0.489999999999981\\
84.79	63.006	0.210000000000008	0.289999999999992\\
86.43	70.000	0.370000000000005	0.629999999999995\\
};
\end{axis}

\end{tikzpicture}%
    \end{minipage} 
    \begin{minipage}{0.35\textwidth}
    \centering
    \includegraphics[width=\textwidth]{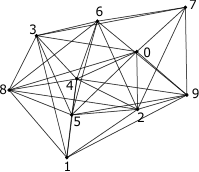}
    \end{minipage} 
    \caption{The figure shows a comparison of the vanilla event-based and the randomized event-based communication strategy (see Sec.~\ref{sec:eventbased}) with a purely random selection of agents. The outcome of a purely random strategy is consistently worse with respect to the resulting trade-off between communication load and classification accuracy.  The right panel visualizes the agent network with ten agents connected with 70 edges.}
    \label{fig:Mnist_on_graph}
\end{figure}

In order to highlight the versatility, we train an MNIST Classifier over a network of agents. We use a multi-layer perceptron that has the same structure as in Sec.~\ref{sec:results} and consider a situation where each agent has only access to the training data of a single digit. Fig.~\ref{fig:Mnist_on_graph} shows the resulting communication load and classification accuracy trade-off on the entire dataset (left), whereas the diagram on the right shows the network structure (only communication along the edges of the graph is allowed). The error bars indicate the range (minimum and maximum) of the classification accuracy among the different agents. 

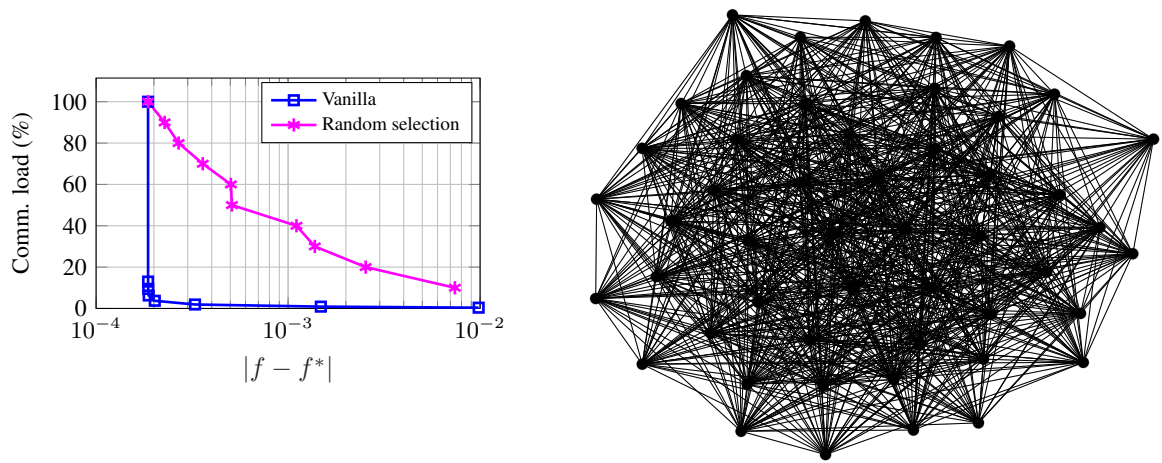
\begin{figure}[]
    \begin{minipage}{0.5\textwidth}
    \centering 
\definecolor{mycolor1}{rgb}{1.00000,0.00000,1.00000}%
\definecolor{mycolor2}{rgb}{0.93725,0.31176,0.94118}%
\definecolor{mycolor3}{rgb}{0.62745,0.85882,0.55686}%
\definecolor{mycolor4}{rgb}{0.95686,0.76078,0.76078}%
\pgfplotsset{every tick label/.append style={font=\small}}
\pgfplotsset{every axis label/.append style={font=\small}}
\pgfplotsset{every axis plot/.append style={line width=1.0pt}} % Set the line width for the whole axis
\pgfplotsset{every mark plot/.append style={scale=0.05}}

\begin{tikzpicture}

\begin{axis}[%
width=2.0in,
height=1.2in,
at={(0.416in,0.359in)},
scale only axis,
xmode=log,
xmin=1e-4,
xmax=0.01,
xtick={1e-5,1e-4,1e-3,1e-2,1e-1,1},
xticklabels={$10^{-5}$,$10^{-4}$,$10^{-3}$,$10^{-2}$,$10^{-1}$,$10^{0}$},
xminorticks=true,
xlabel style={font=\color{white!15!black}},
xlabel={$|f-f^*|$},%ytick={0, 6676904.587, 13353809.17, 20030713.76, 26707618.35, 33384522.94},
ytick={0, 5990800, 11981600, 17972400, 23963200, 29954000},
yticklabels={0,20,40,60,80,100},
ymin=0,
ymax=33384523,
ymode=linear,
ylabel style={font=\color{white!15!black}},
ylabel={Comm. load (\%)},
ylabel near ticks,
scaled ticks=false,
axis background/.style={fill=white},
xmajorgrids,
xminorgrids,
ymajorgrids,
legend style={font=\small, anchor=east, nodes={scale=0.8, transform shape},at={(0.98,0.85)}},
legend cell align={left}
]
\addplot [color=blue, line width=1.0pt, mark size=1.8pt, mark=square, mark options={solid, blue}]
  table[row sep=crcr]{%
0.000186484730534175	29954000\\
0.000186544647871045	3895734\\
0.000186848259115635	2830395\\
0.000187838846347432	1881478\\
0.000202128248677269	1110313\\
0.000327446362099693	573274\\
0.00148075438548645	263465\\
0.00982772418635491	111752\\%0.0631831219473433	45569\\%0.533619409801531	0\\
};\addlegendentry{Vanilla};
\addplot [color=mycolor1, line width=1.0pt, mark size=2.5pt, mark=asterisk, mark options={solid, mycolor1}]
  table[row sep=crcr]{%%0.533619409801531	0\\
0.0073902198059308	2995838\\
0.00253619747298828	5989375\\
0.00137753440029442	8984933\\
0.00110566164294212	11981152\\
0.000509713091233266	14976512\\
0.000503519300679045	17972789\\
0.000359305552155575	20967068\\
0.000268962782943305	23963317\\
0.000227515274040968	26959380\\
0.000186484730534175	29954000\\
};\addlegendentry{Random selection};

\end{axis}

\end{tikzpicture}%
  \end{minipage} 
    \begin{minipage}{0.44\textwidth}
    \centering
\includegraphics[width=\textwidth]{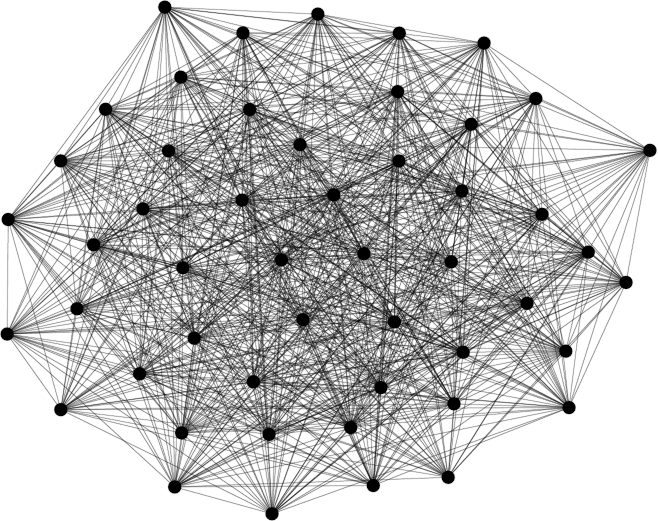}
    \end{minipage} 
  \caption{The first panel shows the comparison of the communication load versus solution accuracy for different communication methods applied to the linear regression problem derived from \eqref{eqn:lasso} ($\lambda=0$). The right panel visualizes the agent network with 50 agents connected with 1762 edges.}
    \label{fig:LS_on_large_graph}
\end{figure}

The results shown in Fig.~\ref{fig:Mnist_on_graph} and  highlight that a purely random selection of agents (suggested in \citep{Yu_Freris_2023_CE} ) results in a worse trade-off curve, which further motivates our event-based strategy. We also apply our algorithm to a much larger distributed learning problem with 50 agents and where the corresponding accuracy versus communication trade-off is shown in Fig.~\ref{fig:LS_on_large_graph}, together with the agent network that has been used.

\begin{table}[h]
    \centering
    \caption{The table summarizes the hyperparameters used for the distributed training of MNIST classifier over a graph (Fig.~\ref{fig:Mnist_on_graph}).}
    \label{tab:hyperparameters_mnist_graph}
    \begin{tabular}{l l}
        \toprule
        \textbf{Hyperparameter} & \textbf{Value} \\
        \midrule
        number of agents (\(N\))  & 10 \\  
        size of neural network layers & [400, 200, 10] \\
        learning rate (gradient descent step-size) & $5\times 10^{-3}$ \\ % 
        augmented lagrangian parameter (\(\rho\))& $5\times 10^{-3}$\\
        number of iterations& $10^{3}$\\  
        number of gradient steps per iteration & 5 \\
        $\Delta^x$ & range between $[0.0,0.2]$\\ 
        \bottomrule
    \end{tabular}
\end{table}

\begin{table}[H]
    \centering
    \caption{The table summarizes the hyperparameters used for the distributed linear regression experiment over a graph (Fig.~\ref{fig:LS_on_large_graph}).}
    \label{tab:hyperparameters_lasso_graph}
    \begin{tabular}{l l}
        \toprule
        \textbf{Hyperparameter} & \textbf{Value} \\
        \midrule
        number of agents (\(N\)) & 50 \\
        augmented lagrangian parameter (\(\rho\)) & $10^{-5}$ \\
        number of iterations & $17\times 10^{3}$\\
        $\Delta^x$ & range between $[0,1]$\\
        \bottomrule
    \end{tabular}
\end{table}

\end{document}